\documentclass[applsci,article,accept,moreauthors,pdftex]{Definitions/mdpi}

\firstpage{1}
\pubvolume{xx}
\issuenum{1}
\articlenumber{5}
\pubyear{2020}
\copyrightyear{2020}
\history{Received: 15 October 2020; Accepted: 22 October 2020; Published: date}
\updates{yes} 

\pdfoutput=1

\usepackage[utf8]{inputenc}

\usepackage{graphicx}
\usepackage{subcaption}
\usepackage{caption}
\usepackage{textcomp}
\usepackage{sidecap}
\usepackage{mathtools}
\usepackage{amsmath}
\usepackage{float}
\usepackage{multirow}
\usepackage{xspace}
\usepackage{tikz}
\usepackage{hyperref}

\Title{Integration of the 3D Environment for UAV Onboard 
	Visual Object Tracking}

\Author{St\'ephane Vujasinovi\'c~*\orcidA{}, Stefan~Becker\orcidB{}, Timo~Breuer, Sebastian~Bullinger\orcidD{}, Norbert~Scherer-Negenborn, and Michael Arens\orcidC{}}

\AuthorNames{St\'ephane Vujasinovi\'c, Stefan~Becker, Timo~Breuer, Sebastian~Bullinger, Norbert~Scherer-Negenborn, and Michael Arens}

\address[1]{Fraunhofer IOSB {(Fraunhofer IOSB is member of the Fraunhofer Center for Machine Learning)},  
\mbox{Institute~of Optronics}, Image~Exploitation, and~System Technologies, 76275 Ettlingen, 
 Germany; stefan.becker@iosb.fraunhofer.de (S.B.); timo.breuer@iosb.fraunhofer.de (T.B.); sebastian.bullinger@iosb.fraunhofer.de (S.B.); norbert.scherer-negenborn@iosb.fraunhofer.de (N.S.-N.); michael.arens@iosb.fraunhofer.de (M.A.)}

\corres{\hangafter=1 \hangindent=1.05em \hspace{-0.82em}Correspondence: stephane.vujasinovic{@}iosb.fraunhoder.de}

\abstract{Single visual object tracking from an unmanned aerial vehicle (UAV) poses fundamental challenges such as object occlusion, small-scale objects, background~clutter, and abrupt camera motion. To~tackle these difficulties, we~propose to integrate the 3D structure of the observed scene into a detection-by-tracking algorithm. We~introduce a pipeline that combines a model-free visual object tracker, a~sparse 3D reconstruction, and~a state estimator. The~3D reconstruction of the scene is computed with an image-based Structure-from-Motion (SfM) component that enables us to leverage a state estimator in the corresponding 3D scene during tracking. By~representing the position of the target in 3D space rather than in image space, we~stabilize the tracking during ego-motion and improve the handling of occlusions, background~clutter, and small-scale objects. We~evaluated our approach on prototypical image sequences, captured~from a UAV with low-altitude oblique views. For~this purpose, we~adapted an existing dataset for visual object tracking and reconstructed the observed scene in 3D. The~experimental results demonstrate that the proposed approach outperforms methods using plain visual cues as well as approaches leveraging image-space-based state estimations. We~believe that our approach can be beneficial for traffic monitoring, video~surveillance, and navigation.}


\begin{document}
\newcommand{\figref}[1]{Figure~\ref{#1}}
\newcommand{\tabref}[1]{Table~\ref{#1}}
\newcommand{\algref}[1]{Algorithm~\ref{#1}}
\newcommand{\secref}[1]{Section~\ref{#1}}
\newcommand{\meqref}[1]{Equation~\eqref{#1}}

\newcommand{\veccs}[1]{\ensuremath{\mathbf{#1}}}
\newcommand{\veccsdos}[2]{\ensuremath{\mathbf{#1}_#2}}
\newcommand{\veccstres}[3]{\ensuremath{\mathbf{#1}_#2^#3}}


\makeatletter
\DeclareRobustCommand\onedot{\futurelet\@let@token\@onedot}
\def\@onedot{\ifx\@let@token.\else.\null\fi\xspace}                
\def\eg{\emph{e.g}\onedot}             
\def\Eg{\emph{E.g}\onedot}
\def\ie{\emph{i.e}\onedot}             
\def\Ie{\emph{I.e}\onedot}
\def\cf{\emph{cf}\onedot}                 
\def\Cf{\emph{Cf}\onedot}
\def\qv{\emph{q.v}\onedot}             
\def\Eg{\emph{Q.v}\onedot}
\def\etc{\emph{etc}\onedot}             
\def\vs{\emph{vs}\onedot}
\def\wrt{w.r.t\onedot}                  
\def\dof{d.o.f\onedot}
\def\etal{\emph{et al}\@onedot}          
\makeatother




\section{Introduction}
\label{introduction}
In recent years, unmanned aerial vehicles (UAVs) have expanded in usage conjointly with the number of applications they provide, such~as video surveillance, traffic~monitoring, aerial~photography, {wildlife protection, cinematography, target~following, disaster~response,} and even delivery.
Initially used in the military field, their~use has gradually become widespread in the civil and commercial field, allowing~new applications to emerge, which~incorporate or eventually will incorporate visual object tracking as a core component.

{Single} visual object tracking is a long-studied computer vision problem relevant for many real-world applications. Its~goal is to estimate the location of an object in an image sequence, given~its initial location at the beginning. By~integrating a state estimator in the tracking process, the~tracking pipeline is referred to as detection-by-tracking; and~without, as tracking-by-detection~\cite{bib:black}.
Despite~solving challenging tasks to a certain extent---e.g., illumination changes, motion~blur, scale~variation---by~using deep learning for visual object tracking, there~are still situations that remain difficult to solve, e.g., partial and full occlusion, abrupt~object motions, or~background clutter.
Current state-of-the-art {single} visual object tracking algorithms follow mainly the tracking-by-detection paradigm, where~the location of the object is estimated based on a maximum-likelihood approach, inferred~from comparing an appearance model of the object with a small search region. For~close-range tracking scenarios, adding~a state estimator often leads to suboptimal performance due~to the problem of filter tuning~\cite{bib:vot_2019}.

\begin{figure}[H]
    \centering
    \includegraphics[trim=0.2cm 0.2cm 0.0cm 0.2cm, clip=true, width=0.95\linewidth]{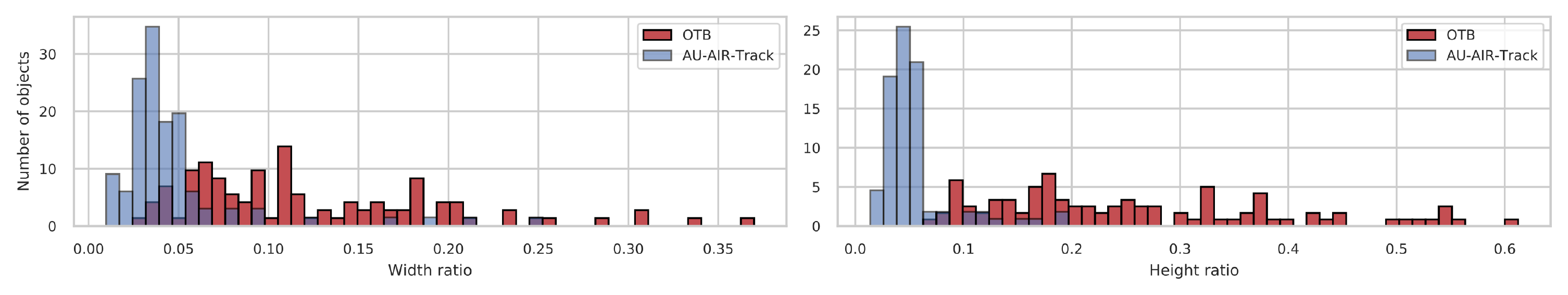}
    \caption{Distribution of bounding box sizes \wrt their image resolution. A~comparison between the {Online Tracking Benchmark (OTB)}~\cite{bib:otb100} in red (ground-level perspective), and the AU-AIR-Track 
    	dataset in blue (unmanned aerial vehicle (UAV) perspective). {A detailed description of our dataset AU-AIR-Track is given in \secref{AU-AIR-Track}.}}
    \label{fig:OTB_vs_AU_prob_dist}
\end{figure}

Despite the general progress of visual object trackers, these~approaches are usually not optimal or suffer from particular limitations when being applied onboard a UAV.
{As highlighted in the Single-Object Tracking~(SOT) challenge of VisDrone 2019~\cite{bib:visdrone}, the~major difficulties for state-of-the-art SOT on UAV scenarios are caused by abrupt camera motion, background~clutter, small-scale objects, and occlusions.}
For instance, objects~being tracked in image sequences captured from a low-altitude oblique view are subject to multiple occlusions caused~by environmental structures such as trees or buildings.
Additionally, {the size of an object} (\wrt to the image resolution) is relatively small in UAV sequences compared to a ground-level perspective as illustrated in \figref{fig:OTB_vs_AU_prob_dist}. {This potentially results in a less powerful discriminative appearance model.
The elevated viewpoint of the scene can also lead to a greater number of objects present simultaneously in the search area. Coupled~with a less-effective appearance model of the object, the~tracker is more inclined to mismatch the object during tracking with a similar background object.
Ego-motion also enhances the difficulties for most image-based tracking, since~the trackers are not directly designed to compensate for camera motion when inferring the object position.}

\begin{figure}[H]
    \centering
    \begin{subfigure}{0.28\linewidth}
        \begin{tikzpicture}
            \node[anchor=south west,inner sep=0] (image) at (0,0) {\includegraphics[width=\linewidth, bb=0 0 1920 1080]{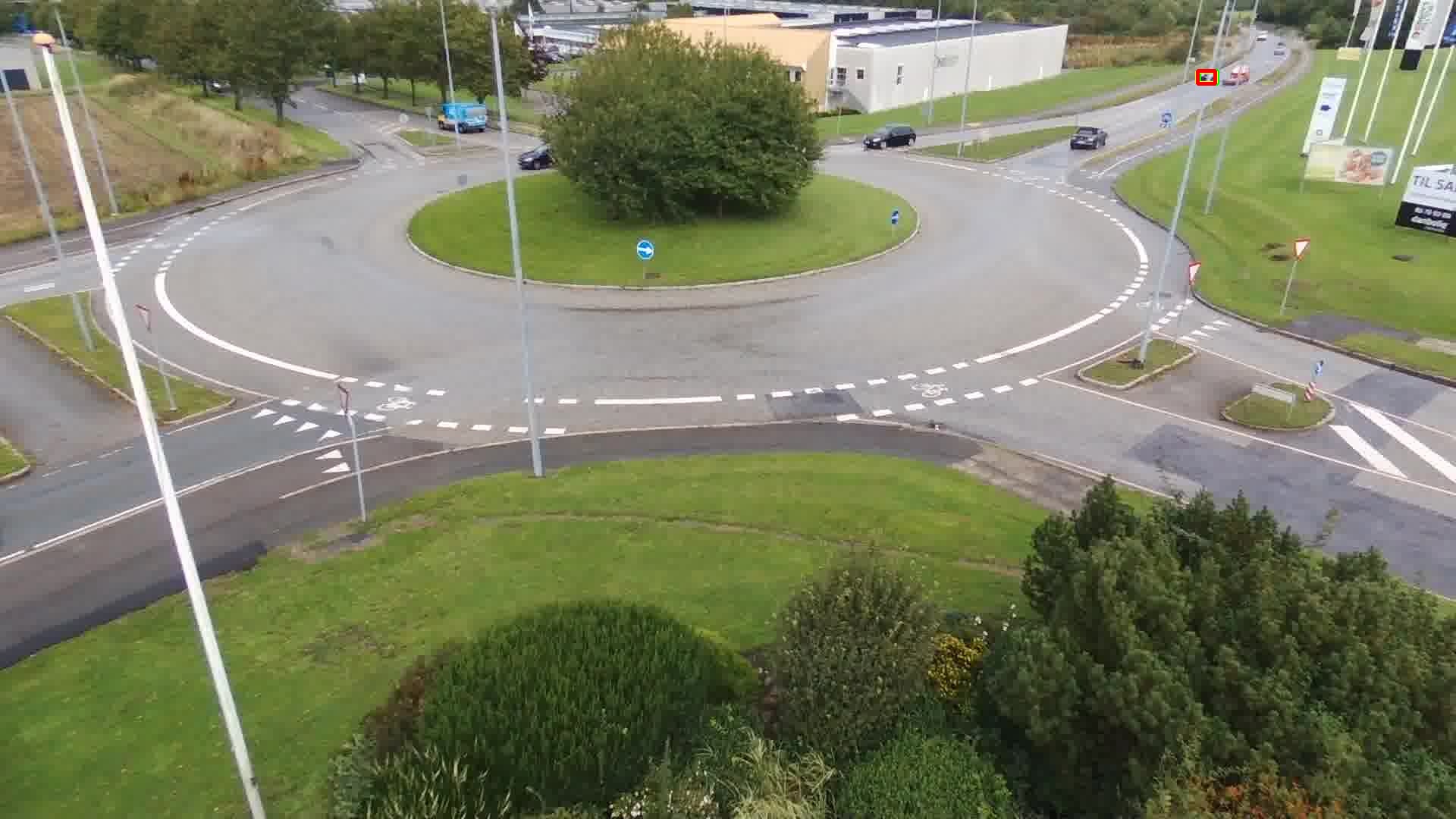}};
            \begin{scope}[x={(image.south east)},y={(image.north west)}]
                \draw[red, thick, dashdotted] (0.74,0.78) rectangle (0.999,0.999);
            \end{scope}
        \end{tikzpicture}
    \end{subfigure}\hspace{0.0005em}
    \begin{subfigure}{0.28\linewidth}
        \begin{tikzpicture}
            \node[anchor=south west,inner sep=0] (image) at (0,0) {\includegraphics[width=\linewidth, bb=0 0 1920 1080]{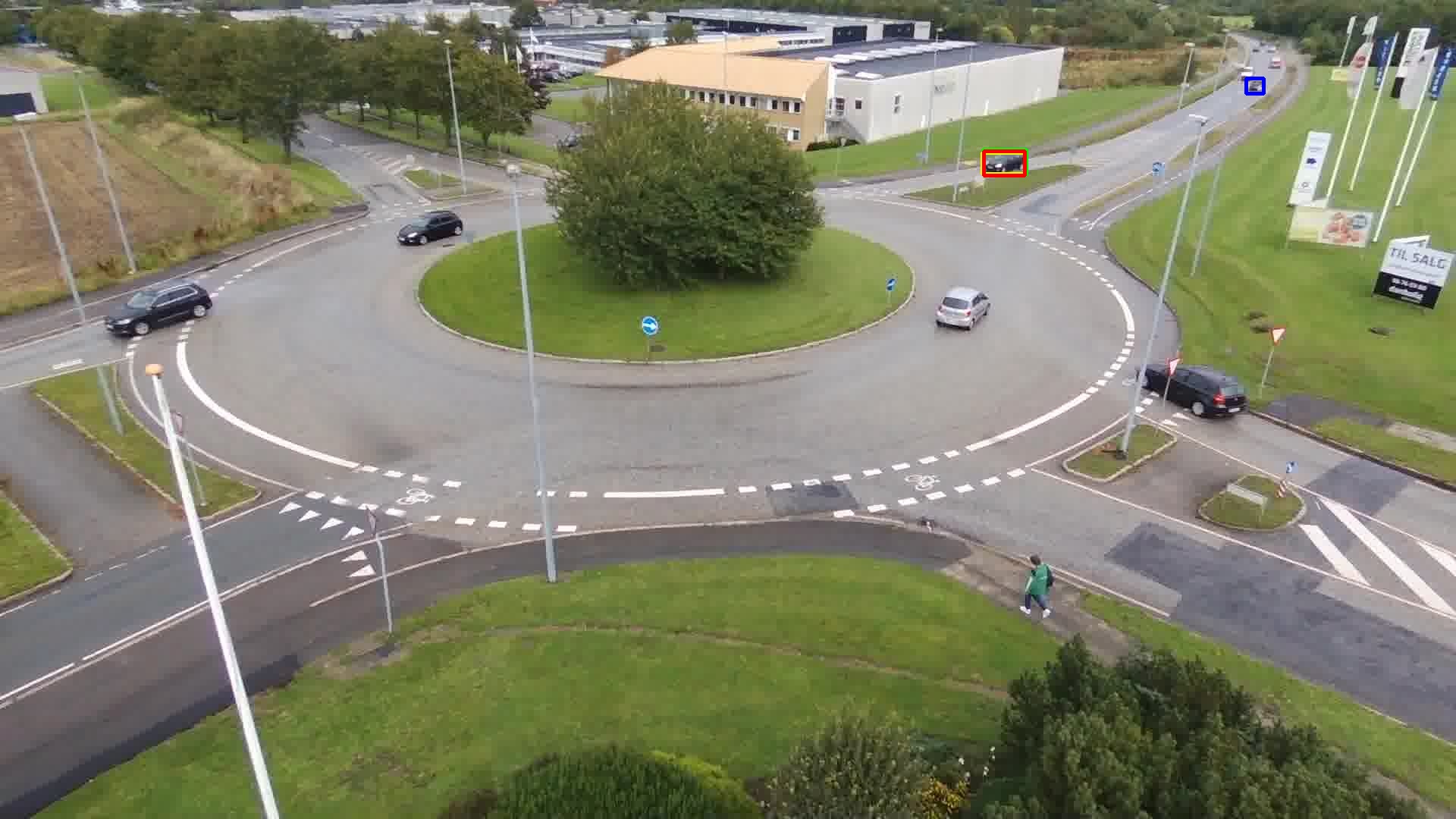}};
            \begin{scope}[x={(image.south east)},y={(image.north west)}]
                \draw[red, thick, dashdotted] (0.62,0.71) rectangle (0.88,0.93);
            \end{scope}
        \end{tikzpicture}
    \end{subfigure}\hspace{0.0005em}
    \begin{subfigure}{0.28\linewidth}
        \begin{tikzpicture}
            \node[anchor=south west,inner sep=0] (image) at (0,0) {\includegraphics[width=\linewidth, bb=0 0 1920 1080]{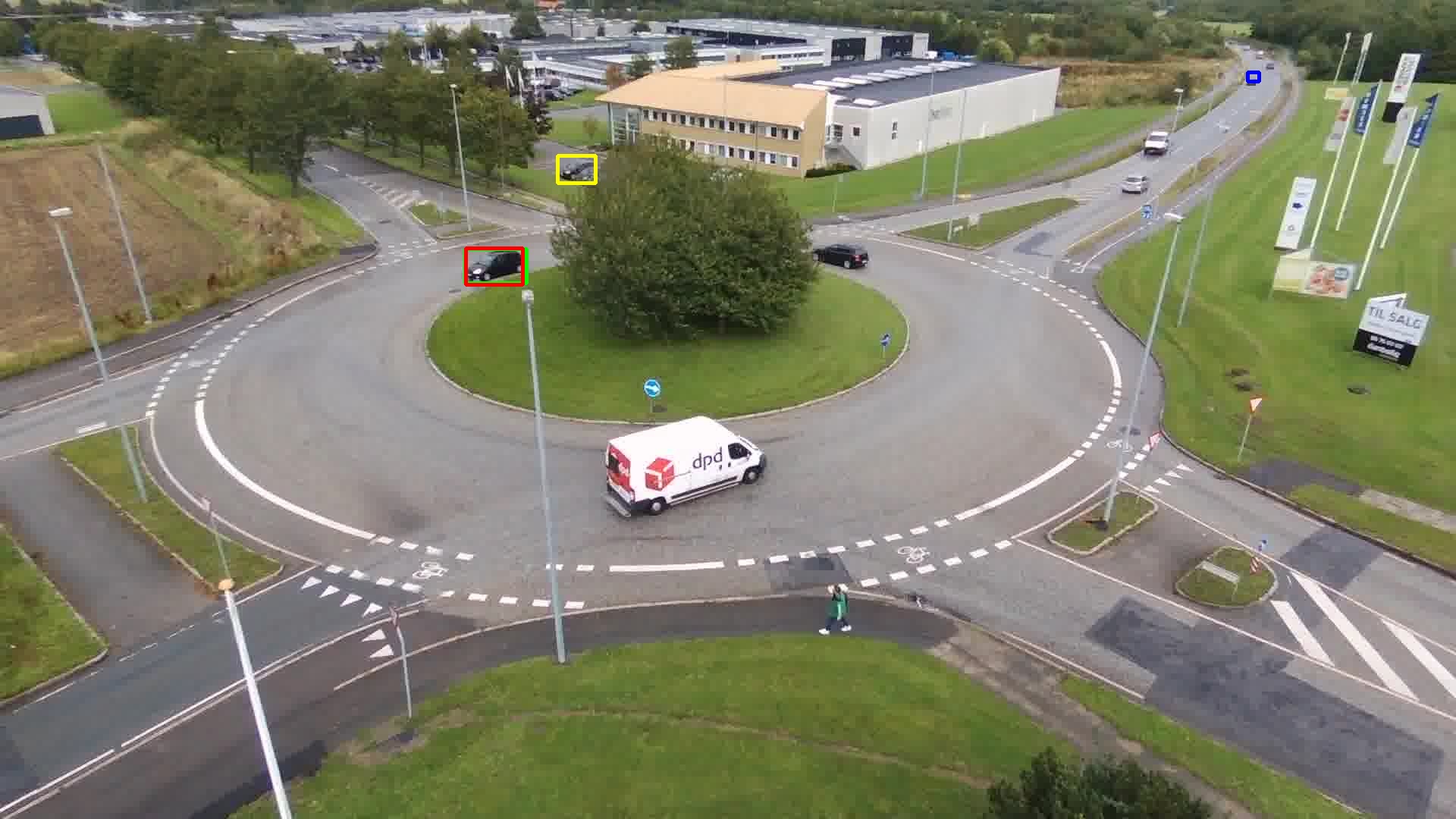}};
            \begin{scope}[x={(image.south east)},y={(image.north west)}]
                \draw[red, thick, dashdotted] (0.29,0.61) rectangle (0.56,0.83);
            \end{scope}
        \end{tikzpicture}
    \end{subfigure}
    \begin{subfigure}{0.28\linewidth}
    \centering
        \includegraphics[width=\linewidth, bb=0 0 1920 1080, trim=50cm 30cm 0cm 0cm, clip=true]{Graphics/introduction/Intro_000.jpg}
    \end{subfigure}\hspace{0.0005em}
    \begin{subfigure}{0.28\linewidth}
        \includegraphics[width=\linewidth, bb=0 0 1920 1080, trim=42cm 27cm 8cm 3cm, clip=true]{Graphics/introduction/Intro_001.jpg}
    \end{subfigure}\hspace{0.0005em}
    \begin{subfigure}{0.28\linewidth}
        \includegraphics[width=\linewidth, bb=0 0 1920 1080, trim=20cm 23.5cm 30cm 6.5cm, clip=true]{Graphics/introduction/Intro_002.jpg}
    \end{subfigure}
    \begin{subfigure}{1\linewidth}
        \centering
        \includegraphics[width=.84\linewidth]{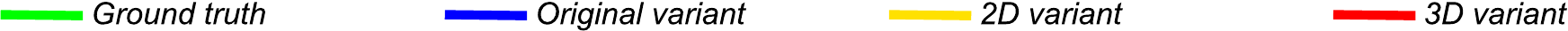}
    \end{subfigure}
    \caption{Qualitative comparison between an original tracking algorithm, its~2D variant, and its 3D variant for UAV onboard visual object tracking. A~closer look of the scenario is shown on the lower part of the figure, based~on a region delimited by a red dash--dot rectangle in the corresponding upper image. Only~the 3D variant is able to overcome the occurring occlusion.}
    \label{fig:Seq_preview}
\end{figure}

\begin{figure}[H]
    \begin{subfigure}{1\linewidth}
        \centering
        \includegraphics[width=.9\linewidth, bb=0 0 1920 1080, trim=1cm 15cm 3cm 10cm, clip=true]{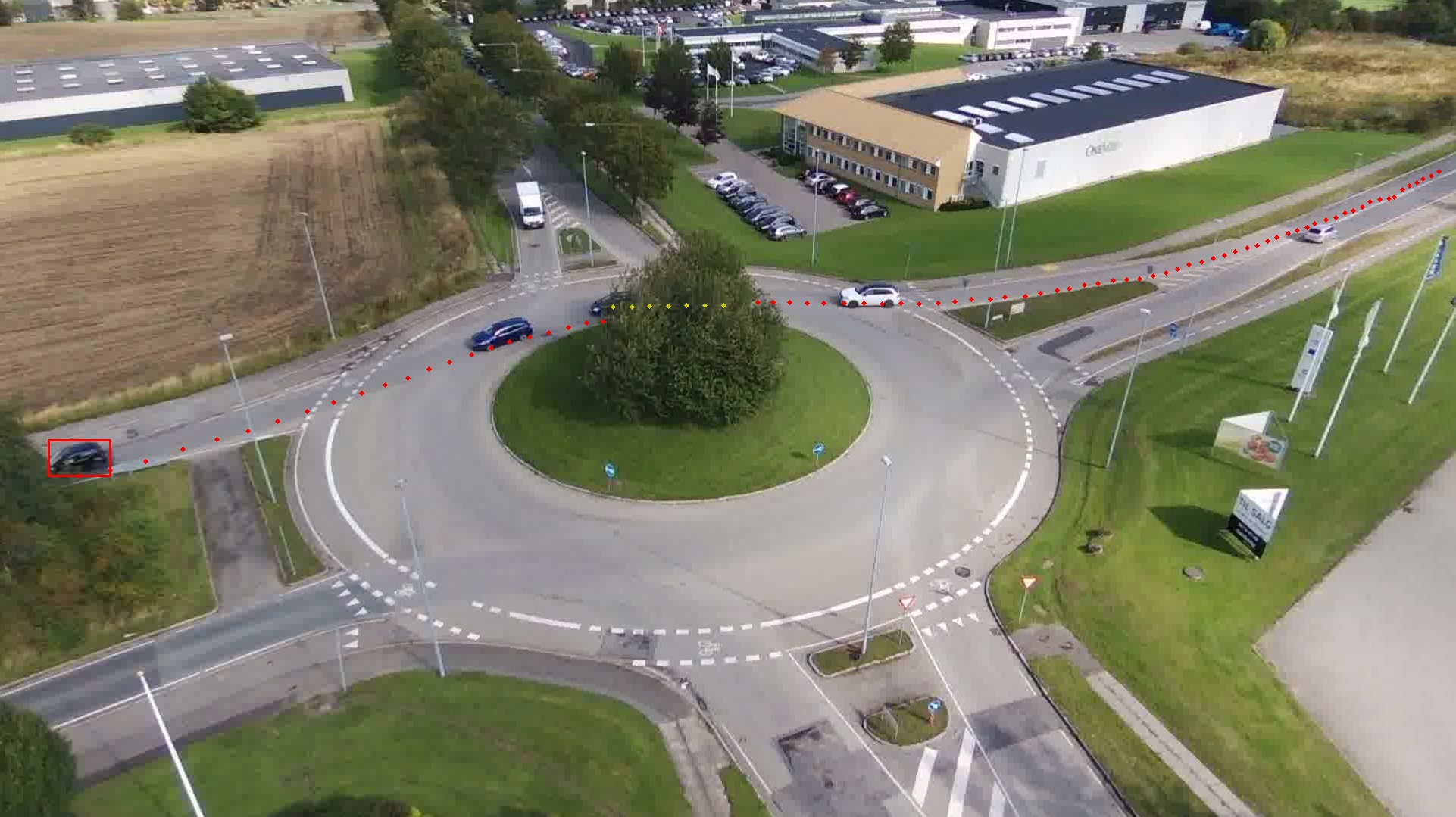}
    \end{subfigure}
    \begin{subfigure}{1\linewidth}
        \centering
        \includegraphics[width=.9\linewidth, bb=0 0 3435 2006, trim=15cm 23cm 17cm 19cm, clip=true]{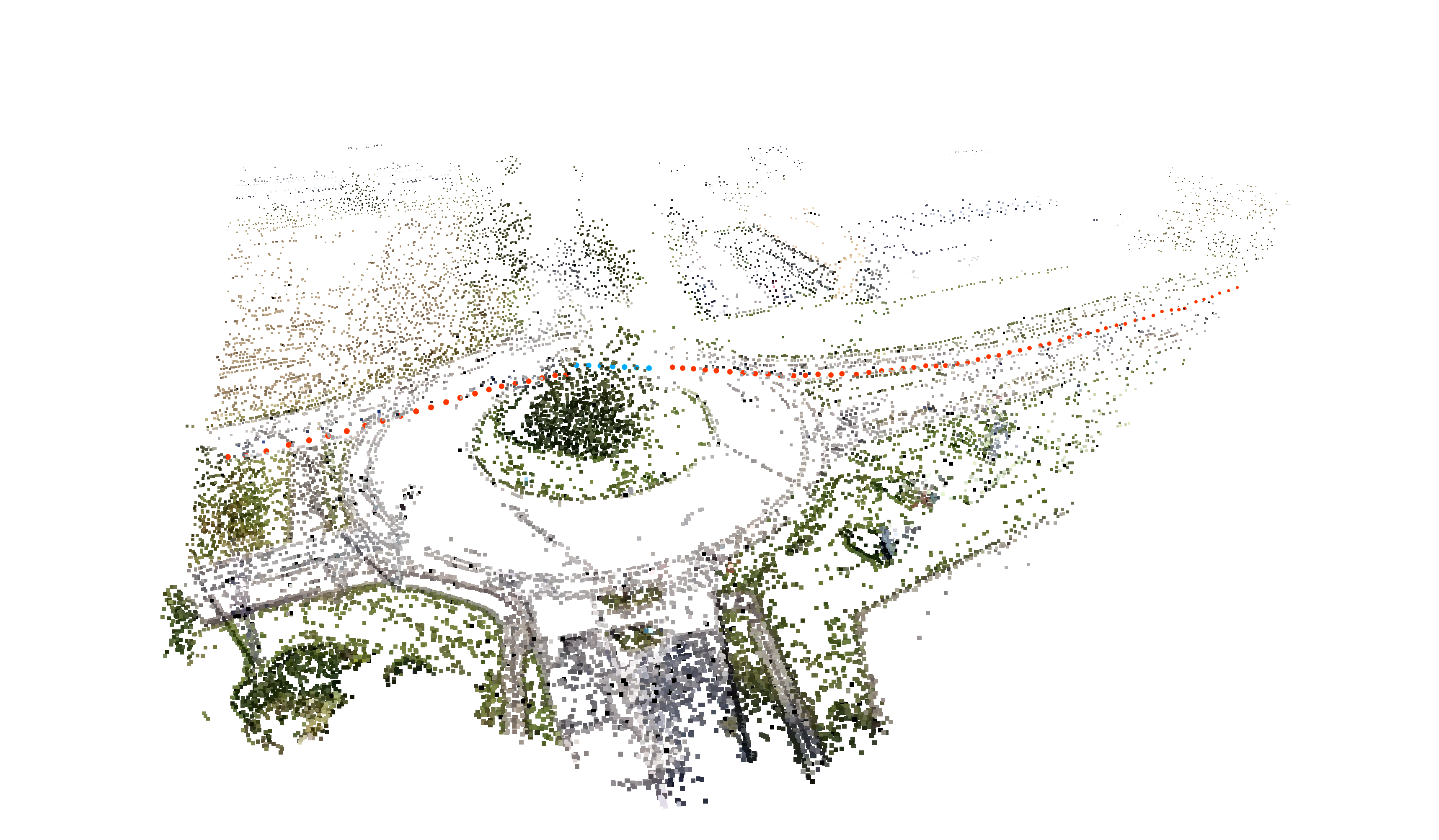}
    \end{subfigure}
    \caption{Qualitative results presenting the trajectory of the object in image space (upper figure) and 3D scene space (lower figure). The~red points in both images indicate the 2D and 3D trajectory extracted during tracking. Yellow~points (2D) and blue points (3D) indicate the occluded trajectory reconstructed by leveraging a state estimator.}
    \label{fig:3D_Traj}
\end{figure}

To better deal with the challenges of low-altitude UAV views, we~propose a {modular} detection-by-tracking pipeline coupled with a 3D reconstruction of the environment. The~core contributions of this work are as follows:
(1) We propose a framework combining three main components. A~visual object tracker, for modeling the appearance model of the object and inferring the position of the object in the image. A~3D reconstruction of the static environment, allowing us to associate pixel positions with a corresponding 3D location. Lastly, a state estimator---i.e., particle filter---for~estimating the position and velocity of the object in the 3D reconstruction.
(2) We show that the incorporation of 3D information into the tracking pipeline has several benefits. A~3D transition model increases the realism of the state estimator predictions, reflecting~the corresponding object dynamics. The~3D camera poses allow to compensate for ego-motions, and the depth information improves the handling of object occlusions~(see \figref{fig:Seq_preview}). The~proposed approach allows us to shift from tracking in 2D image space to tracking in 3D scene space~(see \figref{fig:3D_Traj}).
(3) We improve the processing of false associations---i.e., distractors---through~the usage of a multimodal state estimator.
(4) We create a new dataset called AU-AIR-Track, designed~for visual object tracking from a UAV perspective. The~dataset includes 90 annotated objects as well as annotated occlusions and two 3D reconstructions of the static scenes observed from the UAV.
(5) We demonstrate the effectiveness of our pipeline through quantitative results and qualitative analysis.

The paper is structured as follows. In~\secref{related Work}, an~overview of related work on {single} visual object tracking and their corresponding benchmarks is provided. The~complete system and the individual components are explained in \secref{Method}. {We give a detailed description of the edited dataset used and present metrics with the evaluation protocol for assessing the performance of our approach in} \secref{Evaluation}. Finally, in~\secref{results}, we analyze quantitative as well as qualitative results and present the benefits of the design choices made. In~the end, we~{discuss modifications that can be added to our approach in the \secref{discussion} and give a} conclusion in \secref{conclusion}.

\section{Related Work}
\label{related Work}
In recent years, great~progress regarding {single} visual object tracking has been made, owing~to the abundance of benchmarks available~\cite{bib:vot_2019, bib:otb100, bib:otb50, bib:vot_2018, bib:lasot, bib:alov, bib:oxuva, bib:got-10k, bib:tc128, bib:trackingnet, bib:buaa}. Most~of them are designed towards evaluating tracking algorithms on ground-level perspectives, resulting~in state-of-the-art visual object trackers following the tracking-by-detection paradigm.
Currently, three~tracking designs prevail on those benchmarks: the~discriminative correlation filters~\cite{bib:dcf0, bib:dcf1, bib:dcf2, bib:dcf3, bib:dcf4, bib:dcf5, bib:dcf6}; the~Siamese-based approach~\cite{bib:siam_0, bib:siam_1, bib:siam_2, bib:siam_3, bib:siam_4, bib:siam_5}; and~recently, trackers~inspired by correlation filters that employ a small convolutional neural network for learning the appearance model of the object~\cite{bib:atom, bib:dimp}. In~all three design choices, the~main difference lies in how they learn an appearance model of the object. The~latter style is explored in this paper.

These tracking algorithms are tailored to scenarios presenting ground-level perspectives, but~onboard UAV perspectives present particular challenges. For~instance, the~object occupies a relatively small portion of the image space, resulting~in a less-accurate learned appearance model. This~leads to lower discrimination capabilities when similar objects to the {tracked} object are encountered. In~particular, tracking~an object with an oblique view on the scene from a UAV can present multiple occlusion situations compared to a top-down view.
To analyze the performances of tracking algorithms in UAV scenarios, several~benchmarks have been introduced~\cite{bib:uav123, bib:dtb70, bib:uavdt, bib:visdrone}, ranging~from low to high altitudes, and~propose either an oblique view from the UAV or a top-down view.
Most participating trackers in the {SOT} UAV benchmarks use~state-of-the-art {single} visual object trackers presented in ground-level perspective benchmarks without or only with minor adaptations. However, none~of the adapted trackers participating in the {SOT} UAV benchmarks attempt to utilize 3D information. This~can be explained by the lack of such information being provided in the datasets. In~contrast, the~AU-AIR dataset~\cite{bib:au-air} introduced recently is~oriented toward object detection from a UAV viewpoint. It~offers sequences that capture typical traffic on a roundabout, as~would a surveillance drone for traffic monitoring, and contains sequences that are suitable for reconstructing the observed scene in 3D---sufficient translation movements, not~flying around excessively, and enough structures in the environment.

In contrast to the {SOT} UAV benchmark approaches mentioned previously, there~are different application domains such as autonomous driving, where~objects are tracked in a 3D system of reference through a detection-by-tracking paradigm. Typically,~the 2D image detections generated serve as measurements and are mapped from image space in the ego-motion-compensated reference system of the car. An~example of applications for traffic monitoring is presented in~\cite{bib:3D_0}, where~the authors propose a pipeline including {Multiple Object Tracking (MOT)}, stereo~cues, visual~odometry, optical~scene flow, and a Kalman filter~\cite{welch1995introduction} to enhance tracking performance on the KITTI benchmark~\cite{bib:KITTI_0,bib:KITTI_1}. A~follow up to this study was~\cite{bib:3D_1}, which~reconstructed the static scene and the object in 3D, allowing~the shift from tracking in the 2D image space toward tracking in 3D scene space. In~addition, the~reconstructed object is associated with a velocity, inferred~from the optical flow of the object, which~is afterward associated with tracklets, thus enabling~the authors to tackle occlusion situations and missing detections.

{Most related to our work is the approach presented in~\cite{bib:eye_in_Sky_3D}, where~the authors developed an MOT pipeline for UAV scenarios that also benefits from a 3D scene reconstruction for estimating the object location in 3D. In~contrast to our work, the~authors rely on the tracking-by-detection paradigm by leveraging RetinaNet~\cite{bib:retinaNet} for detecting objects in the image sequence. They~generate tracklets on image-level by integrating visual cues and temporal information to reduce false or missing detections. By~projecting the image-based positions of detected objects on the estimated ground plane---inferred~through visual odometry and multiview-stereo---the~framework is able to assess their 3D positions. However, by~using an object detector, the~authors are only able to track object classes known by the object detector. In~contrast, model-free trackers,~i.e., SOTs, are~able to track arbitrary~objects.}

We are convinced that tracking applications such as {single object} visual tracking from a UAV would~also benefit from a shift towards a detection-by-tracking paradigm by incorporating 3D information. In~this paper, we~apply a model-free single object visual tracker, implying~that the tracker can only track a single object and starts with a blank appearance model---without an offline/pretrained appearance model. Regardless~of the method used for training the tracker---i.e., offline, online---an~appearance model of the object is used to locate the object in the image space. Here, we~consider the state-of-the-art visual trackers ATOM~\cite{bib:atom} and DiMP~\cite{bib:dimp} 
for appearance modeling.

Important for enabling the 2D to 3D mapping is a 3D representation of the observed scene.
{To~this end, a~Structure-from-Motion (SfM) or visual Simultaneous Localization and Mapping (SLAM) approach can be leveraged. SfM~is a photogrammetric technique that estimates the 3D structures of a scene based on a set of images taken from different viewpoints~\cite{bib:colmap0,bib:visualsfm,Moulon2012,snavely2006photo,fuhrmann2014mve,theia-manual}. Visual~SLAM, similarly~to SfM, reconstructs 3D camera poses and scene structures by leveraging specific properties of features in ordered image sets~\cite{bib:orb_slam, bib:orb_slam_2, bib:lsd_slam}. In~addition, UAVs~can improve the robustness of the reconstruction by associating an Inertial Measurement Unit (IMU) with the corresponding SfM or Visual SLAM algorithm~\cite{bib:VI-DSO,bloesch2015robust,li2013high,leutenegger2015keyframe,forster2015imu,usenko2016direct,mur2017visual}. However, in~this work, an established software for SfM called  COLMAP~\cite{bib:colmap0}} is~used to extract camera poses in the scene space and to reconstruct the static scene.

We estimate the state of the object in 3D space by relying on a state estimator,~i.e., a particle filter~\cite{bib:particle_f_tuto}. For~evaluating our approach, the~publicly available dataset AU-AIR~\cite{bib:au-air} is employed. The~dataset is carefully further edited resulting in the AU-AIR-Track dataset, to~best reflect prototypical occlusion situations from a low-altitude oblique UAV perspective.

\section{Single Visual Object Tracking Pipeline for UAV}
\label{Method}
\begin{figure}[H]
    \centering
    \includegraphics[width=.98\linewidth]{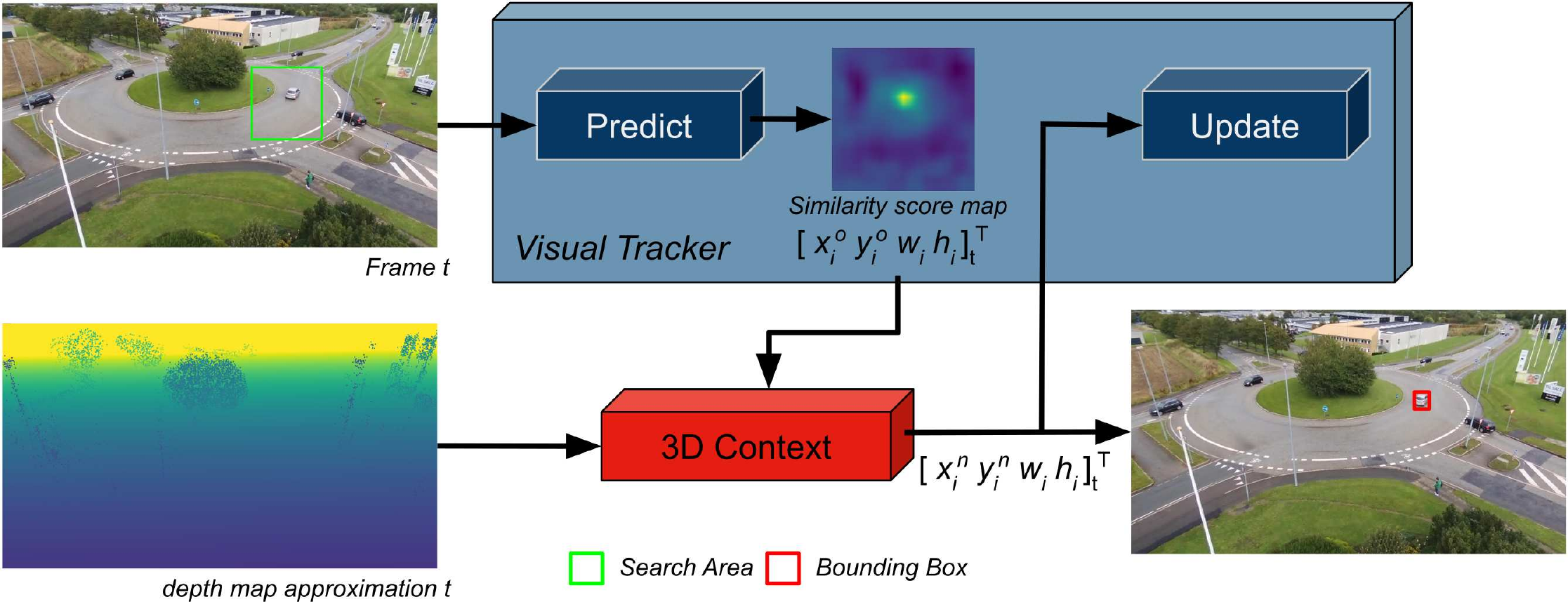}
    \caption{The architecture of the proposed {single} visual object tracking framework. The~visual tracker outputs a similarity score map inferred in the search area. The~similarity score map and depth map approximation corresponding to the image are~processed by the 3D Context component, estimating~a new position for the object in the image.}
    \label{fig:Basic_3D_Tracker}
\end{figure}

The designed framework is intended to be modular, allowing~us to easily substitute different components or add other methods. The~essential architecture of our approach is presented in \figref{fig:Basic_3D_Tracker}.

On an incoming frame, the~visual tracker defines a search area that is based on the previous estimated position and size of the object. The~visual tracker then produces a similarity score map along with estimating an initial position and size of the object~$(x^o_i,y^o_i,w_i,h_i)_t$ in the current frame~$i$ at time step~$t$. The~\emph{3D Context} component estimates the location of the object in the 3D scene space through the similarity score map, the~depth map approximation, and a state estimator---i.e., a particle filter. This~allows the framework to distinguish the object from distractors and also to identify occlusions. Finally, the~3D position estimated by the framework is projected back in the image space as~$(x^n_i,y^n_i)_t$, corresponding to the final estimated position of the object. It~should be noted that no semantic information from the scene---i.e., the road---is~used to facilitate the tracking process.

\secref{Visual_Tracker} describes the visual tracker component. An~overview of the mapping from the 2D image space to the 3D scene space is given in \secref{Sparse_Reconstruction}. The~particle filter is described in \secref{Particle_filter}. Lastly, we~outline the details of our framework in \secref{AAA}.

\subsection{Visual Appearance Modeling of the Object}
\label{Visual_Tracker}
\begin{figure}[H]
    \centering
    \includegraphics[width=.98\linewidth]{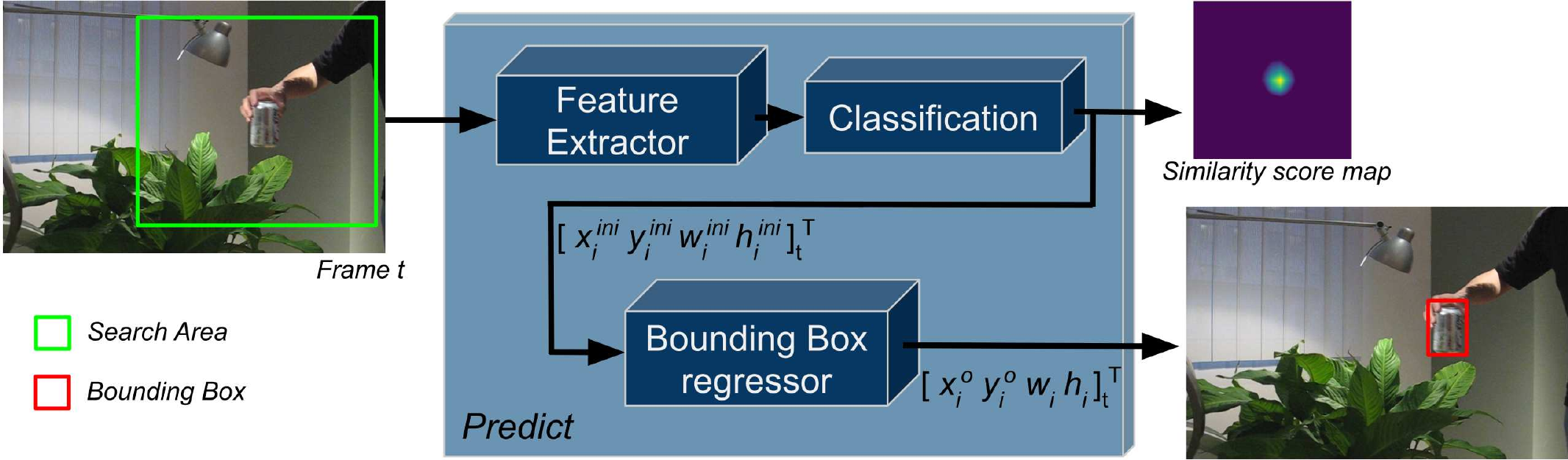}
    \caption{Block diagram of a visual object tracking pipeline---i.e., ATOM and DiMP---without~the update step. During~the tracking cycle, salient~features from the search area are extracted through a feature extractor. Based~on the extracted features, a~similarity score map is inferred, used~in estimating an initial bounding box of the object. Afterward, a~refined bounding box is estimated trough the bounding box regressor, i.e., IoU-Net.}
    \label{fig:Basic_Visual Tracker}
\end{figure}

In this work, we rely on two state-of-the-art visual trackers, ATOM~\cite{bib:atom} and DiMP~\cite{bib:dimp}. The~chosen trackers achieve top ranks against numerous participants on general visual tracking benchmarks~\mbox{\cite{bib:vot_2019, bib:otb100, bib:vot_2018, bib:lasot, bib:got-10k, bib:trackingnet, bib:uav123}}. \figref{fig:Basic_Visual Tracker} shows the main components of a standard pipeline for single object tracking, leveraged by methods such as ATOM and DiMP.

ATOM includes three main components: A {(1) \emph{Feature Extractor}}, a pretrained neural network---e.g., ResNet18, ResNet50~\cite{bib:ResNet}---extracting~salient features. A~{(2) \emph{Classification}} component, composed~of a two-layer convolutional neural network, which~is trained during the tracking process to learn an appearance model of the object. The~{\emph{Classification}} component proposes an initial estimation of the location and size of the object in the image as a bounding box. Lastly, a~{(3) \emph{Bounding Box regressor}} component, based~on the IoU-Net~\cite{bib:iounet} (trained offline), which refines~the initially proposed bounding box.

DiMP is a successor of ATOM and builds on the same elements. The~main difference lies in the extension of the {\emph{Classification}} component. A~new strategy for the initialization of the appearance model is used, expressing~the appearance model with better weights at the start. The~online learning process for updating the appearance model is also refined for faster and more stable convergence.

For each incoming frame, a~search area is created to delimit the possible positions of the object in the frame. The~positioning and  size of the search area depend on the previous estimated position and size of the object. Based~on the features extracted from the search area and the appearance model of the object, a~similarity score map is inferred, reflecting~the resemblance of the extracted features with the learned appearance model. The~highest score in the similarity score map is designated as the position of the object due~to the maximum-likelihood approach. The~object estimation module---i.e.,~IoU-Net---is~used to identify the best fitting bounding box and thus, to~refine the estimated position. To~cope with changes in the appearance of the object---e.g., illumination variations, in-plane rotation, motion~blur, background~clutter---the~appearance model of the object has to be adapted during tracking. This~adaptation is reached by updating the appearance model regularly in ATOM and DiMP. The~updates occur every 10 valid frames for ATOM and every 15 valid frames for DiMP. A~valid frame corresponding to a frame where the object has been identified correctly---with a high similarity score. The~new appearance model is adapted by retraining the classification component with the search areas of those valid frames. The~tracker deals with distractors by recognizing multiple peaks in the similarity score map and immediately updating the appearance model of the object with a high learning rate.

\subsection{Point Cloud Reconstruction of the Environment}
\label{Sparse_Reconstruction}
We employ Structure-from-Motion (SfM) for~achieving the mapping between the 2D image space and the 3D reference system,~i.e., scene space. {Owing to our modular design}, the~mapping between the 2D image space and the 3D scene space can be replaced with an alternative {photogrammetric technique}. For~this paper, we~create {a point cloud representation} of the observed environment, which~is sufficient for the demonstration of our approach.

{We leverage COLMAP ({{in our experiments we used COLMAP 3.6, available publicly under~\url{https://colmap.github.io/} (accessed on 14.10.2020)}})~\cite{bib:colmap0,bib:colmap2}, 
which~is an established approach for SfM. For~a given image set that contains overlapping images of the same scene and taken from different viewpoints, COLMAP~automatically extracts the corresponding camera poses and reconstructs the scene structures in 3D. To~achieve this, the~library follows common SfM steps. (1)~A~correspondence search, where~salient features from an image set are extracted and matched~across the images, and incorrectly matched features are filtered out. (2) The scene is reconstructed as a point cloud by performing image registration, feature~triangulation, and bundle adjustments.}

\begin{figure}[H]
\centering
    \begin{subfigure}{.3\linewidth}
        \includegraphics[width=1\linewidth, bb=0 0 1834 1018]{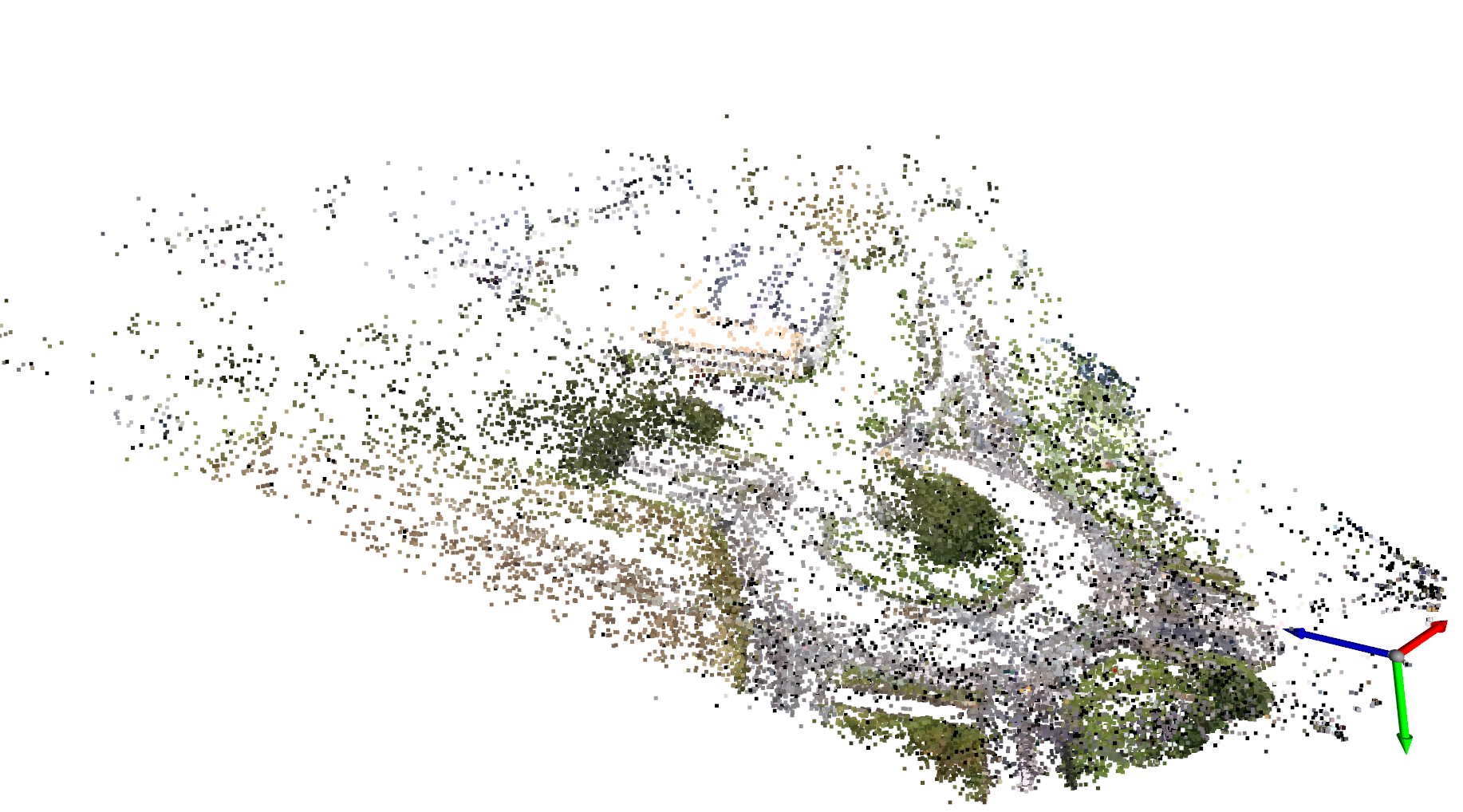}
        \caption{}
        \label{fig:filter_pnt_cld_a}
    \end{subfigure}
    \begin{subfigure}{.3\linewidth}
        \includegraphics[width=1\linewidth, bb=0 0 1874 960]{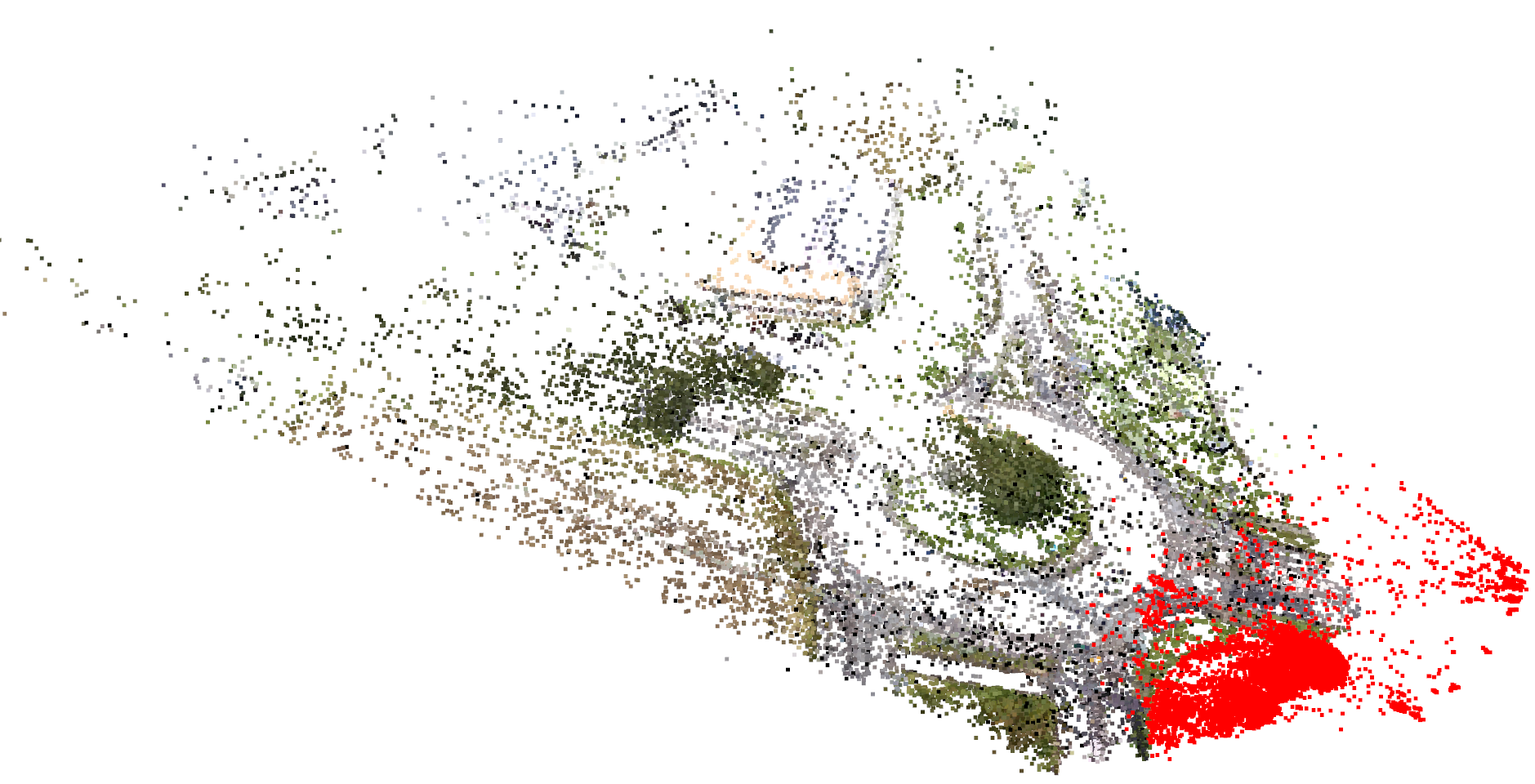}
        \caption{}
        \label{fig:filter_pnt_cld_b}
    \end{subfigure}
    \begin{subfigure}{.3\linewidth}
        \includegraphics[width=1\linewidth, bb=0 0 1900 985]{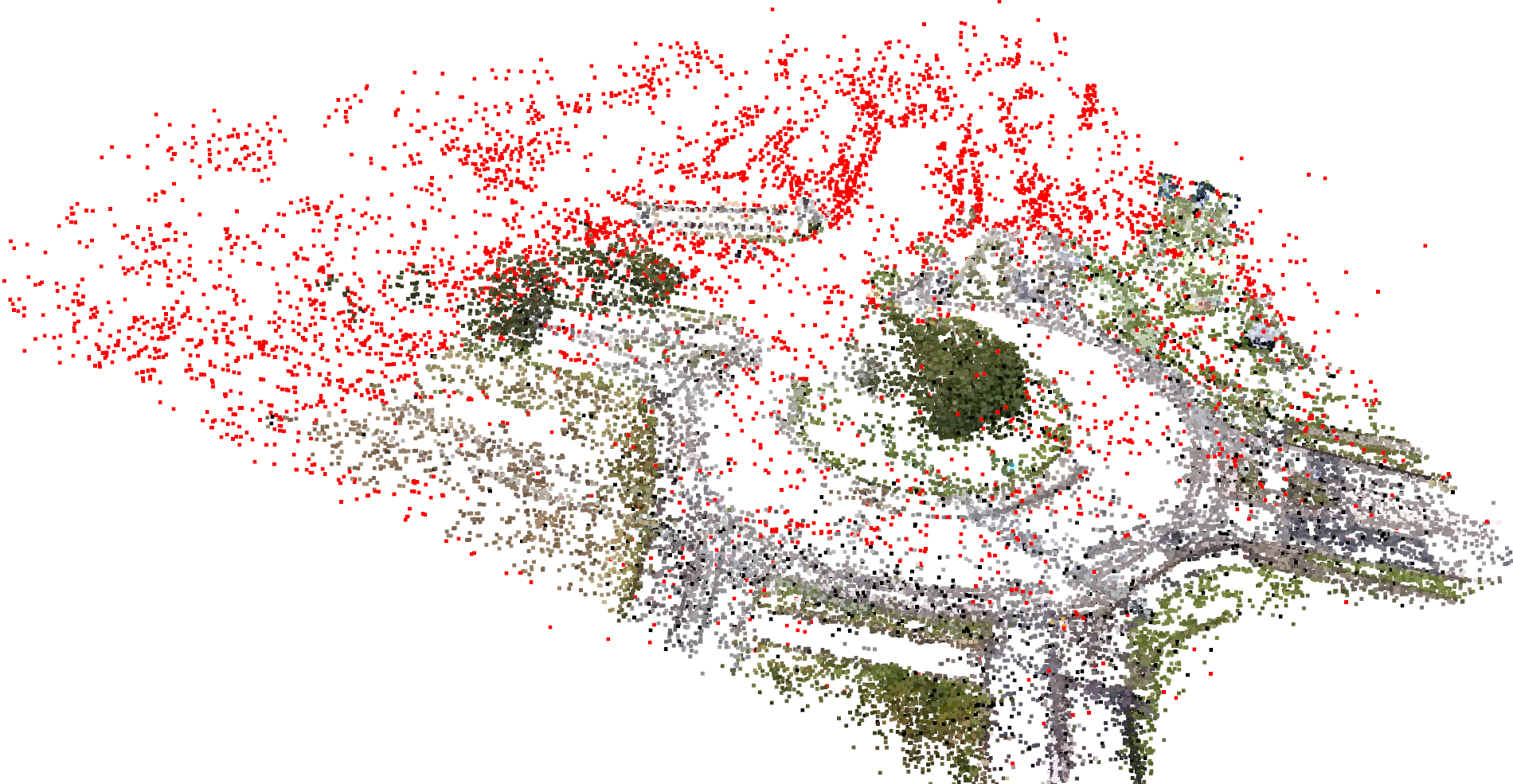}
        \caption{}
        \label{fig:filter_pnt_cld_c}
    \end{subfigure}
    \caption{(\textbf{a}) A point cloud reconstruction using COLMAP. (\textbf{b}) Highlighted noise near the frame of reference in red. (\textbf{c}) Highlighted outliers in red after removing near-frame reference noise.}
    \label{fig:filter_pnt_cld}
\end{figure}

\begin{figure}[H]
    \centering
    \includegraphics[width=0.5\linewidth, bb=100 40 1600 650]{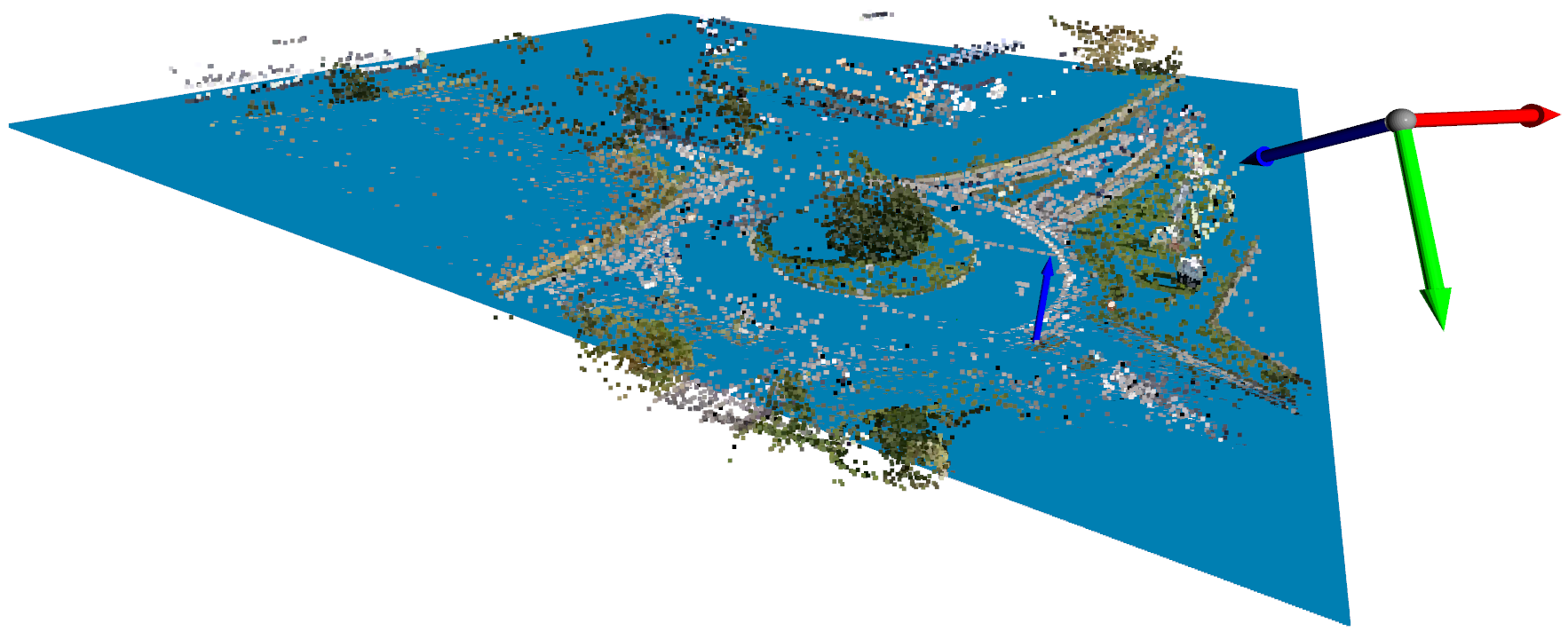}
    \caption{A filtered {point cloud} with the corresponding ground plane~$(\veccsdos{x}{g},\veccsdos{y}{g})$ in blue.}
    \label{fig:plane_est}
\end{figure}

\figref{fig:filter_pnt_cld}a presents a {point cloud reconstruction} leveraged by COLMAP on an image sequence of AU-AIR-Track. The~initial reconstruction of the scene contains noise~near the camera frame and statistical outliers. Points~near the camera frame are falsely triangulated and outliers are points correctly positioned but that are too sparse for reliably representing structures of the scene. To~reduce the number of incorrect points in the reconstruction, we~filter out points close to the camera frame and statistical outliers.
{We discard near-camera points from the point cloud by comparing their Euclidean distance with a threshold.
For statistical outliers, we~proceed as follows: (1) We define a neighborhood of 10 neighbors. The~average distance~$d_i$ of a given point~$i$ to its neighbors is calculated using the Euclidean distance. (2) A standard deviation threshold~$\sigma_{lim}$ is defined and the overall average for all~$d_i$ is computed as~$d_{avg}$. Points~with an average distance~$d_i \notin [d_{avg} - \sigma_{lim}, d_{avg} + \sigma_{lim}]$ are identified as outliers and discarded from the point cloud.
}\figref{fig:filter_pnt_cld}b,c display the removal of points close to the camera frame and statistical outliers. Moreover, a~plane ground~$(\veccsdos{x}{g},\veccsdos{y}{g})$ is estimated by using the Random sample consensus (RANSAC)~\cite{bib:ransac} method (see \figref{fig:plane_est}). Points~below the ground plane~$(\veccsdos{x}{g},\veccsdos{y}{g})$ are discarded.

\begin{figure}[H]
    \centering
    \begin{subfigure}{.45\linewidth}
        \includegraphics[width=1\linewidth]{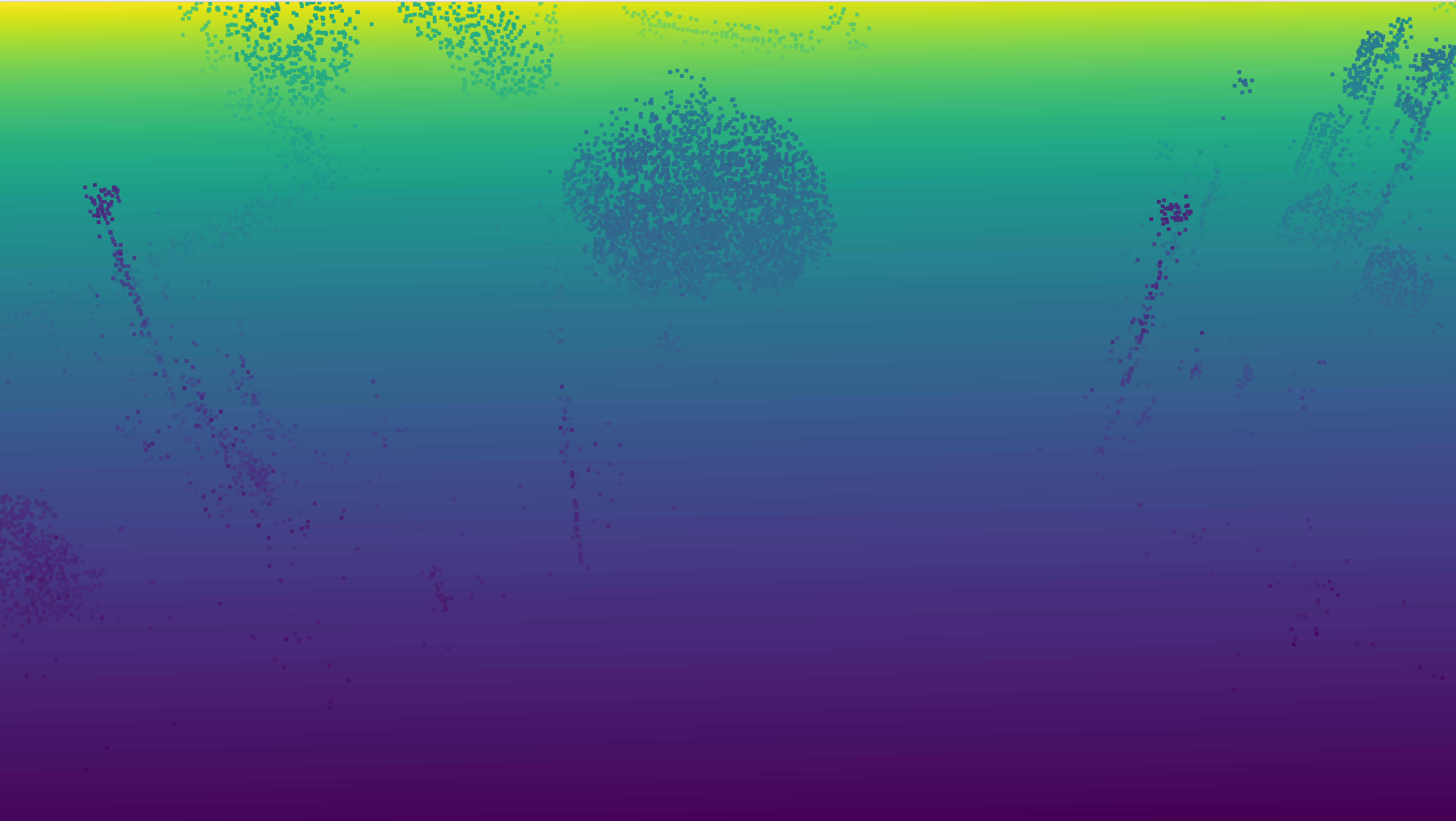}
    \end{subfigure}
    \begin{subfigure}{.45\linewidth}
        \includegraphics[width=1\linewidth]{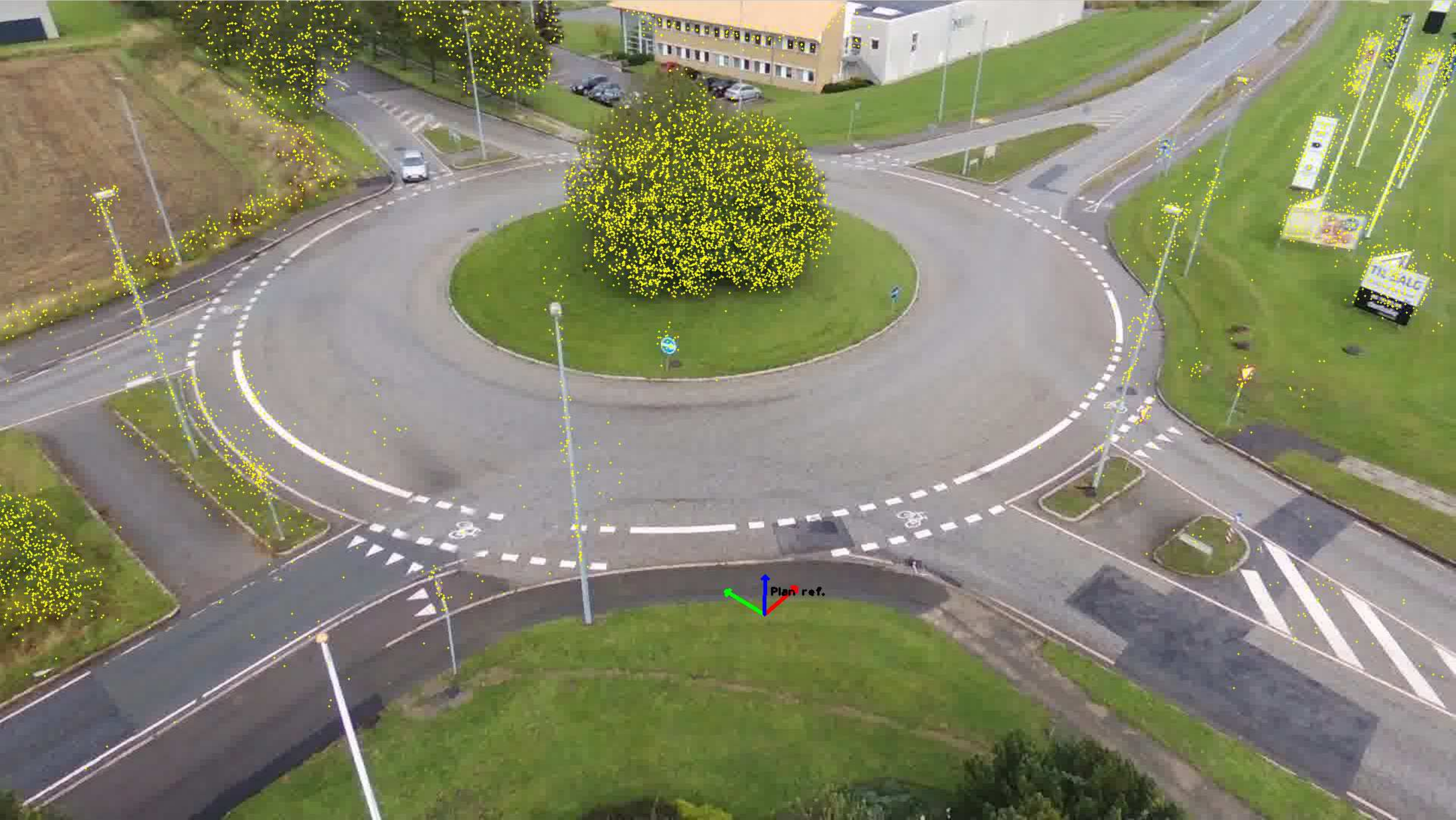}
    \end{subfigure}
    \caption{Comparison between a depth map approximation (\textbf{left}) and the corresponding frame (\textbf{right}). Points~in yellow on the frame indicate visible points from the {point cloud} reconstruction aside from the ground estimation.}
    \label{fig:depth_map}
\end{figure}

We compute depth map approximations based on the new scene reconstructed,~i.e., the filtered {point cloud} with the estimated ground plane. The~depth map approximations encode, for~every pixel, the~distance between the positions of the UAV and the visible points in the scene for the corresponding viewpoint,~i.e., camera pose. For~every frame in the AU-AIR-Track dataset, a~corresponding depth map approximation is constructed {by relying on the pinhole camera model}; thus, enabling~the mapping between the 2D image space and the 3D scene space. \figref{fig:depth_map} displays a depth map approximation example with the corresponding frame.

\subsection{Particle Filter for Modeling the Object in 3D}
\label{Particle_filter}
{In order to overcome the drawbacks of the common maximum-likelihood approach used in SOTs, we~switch to detection-by-tracking by applying an additional state estimator. Additionally, to~robustly handle multiple high similarity scores in the similarity score map, a~state estimator with a multimodal representation of the probability density function is favored. To~that end, a~particle filter~\cite{bib:particle_f_tuto} is used, which~will estimate the state of the object---i.e., the position and velocity---in~the 3D scene for every time step~$t$}. The~particle filter estimates the posterior density function of the object~$p(s_t)$ based on a transition model~$f$, the~current observation~$z_t$, and the prior density function~$p(s_{t-1})$. The~general idea is shown in \meqref{eq:Pf_1}:

\begin{equation}
    p(s_t) = f(p(s_{t-1}), z_t)
    \label{eq:Pf_1}.
\end{equation}

In order to approximate the probability density function~$p(s_t)$, the~particle filter uses particles. Each~particle denotes a hypothesis on the state. Particles~from the prior probability density function~$p(s_{t-1})$ are propagated through the transition model~$f$. Weights~$w_t^{i}$ at time step~$t$ are assigned to~$n$ particles with~$i \in I = \{1,...,n\}$, mirroring~how strongly particles match with the current observation~$z_t$. Let~$p(s_t|z_{1:t})$ represent the probability density function of the posterior state given all observations~$z_{1:t}$ up to time step~$t$ in \meqref{eq:Pf_2}. Each~particle has a corresponding weight~$w_t^{i}$ at time step~$t$. Let~$s_t^{i}$ denote the hypothesis on the state of the~$i$-th particle and~$s_t$ the state estimation at time~$t$. $\delta$ is the Dirac delta function. The~weights are normalized such that~$\sum_{i}^{n} w_t^{i} = 1$. Particles~are weighted according to their matching similarity with the current observation~$z_t$.

\begin{equation}
    p(s_t|z_{1:t}) \approx \displaystyle\sum_{i\in{I}}^{} w_t^{i}\delta(s_t - s_t^{i}).
    \label{eq:Pf_2}
\end{equation}

A resampling is performed after the weighting of the particles whenever~$\hat{n}_{eff}$, presented~in \meqref{eq:Pf_3}, is~below a certain threshold. The~resampling allows the particle filter to discard low-weighted particles and create new particles based on the stronger-weighted ones, allowing~a refined approximation of~$p(s_t|z_{1:t})$.

\begin{equation}
    \hat{n}_{eff} = \frac{1}{\displaystyle\sum_{i\in{I}}^{}{(w_t^{i})}^2}.
    \label{eq:Pf_3}
\end{equation}

In our case, the~observation~$z_t$ is the similarity score map produced by ATOM or DiMP. We~apply a constant velocity model for the transition model~$f$ and fix the velocity~$v_{\veccsdos{z}{g}}$ on the~$\veccsdos{z}{g}$ axis, as~$v_{\veccsdos{z}{g}}~=~0$---where~$\veccsdos{z}{g}$ is perpendicular to the estimated ground plane~$(\veccsdos{x}{g},\veccsdos{y}{g})$ of the reconstructed scene. This~results in particles being only able to move on the ground.


\subsection{Tracking Cycle and Occlusion Handling}
\label{AAA}
During initialization, the~visual tracker learns an appearance model of the object based on the initial bounding box. At~the same time, the~position of the object in the image is projected onto the estimated ground plane of the 3D reconstruction,~i.e., the scene space. {
To this end, we~rely on the pinhole camera model for estimating the depth value of the object in the scene space. Supposing~that the object does not leave the ground in the real world, we~can assume that the object is bound to only move on the plane ground~$(\veccsdos{x}{g},\veccsdos{y}{g})$. Let~$\veccsdos{n}{c} = (a_c,b_c,c_c)$ describe the normal vector of the plane ground and~$\veccstres{t}{t}{c} = (x^o_c,y^o_c,z^o_c)_t$ describe the position of the object on the plane---expressed in camera frame coordinates. In~\meqref{eq:standard_plane_eq}, we describe the plane in standard form for a generic point on the plane. In~\meqref{eq:homo}, we define~$x^o_c$ and~$y^o_c$, where~$(x^o_i,y^o_i)_t$ are the image coordinates of the object inferred by the visual tracker and~$f$ is the focal length of the camera. By~replacing~$x^o_c$ and~$y^o_c$ in \meqref{eq:standard_plane_eq} and simplifying it, we~obtain \meqref{eq:final_eq}, allowing~us to infer the depth value for the object based on its image coordinates~$(x^o_i,y^o_i)_t$. Now~that the depth value~$z_c$ is estimated, we can determine the missing coordinates~$x^o_c$ and~$y^o_c$ of~$\veccstres{t}{t}{c}$ based on \meqref{eq:homo}. By~applying a rigid body transformation~$\mathbf{H}^{c}_{w}$ from the camera to the world frame of reference, we~obtain the position of the object in the scene space as~$\veccstres{t}{t}{w}= (x^o_w,y^o_w,z^o_w)_t$.

\begin{equation}
    a_c x^o_c + b_c y^o_c + c_c z^o_c - d_c = 0,
    \label{eq:standard_plane_eq}
\end{equation}

\begin{equation}
    x^o_c = \frac{x^o_i z^o_c}{f} \quad\text{and}\quad y^o_c = \frac{y^o_i z^o_c}{f},
    \label{eq:homo}
\end{equation}

\begin{equation}
    z^o_c = \frac{f  dc}{a_c  x^o_i + b_c  y^o_i + c_c  f}.
    \label{eq:final_eq}
\end{equation}

}

The projected position~$\veccstres{t}{t}{w}$ for~the first frame $t=1$, expressed~in 3D coordinates, is considered as the initial position for the state~$s_{-1}$. In~the following frame, the~\emph{visual tracker} component computes a bounding box delimiting the position and size of the object in the frame. Additionally, we~extract the current search area and the similarity score map. Particles~are generated uniformly onto the ground of the 3D reconstruction but are delimited on the projected surface of the search area. Particles~are then weighted accordingly to the similarity score map of the~\emph{visual tracker} component. Following~is a resampling step, which~shrinks the possible locations where the object might be located by regenerating new particles where previous high-weighted particles were located. In~consequence, particles~are mostly located around a high similarity response, giving~us an estimated position of the object in 3D. The~current 3D position~$(x_w,y_w,z_w)_t$ along with the previous 3D position of state~$s_{-1}$ are used to determine an initial velocity. As~a result, an~initial state~$s_0$ for the object is estimated with velocity and position.

In the following frame, the~particle filter can be used to predict the position of the object in the scene space as~$(x^p_w,y^p_w,z^p_w)_t$.
To deal with the uncertainties in the transition model---i.e., the constant velocity model---a~Gaussian noise term is added along \veccsdos{x}{g} and \veccsdos{y}{g}.

\begin{figure}[H]
    \centering
    \includegraphics[width=.96\linewidth]{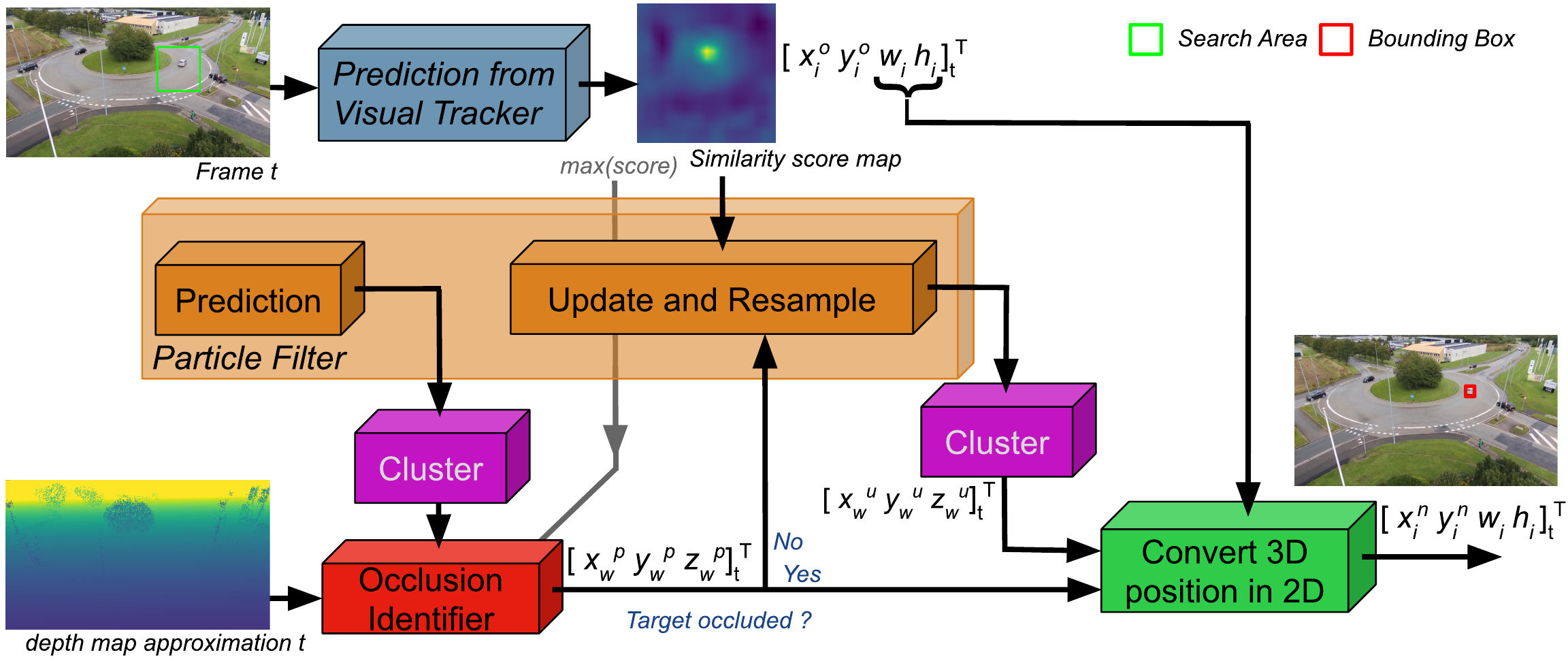}
    \caption{Detailed view of the architecture with the \textit{Context 3D} component. The~prediction of the particle filter and the depth map approximation are used to identify occluded particles. Particles~are clustered depending on their position in the image to identify potential groups, such~as the object and distractors. In~case the object is considered as occluded, the~3D coordinate of the prediction provided by the particle filter~$(x^p_w,y^p_w,z^p_w)_t$ is used as the presumed position of the object in the scene space and is reprojected in the image as~$(x^n_i,y^n_i)_t$. In~case the object is not identified as occluded, an~update and resampling are performed on the particles before clustering them for a second time. Following~is the determination of the cluster representing the object. Lastly, the~3D coordinates~$(x^u_w,y^u_w,z^u_w)_t$ of the cluster, modeling~the object in the scene space, are~reprojected in the image as~$(x^n_i,y^n_i)_t$.}
    \label{fig:3D_Context_helper}
\end{figure}

After initialization, our~tracking framework enters an online tracking cycle. \figref{fig:Basic_3D_Tracker} presents the essential architecture of the framework during online tracking. On~an incoming frame, the~\emph{visual tracker} component defines a search area and produces a similarity score map along with estimating a bounding box for the object on frame~$t$. During~the \emph{3D Context} computation step, we~estimate the 3D position of the object in the scene space, distinguish~the object from distractors, and recognize occlusions. The~new 3D position of the object~$(x_w,y_w,z_w)_t$ expressed in the scene space is projected back onto the frame~$t$, corresponding to the new estimated position~$(x^n_i,y^n_i)_t$ of the object in image space. \figref{fig:3D_Context_helper} displays the different building blocks of the \emph{3D Context} component, which~contains the particle filter that estimates the object state,~i.e., the 3D position and 3D velocity in the scene space. In~the first step, the~particle filter predicts particles on an incoming frame. The~predicted particles are clustered in the image space. We~then identify the cluster representing the object based~on how close each cluster is to the previously estimated position of the object in the scene space.

The \emph{occlusion identifier} component identifies occluded particles composing the object cluster through the depth map approximation. To~this end, we~compare the depth value of the predicted particles from the perspective of the UAV and the depth value from the depth map approximation. We~identify particles hidden by structures if there is a discrepancy between the predicted depth and the observed depth. We~considered that if 50\% of the particles from the object cluster are occluded, then~the object is also occluded. Similarly, the~object is automatically considered as hidden if the similarity score map is flat and widely spread over the search area. Should~the object be identified as occluded, the~tracking framework would rely on the predictions provided by the particle filter~$(x^p_w,y^p_w,z^p_w)_t$.

To identify the reappearance of the object, a~high similarity score close to the expected position is required. When~more than 50\% of the predicted particles are not labeled as occluded, we~consider that the object is potentially visible again. A~specific threshold for the similarity score map is set to consider the object as being truly visible again. A~similarity score greater-than or equal-to this threshold must be reached to consider updating the particle filter with the observation,~i.e., the similarity score map. If~distractors are present when the object reappears, then~the group of particles closest to the prediction is considered to represent the object.

\begin{figure}[H]

    \begin{subfigure}{0.7\linewidth}
        \centering
        \includegraphics[width=.9\linewidth, bb=0 0 1838 1032]{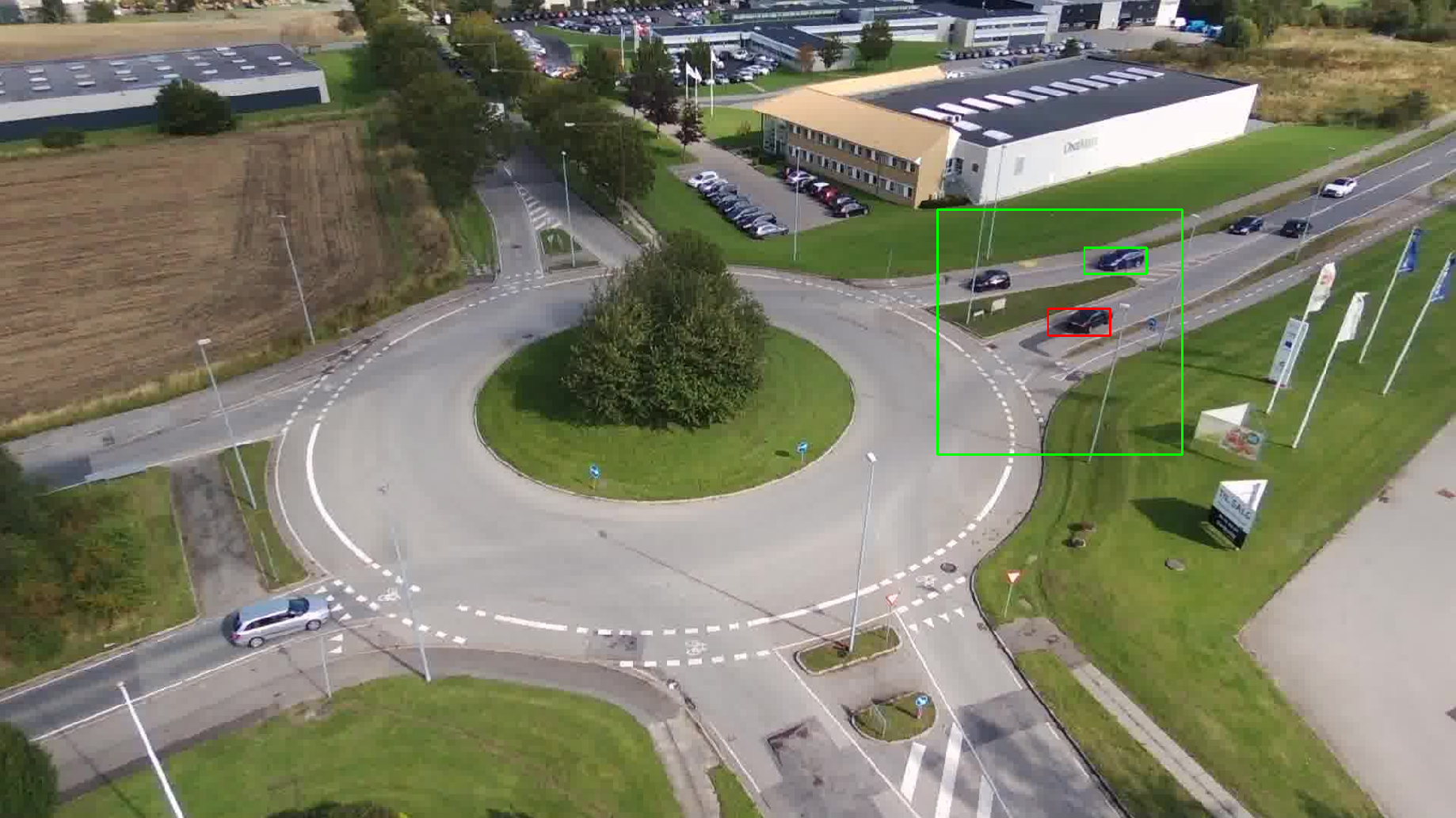}
    \end{subfigure}
    \begin{subfigure}{0.3\linewidth}
        \begin{subfigure}{1\linewidth}
            \includegraphics[width=0.48\linewidth, bb=0 0 250 250]{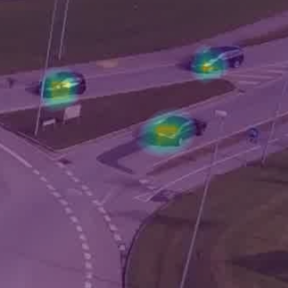}
        \end{subfigure}
        \begin{subfigure}{1\linewidth}
            \includegraphics[width=0.75\linewidth, bb=0 0 800 800]{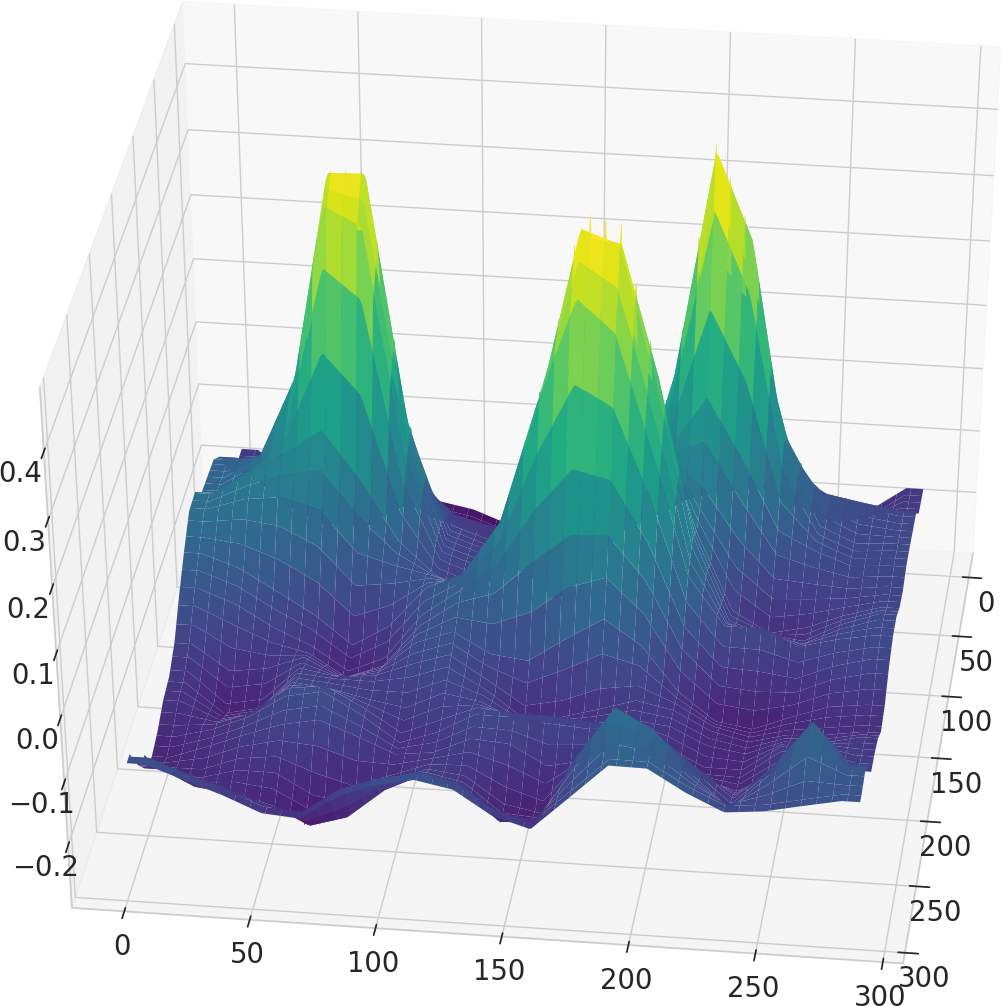}
        \end{subfigure}
    \end{subfigure}
    \caption{The small green bounding box represents the estimated bounding box of the unmodified visual object tracker and the red bounding box represents the estimated bounding box of a visual object tracker coupled with a particle filter. The~large green bounding box corresponds to the search area of the visual tracker. In~this scenario, the~similarity score map has three peaks. Due~to the maximum-likelihood approach of the visual tracker, it~mistakes a distractor with the object; whereas~the visual tracker coupled with a particle filter manages to stay on the object, even~though the highest similarity score is attached to a distractor.}
    \label{fig:Distractor jumps}
\end{figure}

When the object is visible, a~second step is to update the belief of the particle filter with the observation,~i.e., similarity score map. However,~before updating a small percentage,~i.e., 10\%, of~particles are uniformly redistributed across the projected search area on the ground. This~redistribution ensures that we maintain a multimodal distribution. Without~redistributing a portion of the particles, particles~would clump around the object and only a small portion of the similarity score map would be considered for updating the weights. The~weights of the particles are updated by projecting their 3D position in the image space, allowing~us to weigh them accordingly to their 2D location in the similarity score map. We~then resample particles through stratified resampling~\cite{Resampling}. By~using a particle filter, we~can model multimodal tracking, preventing~instantaneous switching from object to distractors as illustrated in \figref{fig:Distractor jumps}.

After particles are resampled, the~framework clusters them based on their position in the image space. We~identify the cluster describing the object, by~comparing the position~$(x^c_w,y^c_w,z^c_w)$ for every cluster~$c$ against the predicted position of the object~$(x^p_w,y^p_w,z^p_w)_t$. The~closest cluster to the predicted position is considered to describe the object state. Thus, the~new/updated position of the object is based on this identified cluster as~$(x^u_w,y^u_w,z^u_w)_t$. The~coordinates in the scene space are then projected in the image space as~$(x^n_i,y^n_i)_t$ and used for updating the appearance model of the~\emph{visual tracker} component. This~avoids adding incorrect training samples, i.e., distractors.

\section{Dataset and Evaluation Metrics}
\label{Evaluation}
{In \secref{AU-AIR-Track}, we~present our edited AU-AIR-Track dataset. In~\secref{Evalu_met}, we~define the metrics used for evaluating the trackers on the dataset.}

\subsection{AU-AIR-Track Dataset}
\label{AU-AIR-Track}
Using our approach, we~want to tackle occlusion occurrences, false~associations, {and ego-motion} using a 3D reconstruction of the static scene. To~that end, we~need a UAV dataset that provides visual object tracking annotations with 3D reconstructions of the scene. As~stated before, current~UAV datasets~\cite{bib:uav123, bib:dtb70, bib:uavdt, bib:visdrone} with visual tracking annotations do not provide 3D information. Therefore,~we created the AU-AIR-Track dataset ({AU-AIR-Track alongside AU-AIR~\cite{bib:au-air} are available under~\url{https://github.com/bozcani/auairdataset} (accessed on 14.10.2020)})  
which includes the following: bounding box annotations with identification numbers, occlusion~annotations, 3D reconstructions of the scene with the corresponding depth map approximations, and camera poses.

AU-AIR-Track is distilled from the AU-AIR dataset~\cite{bib:au-air}, which~provides real-world sequences suitable for traffic surveillance and reflects prototypical outdoor situations captured from a UAV. AU-AIR contains sequences taken from a low flight altitude ranging from 10 to 30 m, under~different camera angles ranging from approximately 45 to 90 degrees. For~each frame, the~dataset provides recording time stamps, Global Positioning System (GPS)~coordinates, altitude~information, IMU data, and~the velocity of the UAV. A~criterion in favor of AU-AIR is~the low range altitude flights with the oblique point of view towards the scene it provides, the~multiple occlusions offered by the tree in the roundabout and the duration of a scene is observed.

\begin{figure}[H]
    \centering
        \begin{subfigure}{0.32\linewidth}
            \includegraphics[width=\linewidth, bb=0 0 1920 1080]{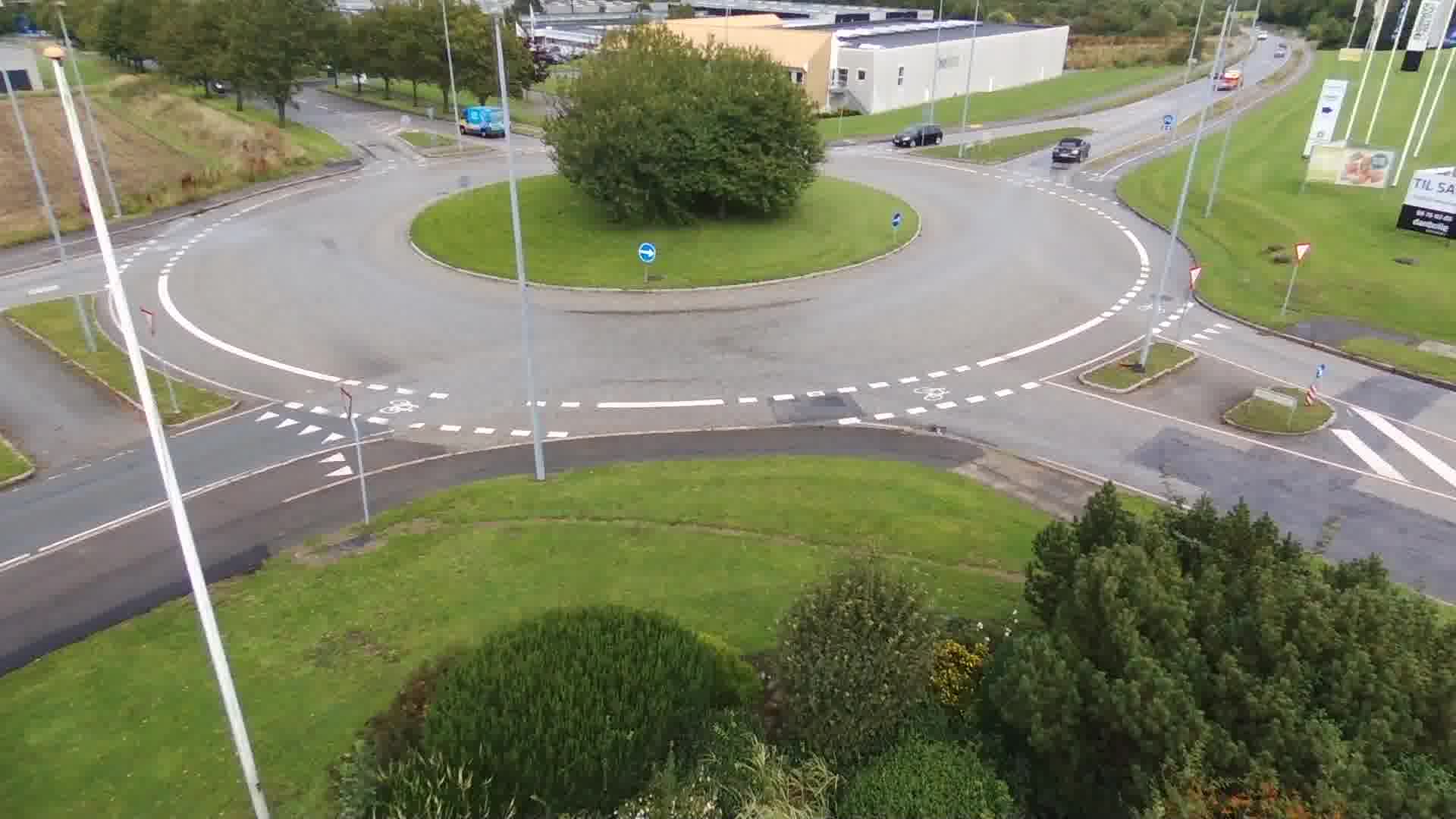}
        \end{subfigure}
        \begin{subfigure}{0.32\linewidth}
            \includegraphics[width=\linewidth, bb=0 0 1920 1080]{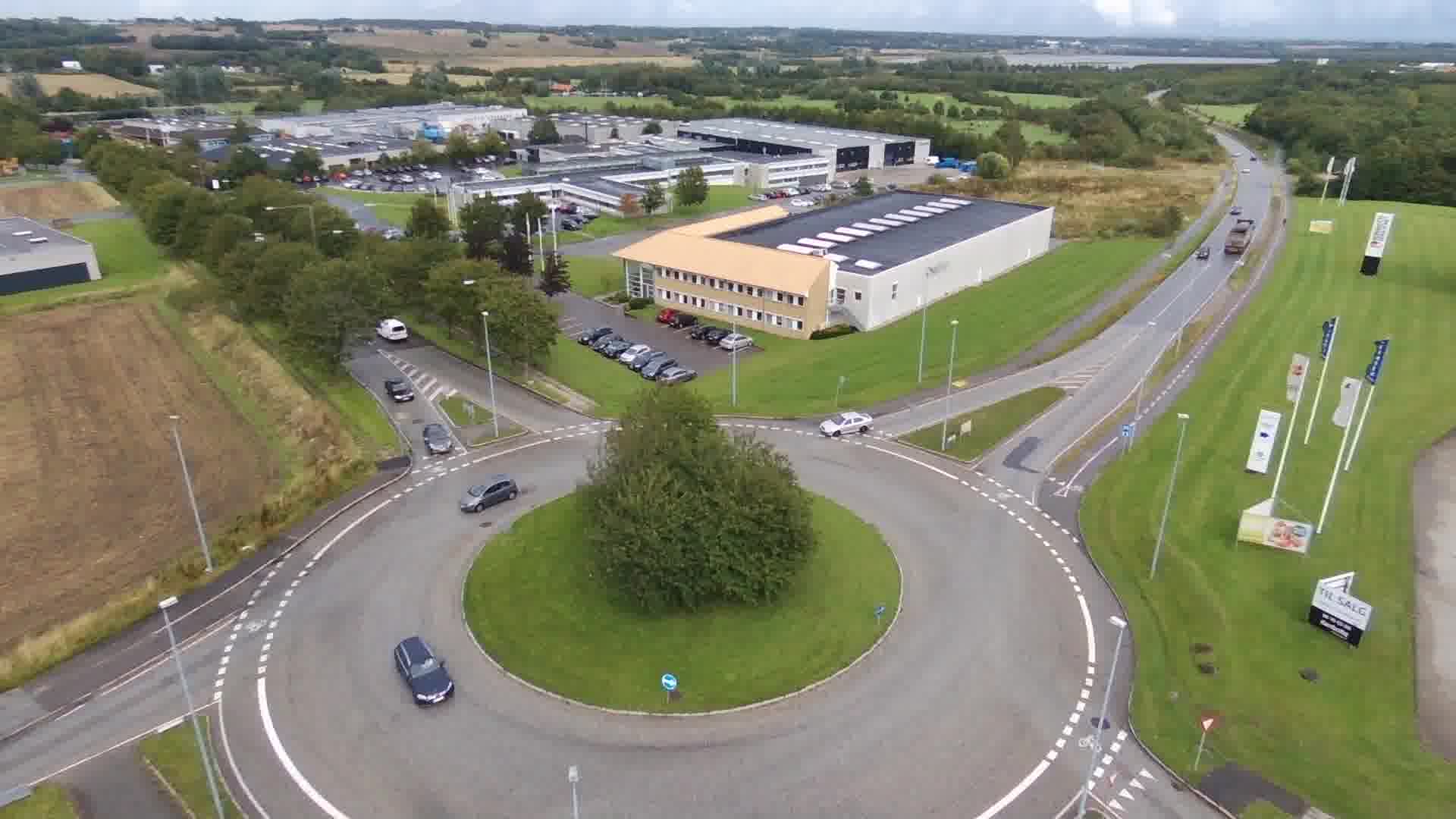}
        \end{subfigure}
        \begin{subfigure}{0.32\linewidth}
            \includegraphics[width=\linewidth, bb=0 0 1920 1080]{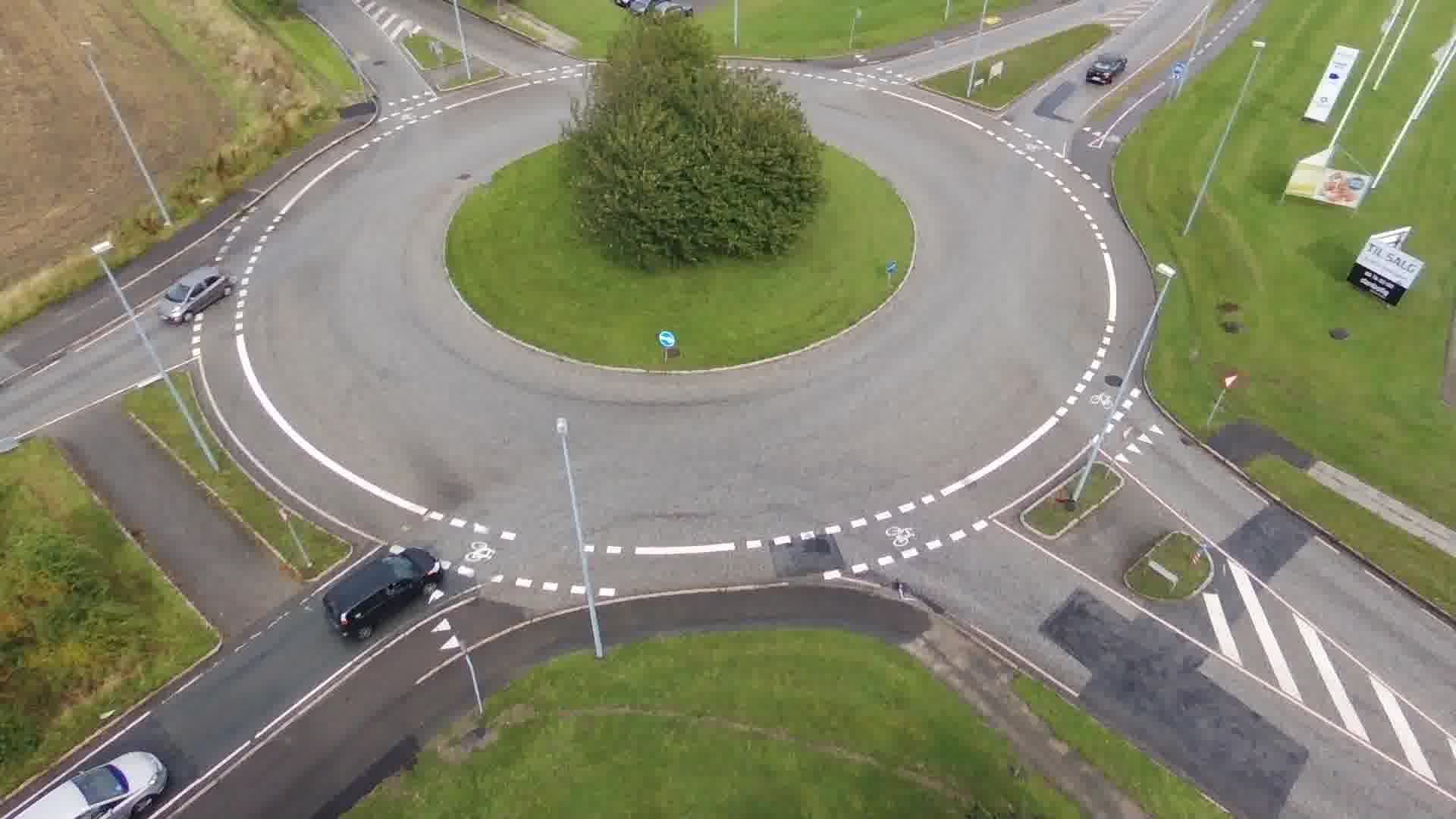}
        \end{subfigure}
    \caption{Examples of images from sequence 0 of AU-AIR-track.}
    \label{fig:AU_AIR_TRACK_prev_0}
\end{figure}

\begin{figure}[H]
     \centering
    \begin{subfigure}{0.32\linewidth}
        \includegraphics[width=\linewidth, bb=0 0 1922 1079]{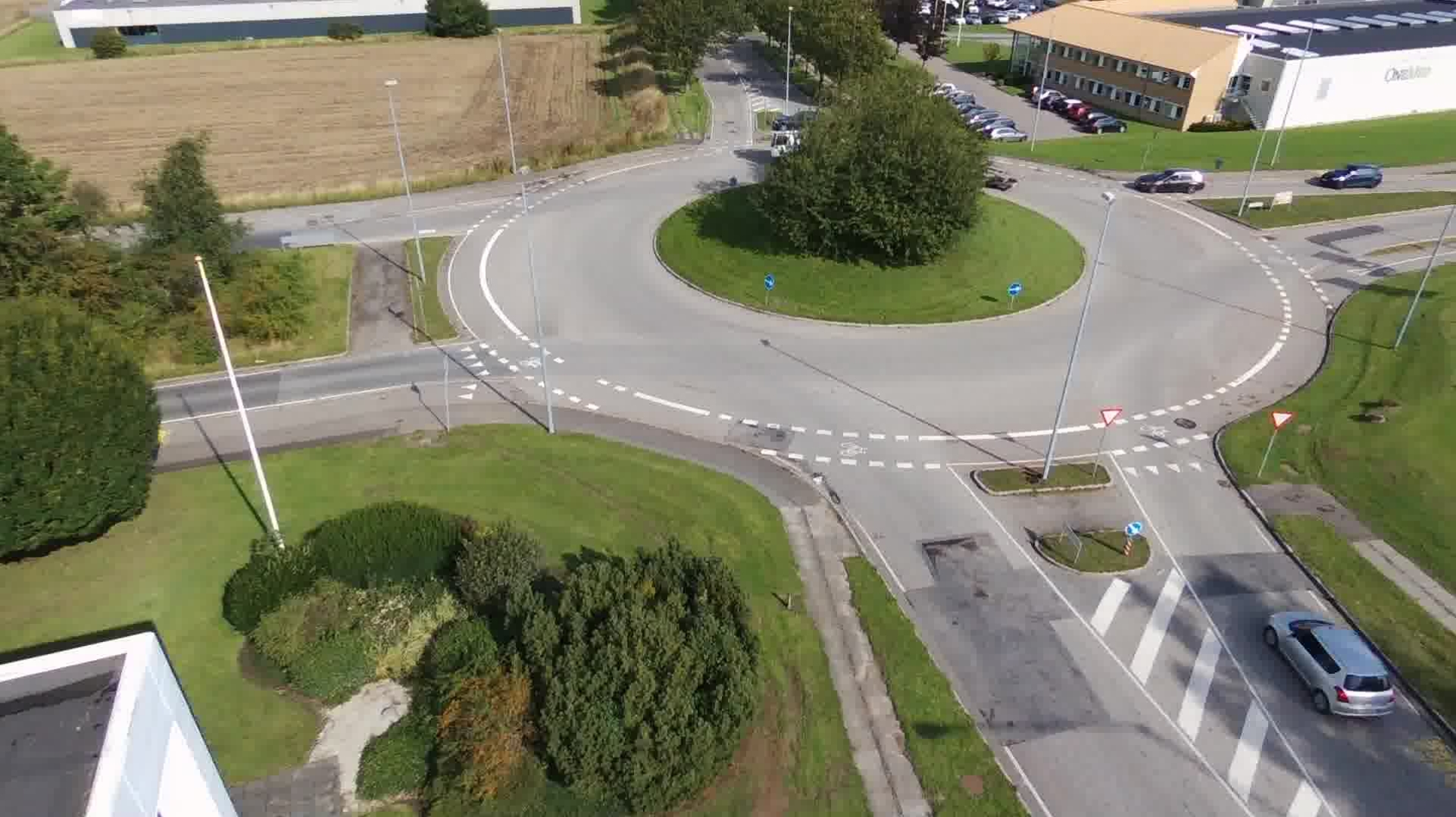}
    \end{subfigure}
    \begin{subfigure}{0.32\linewidth}
        \includegraphics[width=\linewidth, bb=0 0 1922 1079]{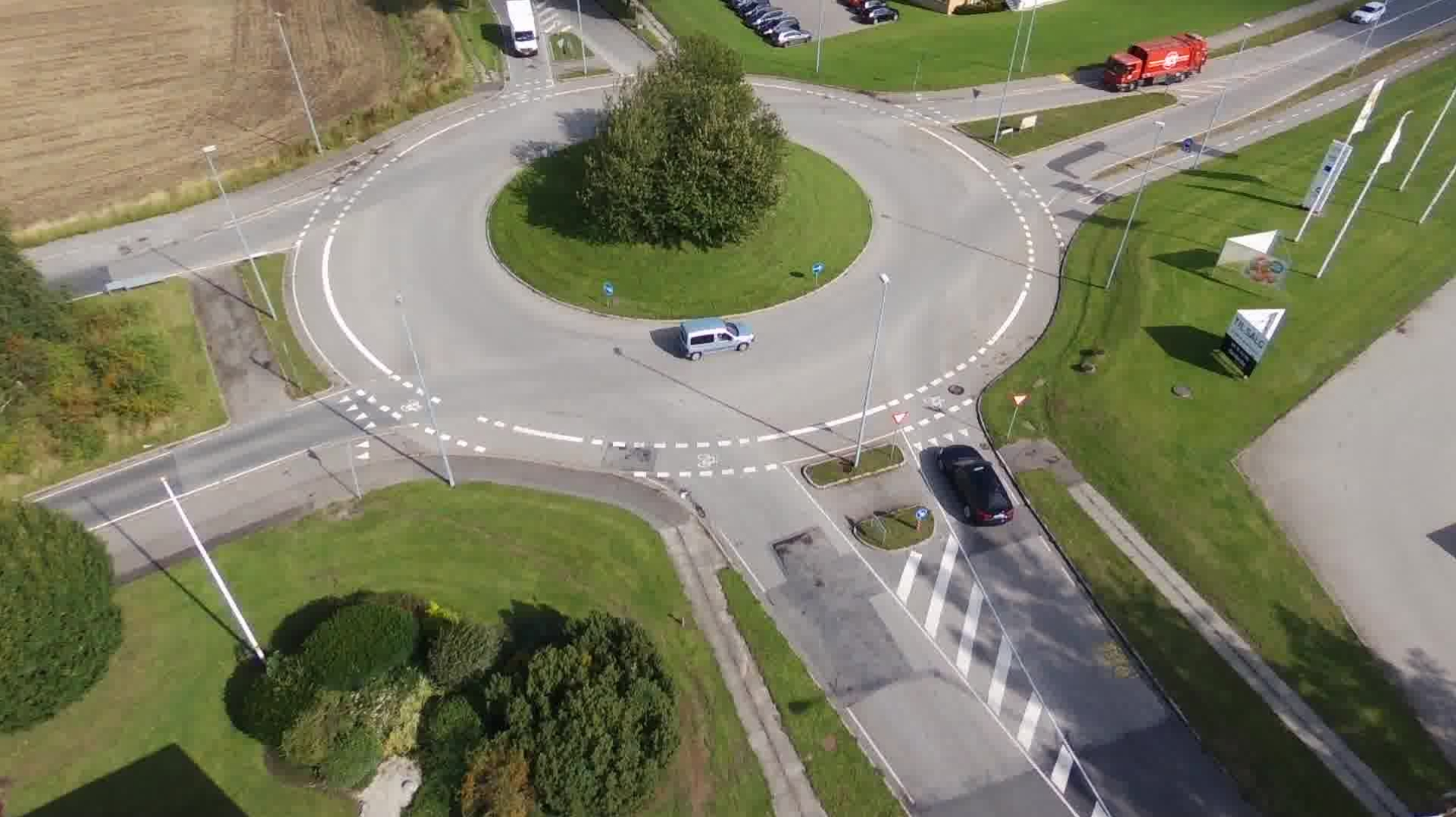}
    \end{subfigure}
    \begin{subfigure}{0.32\linewidth}
        \includegraphics[width=\linewidth, bb=0 0 1922 1079]{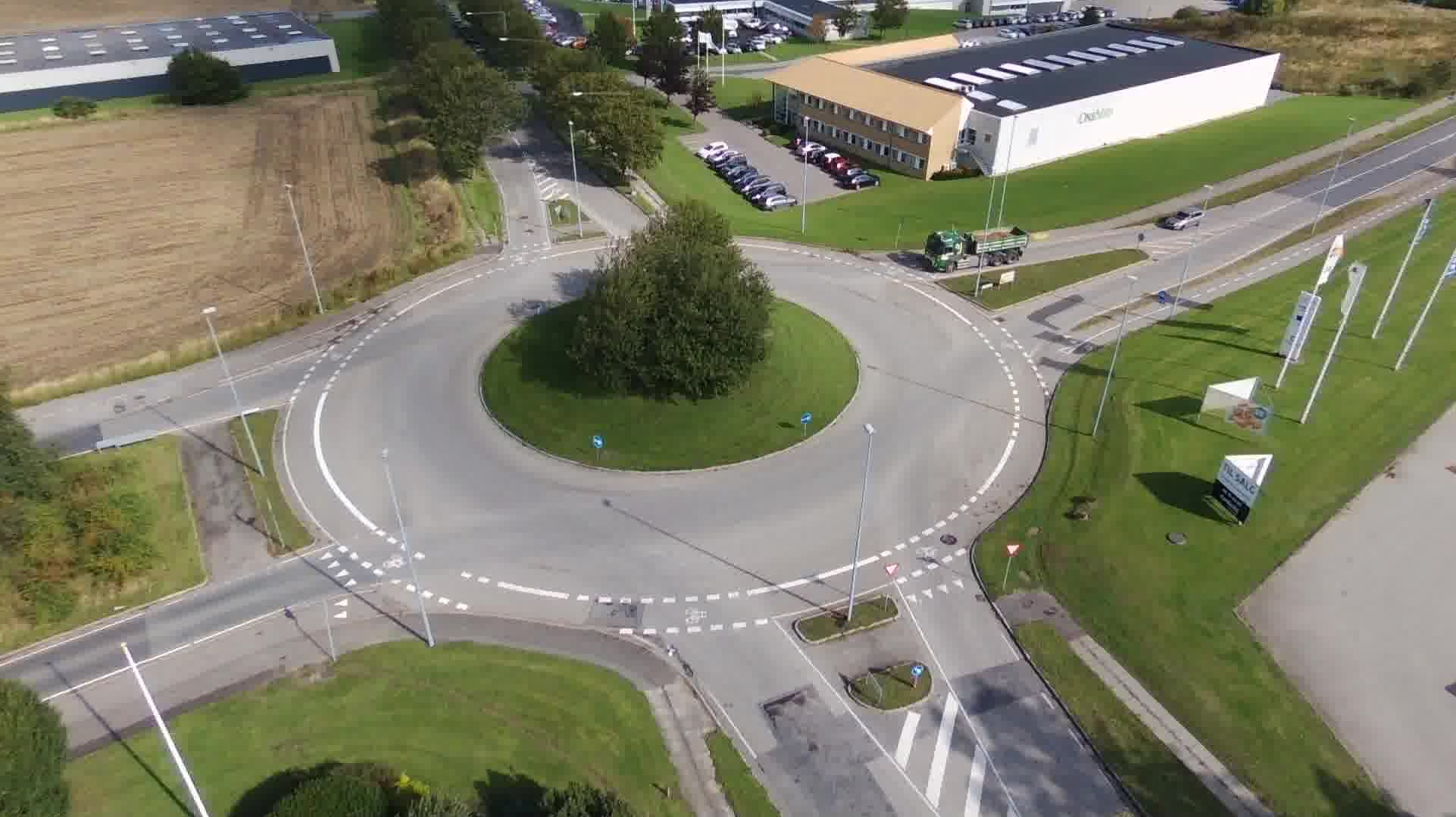}
    \end{subfigure}
    \caption{Examples of images from sequence 1 of AU-AIR-track.}
    \label{fig:AU_AIR_TRACK_prev_1}
\end{figure}

The AU-AIR-Track dataset consists of two sequences, designated~as 0 and 1. With~a total of 90 annotated objects, sequence~0 contains 887 frames and 63 annotated objects; and sequence 1 has 512 frames with 27 annotated objects. Both~sequences have been extracted at 5 frames per second and their resolution is 1920 $\times$ 1080 and 1922 $\times$ 1079 pixels, respectively. Figures~\ref{fig:AU_AIR_TRACK_prev_0} and \ref{fig:AU_AIR_TRACK_prev_1} display a few images from both sequences, which~present only oblique views of the scene taken from a nonstationary~UAV.

\begin{figure}[H]
    \centering
    \begin{subfigure}{0.45\linewidth}
        \includegraphics[width=\linewidth, bb=0 0 1920 1081]{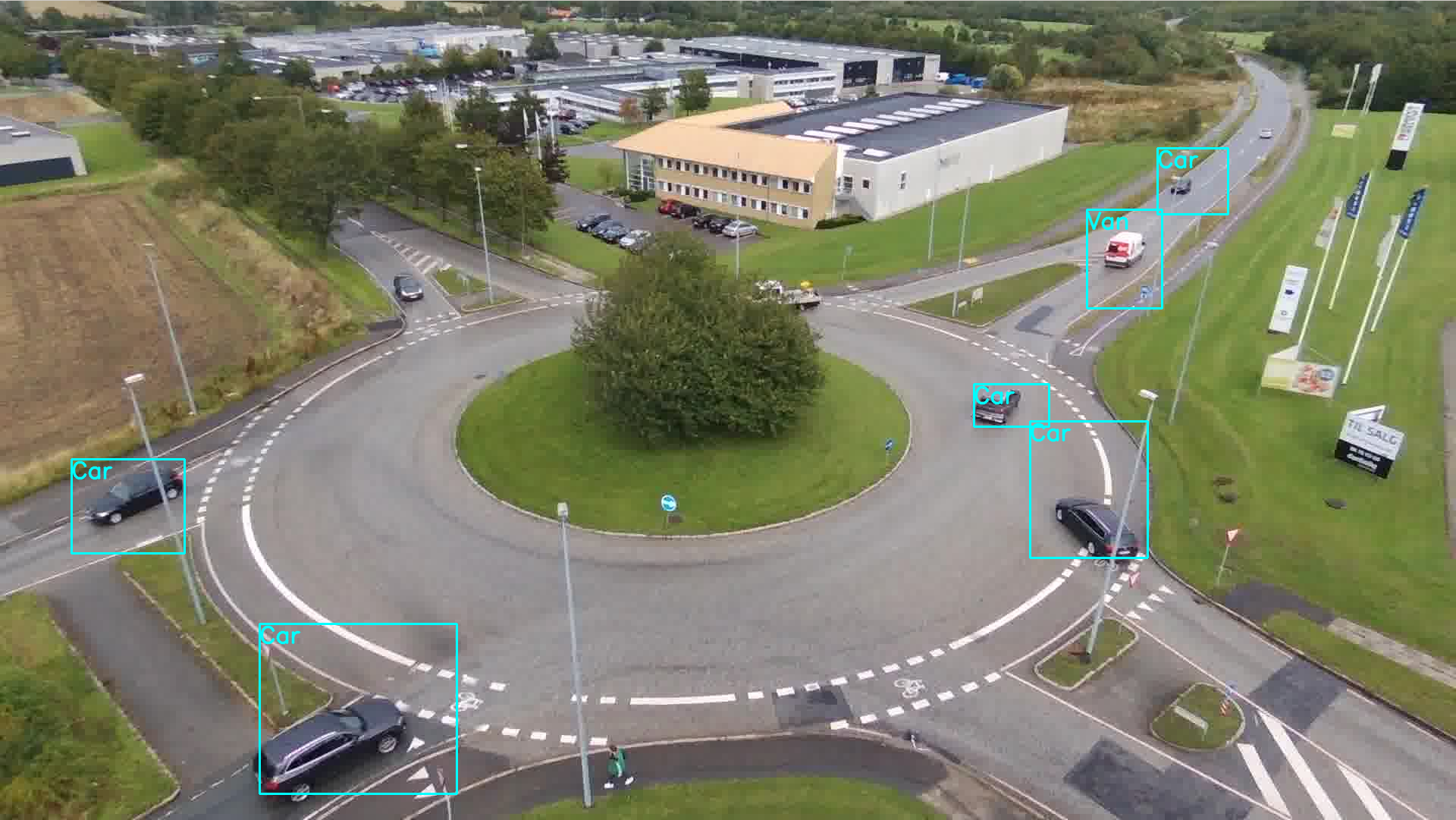}
    \end{subfigure}
    \begin{subfigure}{0.45\linewidth}
        \includegraphics[width=\linewidth, bb=0 0 1919 1079]{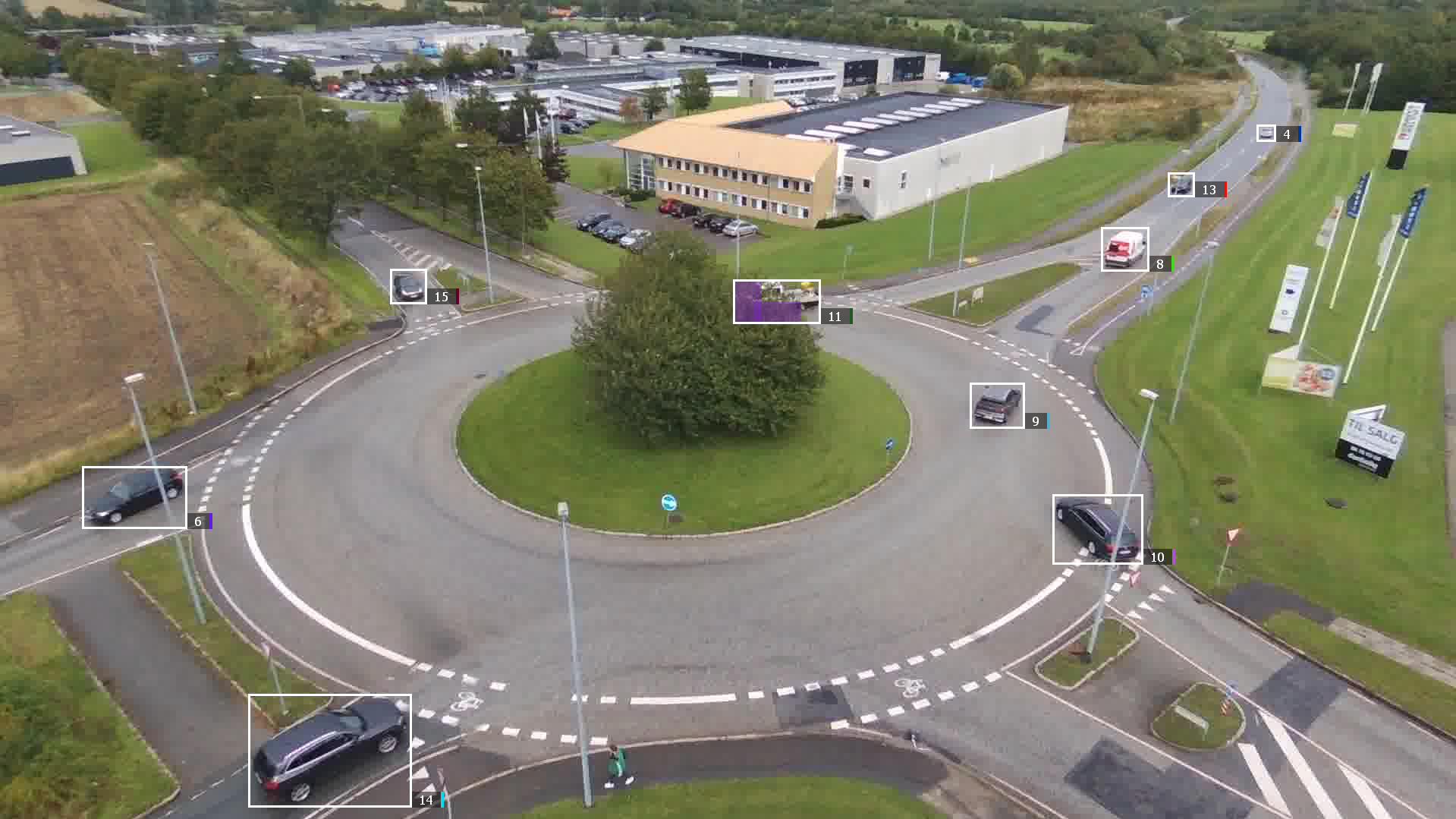}
    \end{subfigure}
    \caption{Comparison between original annotation from AU-AIR (\textbf{left}) and adapted annotations for visual object tracking (\textbf{right}). Purple~color indicates that a region of the bounding box is occluded.}
    \label{fig:old_new_anno}
\end{figure}

As a result, the~main challenges captured in AU-AIR-Track are the constant camera motion, the~low image resolution, the~presence of distractors, and~most importantly, frequent object occlusions. Since~AU-AIR annotations are designed for object detection, we~adapted them in AU-AIR-Track for visual object tracking. \figref{fig:old_new_anno} presents the original AU-AIR annotations and the adapted annotations for AU-AIR-Track (where only moving objects are annotated).

\begin{figure}[H]
    \centering
    \begin{subfigure}{0.45\linewidth}
        \includegraphics[width=1\linewidth, bb=0 0 1100 500]{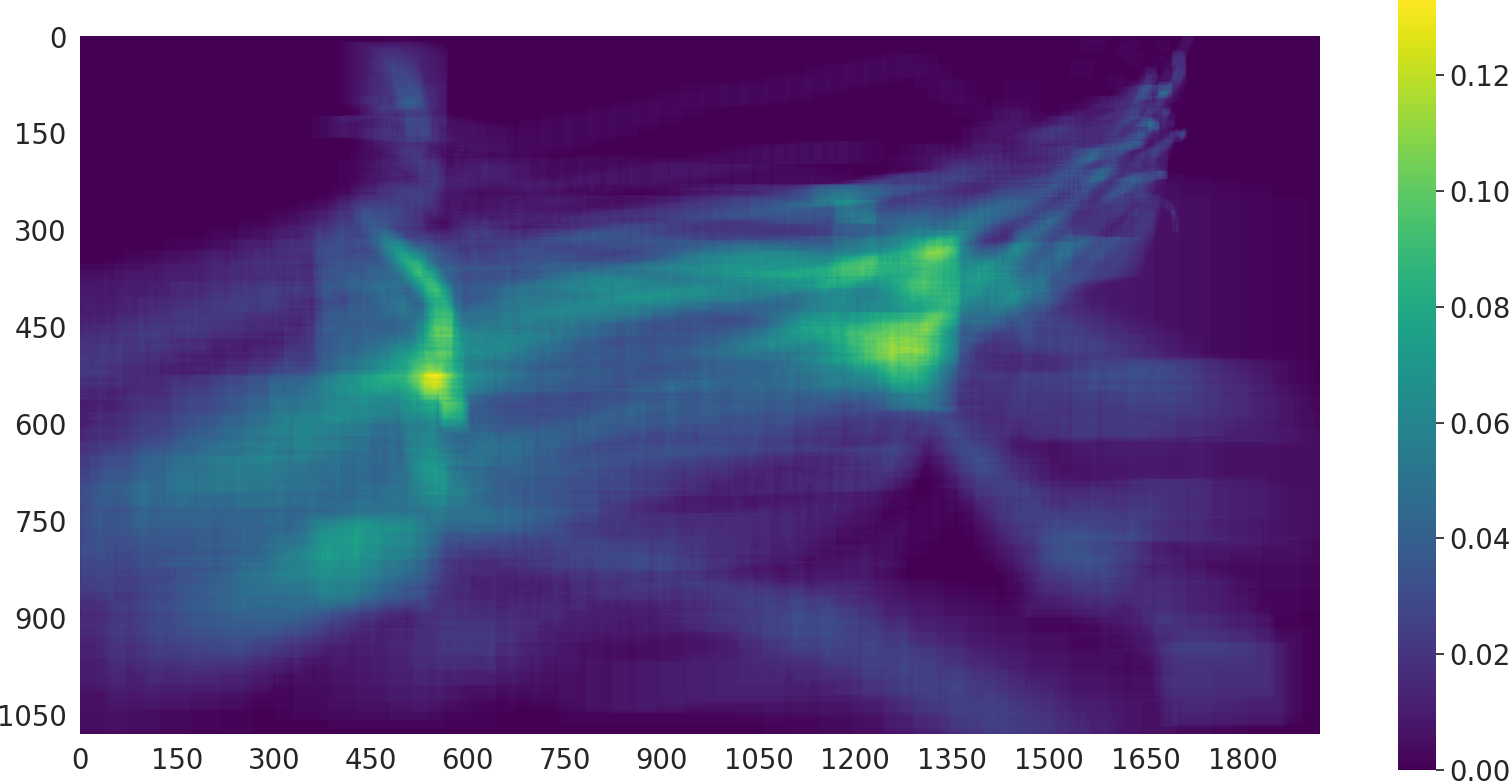}
    \end{subfigure}
    \begin{subfigure}{0.45\linewidth}
        \includegraphics[width=1\linewidth, bb=0 0 1100 500]{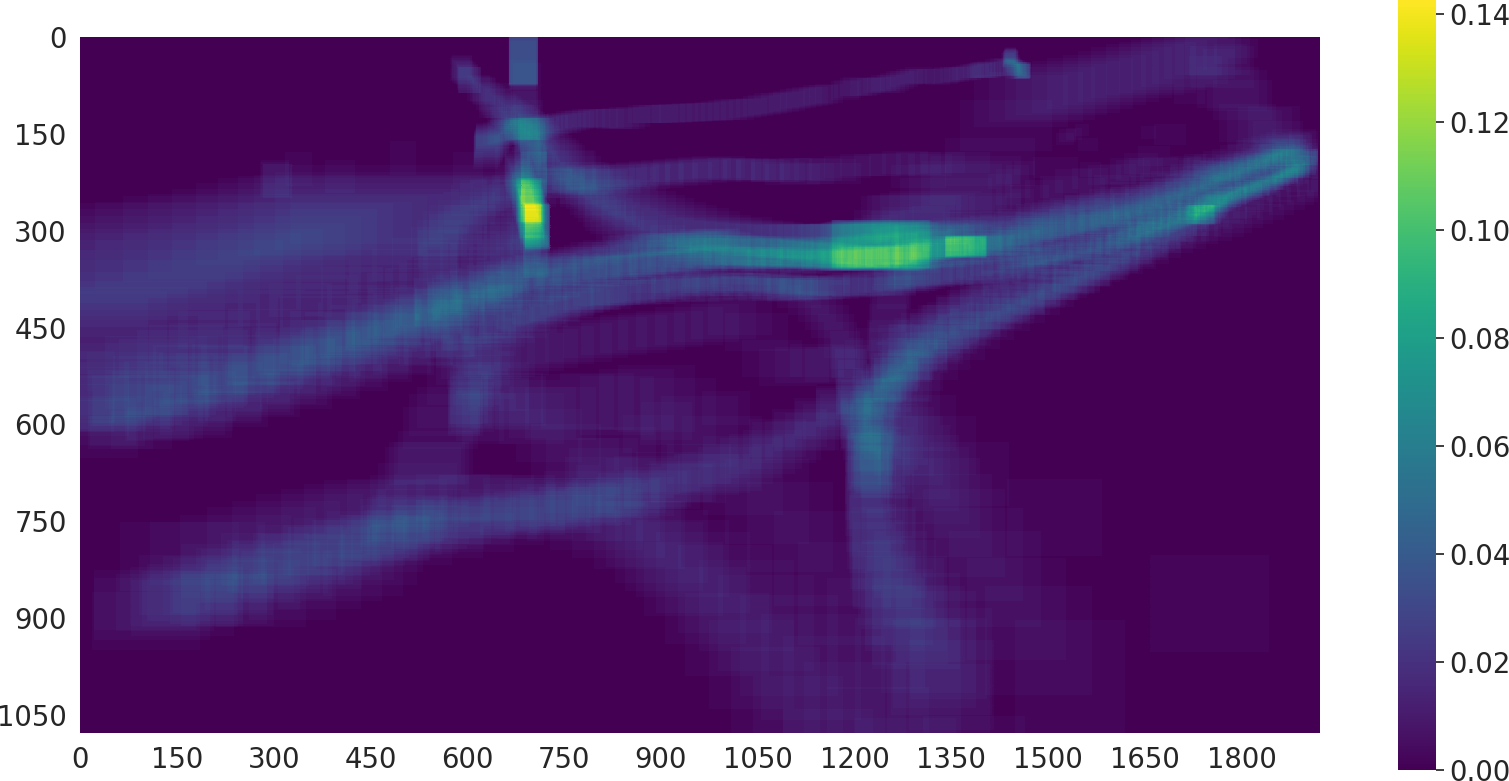}
    \end{subfigure}
    \caption{The probability distribution of ground-truth bounding boxes over the total sequence length for sequence 0 (\textbf{left}) and sequence 1 (\textbf{right}) from AU-AIR-Track.}
    \label{fig:distribution_of_GT_bbox_in_seq}
\end{figure}

\begin{figure}[H]
    \centering
    \begin{subfigure}{.45\linewidth}
        \includegraphics[width=1\linewidth, bb=0 0 1553 653]{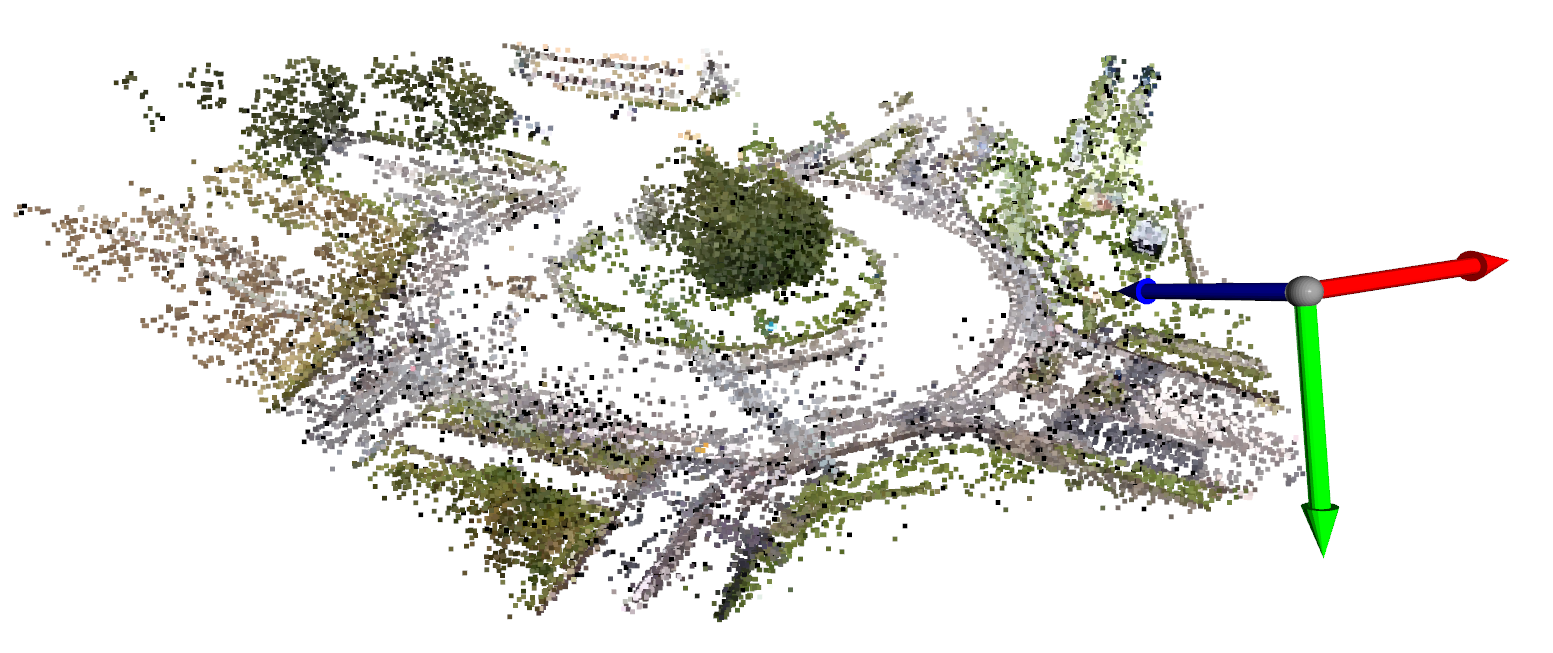}
    \end{subfigure}
    \begin{subfigure}{.45\linewidth}
        \includegraphics[width=\linewidth, bb=0 0 1777 752]{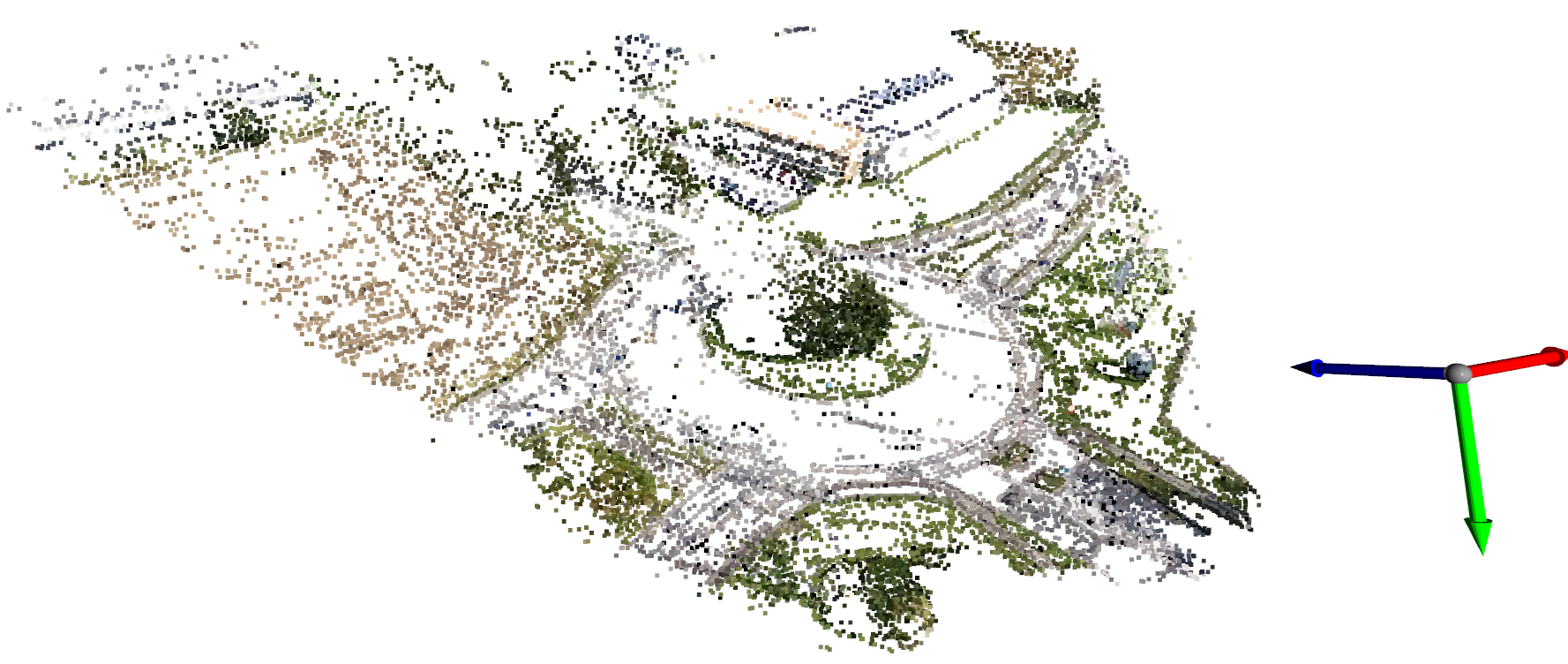}
    \end{subfigure}
    \caption{Filtered point cloud for both sequences composing the dataset. Left~is a reconstruction of sequence 0 and right of sequence 1.}
    \label{fig:filter_pnt_cld_plane}
\end{figure}

\figref{fig:distribution_of_GT_bbox_in_seq} shows the distribution of the ground-truth bounding box locations for both sequences. The~value of each pixel denotes the probability of a bounding box to cover that pixel over an entire sequence. It~can be seen that most objects follow the underlying scene structure,~i.e., the road. From~the 63 possible objects present in sequence 0, 45 objects undergo an occlusion and 25 out of 27 objects in sequence 1. As~stated before, AU-AIR-Track provides the 3D reconstructions of both sequences (see \figref{fig:filter_pnt_cld_plane}). The~3D information available in our dataset are the~sparse reconstructions with their respective ground plane estimations, camera~poses, fundamental~matrices, and transformation~matrices.

\subsection{Evaluation Metrics}
\label{Evalu_met}
Similar to long-term tracking, the~tracked object can disappear and reappear. Thus, no~manual reinitialization is done when the tracker loses the object. To~measure the performance, we~utilize common long-term metrics---tracking precision~$Pr$, tracking~recall~$Re$, and tracking F1-score~$F(\tau_\theta)$---at a given~$\tau_\theta$, introduced~in~\cite{bib:longterm_metric} and used in~\cite{bib:vot_2018, bib:vot_2019}.

Let~$G_t$ be the ground truth object bounding box, and~$A_t$ the bounding box estimation given by the tracker at frame~$t$. Further, let~$\theta_t$ denote the prediction confidence, which~in our case, is the maximal score given by the tracker regarding its confidence on the presence of the object in the current frame~$t$. If~the object is partially/fully
~occluded, we~set~$G_t = 0$; and~similarly, if~the trackers predicts an object with a confidence~$\theta_t$ below~$\tau_\theta$, we~set~$A_t = 0$. Furthermore, let~$n_g$ denote the number of frames with~$G_t \neq 0$, where~$t_g \in N_g = \{1,...,n_g\}$,~$n_p$ is the number of frames with~$A_t \neq 0$, and~$G_t \neq 0$ with~$t_p \in N_p =\{1,...,n_p\}$. Lastly,~$\Omega(A_t(\theta_t),G_t)$ describes the intersection-over-union (IoU) between~$G_t$ and~$A_t$.

\begin{equation}
    Pr(\tau_\theta) = \frac{1}{|N_p|}\displaystyle\sum_{t_p \in N_p}^{}\Omega(A_{t_p}(\theta_{t_p}),G_{t_p}),
    \label{eq:tracking_precision}
\end{equation}
\begin{equation}
    Re(\tau_\theta) = \frac{1}{|N_g|}\displaystyle\sum_{t_g\in N_g}^{}\Omega(A_{t_g}(\theta_{t_g}),G_{t_g}),
    \label{eq:tracking_recall}
\end{equation}
\begin{equation}
    \mathcal{F}(\tau_\theta) = \frac{2Pr(\tau_\theta)Re(\tau_\theta)}{Pr(\tau_\theta) + Re(\tau_\theta)}.
    \label{eq:F_measure}
\end{equation}

The combination of tracking precision~$Pr(\tau_\theta)$ and tracking recall~$Re(\tau_\theta)$ as a single score is defined as the tracking F1-score~$\mathcal{F}(\tau_\theta)$~\cite{bib:longterm_metric}. Similarly~to the long term challenges presented in \cite{bib:vot_2018, bib:vot_2019}, the final tracking F1-score is used to rank the different tracking algorithms.

The evaluation protocol is as follows: the trackers are evaluated on all objects present in the AU-AIR-Track. The~annotated first frame of the object is used to initialize the tracker. From~there, the tracker outputs a prediction bounding box for every subsequent frame where the object is annotated---even during occlusions, no reset is allowed. Tracking~precision, tracking~recall, and tracking F1-score are computed accordingly to Equations~\eqref{eq:tracking_precision}--\eqref{eq:F_measure}.
To avoid statistical errors caused by the classification component of the visual tracker, which~describes the appearance of the object through learned weights, we~run 
an evaluation of every tracker five times on both sequences of the AU-AIR-Track. For~every evaluation~$e$ and for every object~$i$ present in a sequence, we~take the maximum tracking F1-score for that object~${f_{\mathrm{max}}}^{e}_{i}$. Considering~the maximum tracking F1-score, regardless~of~$\tau_\theta$, allows~us to examine how the tracker would work without human intervention.

For computing the final F1-score~$f_{\mathrm{final}}$ of a tracker on a sequence of AU-AIR-Track: (1) we average maximum tracking F1-scores~${f_{\mathrm{max}}}^{e}_{i}$ extracted from different evaluations~$e$ for the same object~$i$; then, (2)~we average all~${f_{\mathrm{max}}}^{e}_{i}$ belonging to the same sequence.
\meqref{eq:Final_measure} describes how we computed $f_{\mathrm{final}}$. Let~~$i$ denote the index of the object, with~$i \in I=\{1,...,n\}$; and let~$e$ describe the index for an evaluation, where~$e \in E=\{1,...,m\}$. In~our case, we performed five evaluations for each tracker on the~AU-AIR-Track.

\begin{equation}
    f_{\mathrm{final}} = \frac{1}{|I||E|}\displaystyle\sum_{i \in I}^{}\displaystyle\sum_{e \in E}^{}{f_{\mathrm{max}}}_{i}^{e}.
    \label{eq:Final_measure}
\end{equation}

\section{Results}
\label{results}
In this section, we~demonstrate the effectiveness of our approach on quantitative results in \secref{quant} and through a qualitative analysis in \secref{qual}.

\subsection{Quantitative Results}
\label{quant}
In this section, we~use the following terms: (1) ``original'', which refers to the unmodified visual object tracker ATOM and DiMP presented in~\cite{bib:atom, bib:dimp}. (2) ''2D variant``, denoting ATOM and DiMP---i.e., ATOM-2D, DiMP-2D---coupled~with a particle filter, working~in the 2D image space. (3) Lastly, the~``3D variant'' refers to ATOM and DiMP---i.e., ATOM-3D, DiMP-3D---utilizing~3D information combined with a particle filter operating in this 3D scene space. Table~\ref{tab:F_1_atom_dimp_seq_0_seq_1} summarizes the final tracking F1-scores~$f_{\mathrm{final}}$ of ATOM and DiMP for every variation. Best~$f_{\mathrm{final}}$ on each sequence and for each variant are indicated in bold.

\begin{table}[H]
\centering
\caption{Final tracking F1-scores~$f_{\mathrm{final}}$ and standard deviation for every tracker variation evaluated on AU-AIR-Track.}
\begin{tabular}{llllll}
\toprule
                                                &               & \multicolumn{2}{c}{\textbf{Sequence 0}}                        & \multicolumn{2}{c}{\textbf{Sequence 1}}                        \\ \cmidrule(lr){3-4}\cmidrule(lr){5-6}
\textbf{Variants}                                        & \textbf{Tracker}       & \boldmath{$f_{\mathrm{final}}$} \boldmath{$\uparrow$}   & \textbf{std} \boldmath{$\downarrow$}  & \boldmath{$f_{\mathrm{final}}$} \boldmath{$\uparrow$}   & \textbf{std} \boldmath{$\downarrow$}  \\ \midrule
\multirow{2}{*}{Original}                       & ATOM          & 0.625                             & 0.233             & 0.577                             & 0.232             \\
                                                & DiMP          & 0.593                             & 0.239             & 0.579                             & 0.239             \\ \midrule
\multirow{2}{*}{2D Variant (ours)}
  & ATOM-2D       & 0.641                             & 0.223             & 0.605                             & 0.242             \\
                                                & DiMP-2D       & 0.663                             & 0.201             & 0.625                             & 0.235             \\ \midrule
\multirow{2}{*}{\textbf{3D Variant (ours)}}     & ATOM-3D       & \textbf{0.684}                    & \textbf{0.193}    & \textbf{0.691}                    & \textbf{0.179}    \\
                                                & DiMP-3D       & \textbf{0.715}                    & \textbf{0.161}    & \textbf{0.701}                    & \textbf{0.143}    \\
\bottomrule
\end{tabular}
\label{tab:F_1_atom_dimp_seq_0_seq_1}
\end{table}

Based on Table~\ref{tab:F_1_atom_dimp_seq_0_seq_1}, we~observe that the original variations of ATOM and DiMP attain the lowest scores and the least stable results. As~stated before, the~original methods are designed for short-term tracking from a ground-level perspective by relying on visual cues. The~original variants have no specific module integrated for handling partial or full occlusions apart from using a similarity score map with a threshold. If~the similarity score map has no peaks (under a set threshold) and is widely spread, the~tracker is able to recognize that the object is missing but is not able to predict the next position. The~original variation is also more prone to switching the object with a false association---i.e., the distractor---because~of the maximum-likelihood approach. Since~the original trackers rely only on the learned appearance model, they~are extremely dependent on the number of pixels that encode the object. This~results in the tracker losing most of the tracked object when they are described with a low amount of pixels, i.e., small-scale objects.

ATOM-2D and DiMP-2D also recognize occlusion only through visual cues, achieved~by setting a minimum required similarity score as a threshold. Occlusion~is identified by obtaining a similarity score that is below the set threshold. During~occlusion, the~position of the object cannot be inferred visually but can be estimated (to some extent) through the predictions of the particle filter. Relying~solely on visual cues for identifying occlusion is limited since the tracker potentially misinterprets a fast appearance change with an occlusion. Another~limitation is that only occlusions without ego-motion can be handled because the particle filter estimates positions in the 2D image space. Overall, there~is an increase in the tracking F1-score compared to the original variations, but~this gain is essentially due to better recognizing and handling of distractors and small-scale objects, leveraged~by a multimodal state estimator. With~the multimodal property, we~allow groups of particles to form where high responses in the similarity score map are found. The~different groups of particles are clustered depending on their locations, enabling~the 2D trackers to consider and distinguish the object and distractors. Whereas, the~original trackers, utilizing~a maximum-likelihood approach, can~only handle the highest response in the similarity score map. Using~a state estimator also offers the benefit of being less dependent on the number of pixels used for encoding the object.

Regarding ATOM-3D and DiMP-3D, this~variation achieves the best performance on the AU-AIR-Track dataset. ATOM-3D and DiMP-3D can identify occlusions not only based on visual cues but also through depth information leveraged by the 3D reconstruction of the scene (depth map approximations). This~allows them to recognize a hidden object more reliably than the previous variations. Using~a particle filter in the 3D scene space enables the usage of a 3D transition model, which~adequately describes real-world motions in comparison to a particle filter in 2D image space. This~results in improved stability of corresponding predictions~\wrt ego-motions. Thus, the~predictions of the state estimator are more accurate than in the 2D variants when the object is hidden, allowing~them to potentially estimate the position of the object for a longer period.
Additionally, in~this variation, the~particle filter enhances the ability (similarly to ATOM-2D and DiMP-2D) of the tracker to distinguish the objects from distractors and to be less dependent on the number of pixels describing the object. Based~on these results, the~ATOM-3D and DiMP-3D display better performance in comparison to the original and 2D~variants.

\subsection{Qualitative Analysis}
\label{qual}
In this section, we~discuss selected qualitative tracking scenarios to verify the overall viability of the different methods. To~illustrate the results, an~in-depth look is provided by examining the variants of DiMP. For~every figure, a~closer look at the scenario is shown on the lower part of the figure, based~on a region delimited by a red dash--dot rectangle in the corresponding upper image.

\begin{figure}[H]
    \centering
    \begin{subfigure}{0.28\linewidth}
        \begin{tikzpicture}
            \node[anchor=south west,inner sep=0] (image) at (0,0) {\includegraphics[width=\linewidth, bb=0 0 1920 1080]{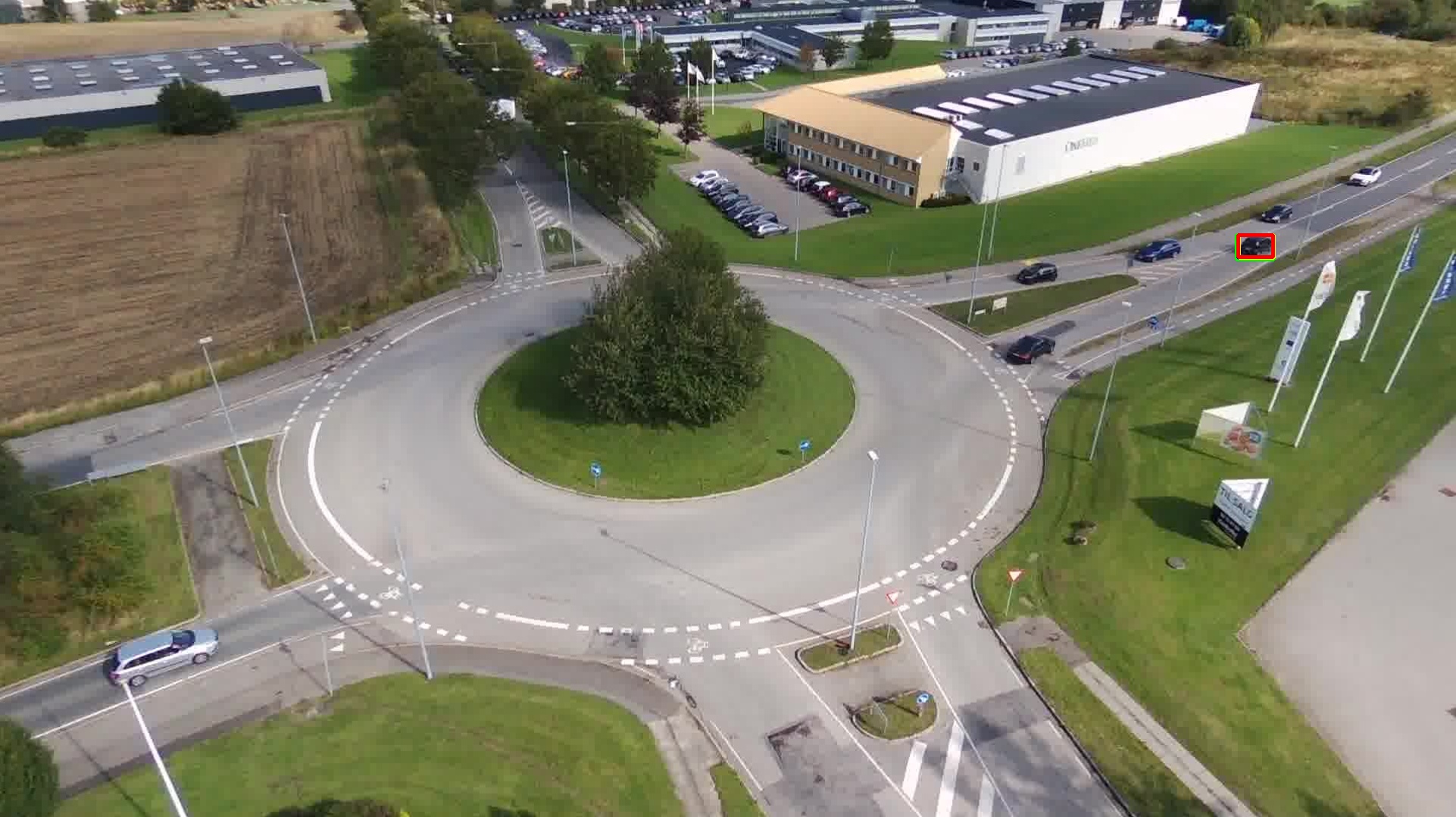}};
            \begin{scope}[x={(image.south east)},y={(image.north west)}]
                \draw[red, thick, dashdotted] (0.71,0.60) rectangle (0.97,0.82);
            \end{scope}
        \end{tikzpicture}
    \end{subfigure}\hspace{0.0005em}
    \begin{subfigure}{0.28\linewidth}
        \begin{tikzpicture}
            \node[anchor=south west,inner sep=0] (image) at (0,0) {\includegraphics[width=\linewidth, bb=0 0 1920 1080]{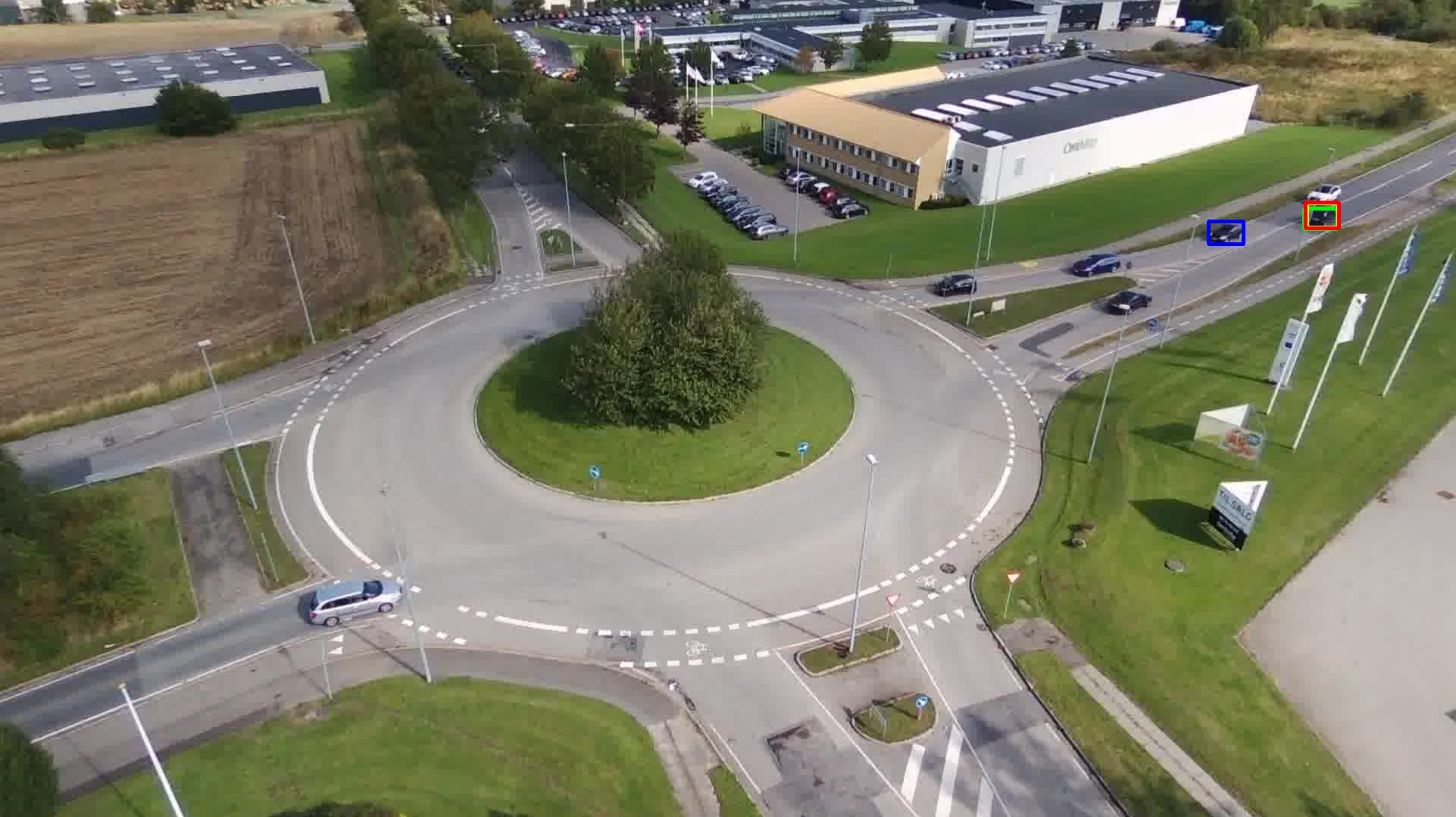}};
            \begin{scope}[x={(image.south east)},y={(image.north west)}]
                \draw[red, thick, dashdotted] (0.74,0.63) rectangle (1.,0.84);
            \end{scope}
        \end{tikzpicture}
    \end{subfigure}\hspace{0.0005em}
    \begin{subfigure}{0.28\linewidth}
        \begin{tikzpicture}
            \node[anchor=south west,inner sep=0] (image) at (0,0) {\includegraphics[width=\linewidth, bb=0 0 1920 1080]{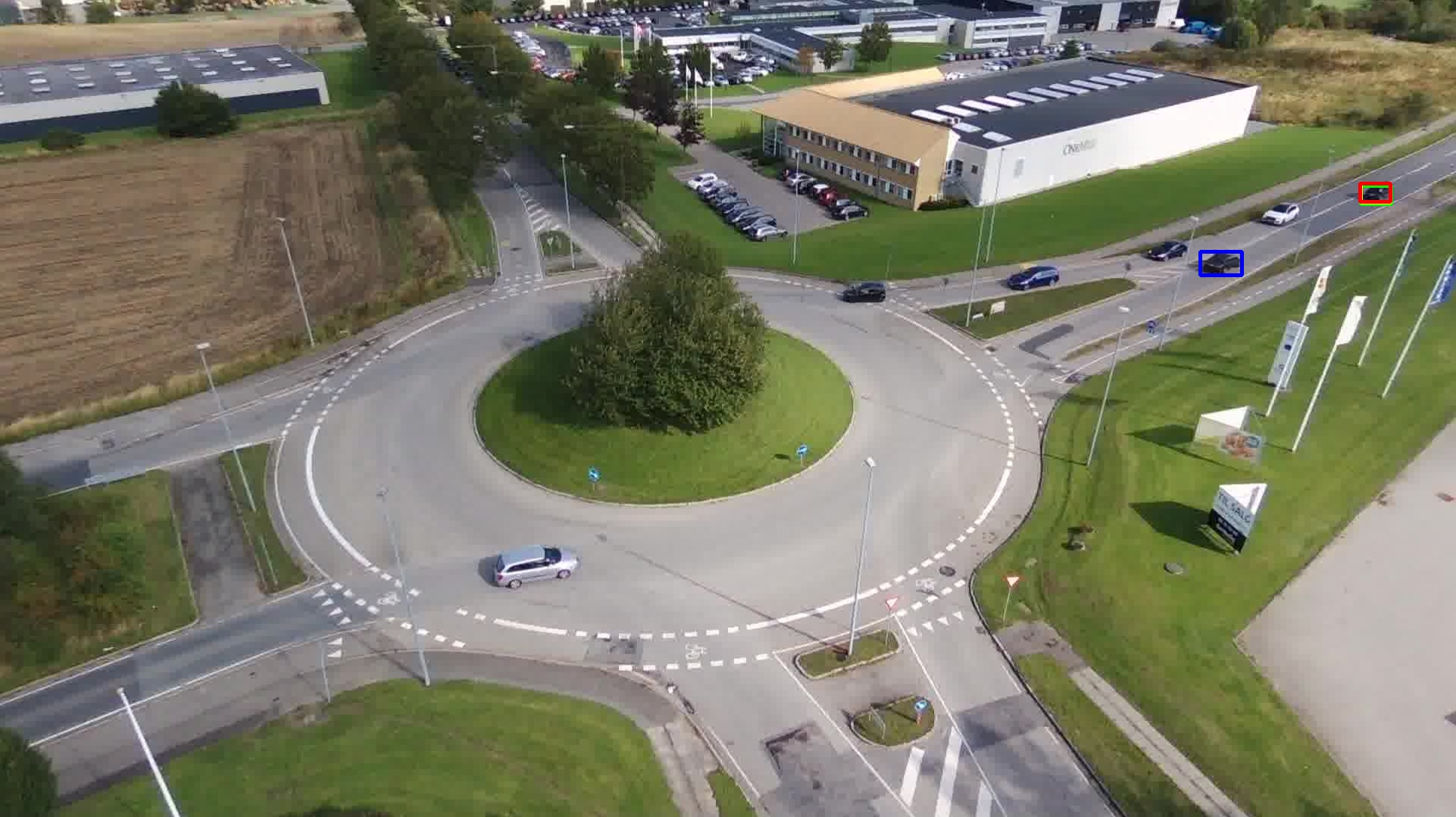}};
            \begin{scope}[x={(image.south east)},y={(image.north west)}]
                \draw[red, thick, dashdotted] (0.74,0.63) rectangle (1.,0.84);
            \end{scope}
        \end{tikzpicture}
    \end{subfigure}
    \begin{subfigure}{0.28\linewidth}
    \centering
        \includegraphics[width=\linewidth, bb=0 0 1920 1080, trim=48cm 23cm 2cm 7cm, clip=true]{Graphics/results/Seq_1_t_7_01.jpg}
    \end{subfigure}\hspace{0.0005em}
    \begin{subfigure}{0.28\linewidth}
        \includegraphics[width=\linewidth, bb=0 0 1920 1080, trim=50cm 24cm 0cm 6cm, clip=true]{Graphics/results/Seq_1_t_7_02.jpg}
    \end{subfigure}\hspace{0.0005em}
    \begin{subfigure}{0.28\linewidth}
        \includegraphics[width=\linewidth, bb=0 0 1920 1080, trim=50cm 24cm 0cm 6cm, clip=true]{Graphics/results/Seq_1_t_7_03.jpg}
    \end{subfigure}
    \begin{subfigure}{1\linewidth}
        \centering
        \includegraphics[width=.84\linewidth]{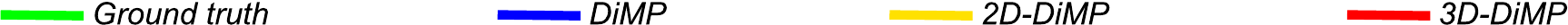}
    \end{subfigure}
    \caption{Comparison of the evaluated trackers. DiMP-2D and DiMP-3D are able handle the presence of a distractors by taking advantage of the multimodal representation provided by a particle filter. In~contrast, using the maximum-likelihood approaches, DiMP~switches to a distractor.}
    \label{fig:18}
\end{figure}

\begin{figure}[H]
    \centering
    \begin{subfigure}{0.28\linewidth}
        \begin{tikzpicture}
            \node[anchor=south west,inner sep=0] (image) at (0,0) {\includegraphics[width=\linewidth, bb=0 0 1920 1080]{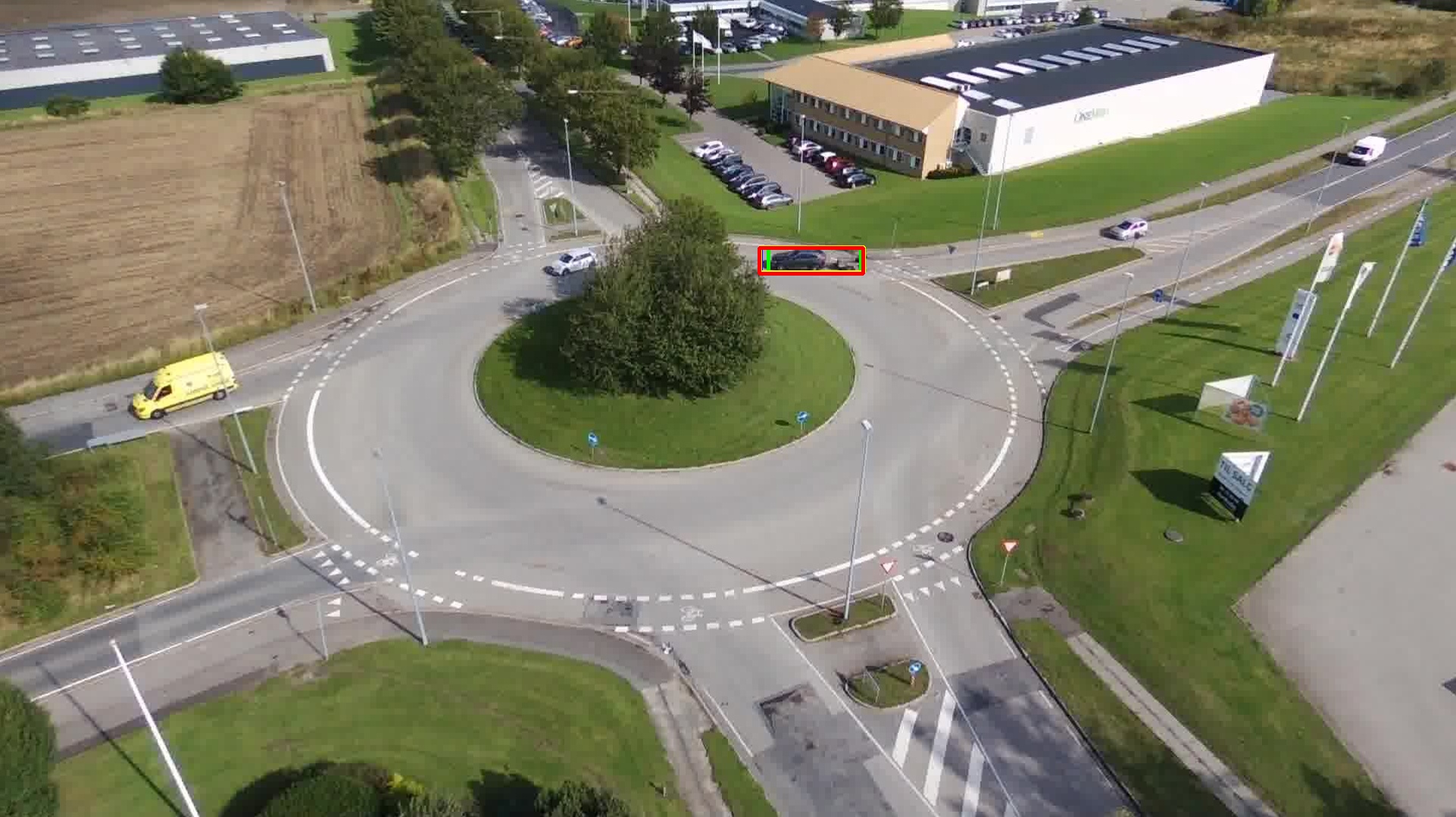}};
            \begin{scope}[x={(image.south east)},y={(image.north west)}]
                \draw[red, thick, dashdotted] (0.43,0.58) rectangle (.69,0.78);
            \end{scope}
        \end{tikzpicture}
    \end{subfigure}\hspace{0.0005em}
    \begin{subfigure}{0.28\linewidth}
        \begin{tikzpicture}
            \node[anchor=south west,inner sep=0] (image) at (0,0) {\includegraphics[width=\linewidth, bb=0 0 1920 1080]{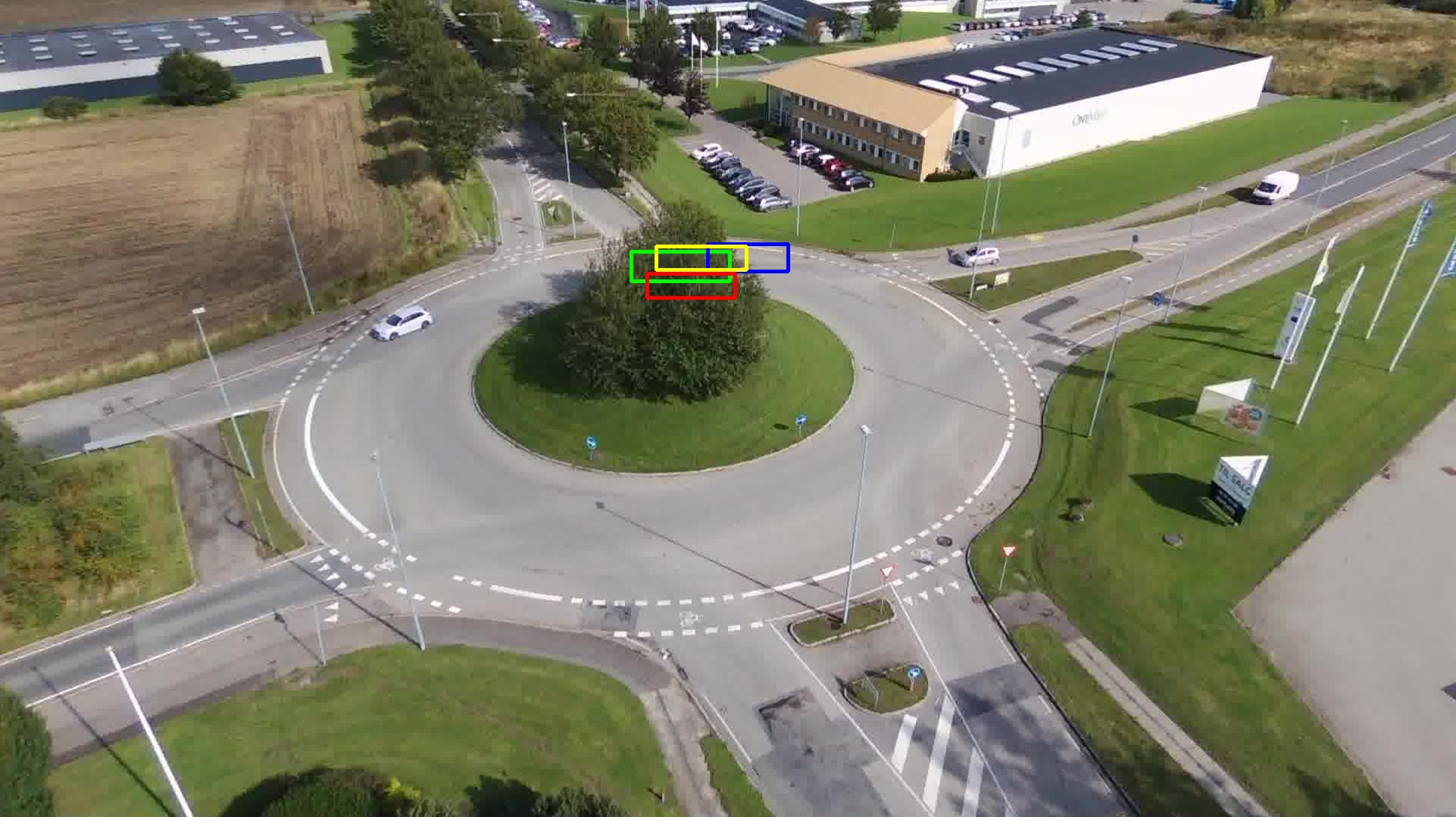}};
            \begin{scope}[x={(image.south east)},y={(image.north west)}]
                \draw[red, thick, dashdotted] (0.37,0.58) rectangle (.63,0.78);
            \end{scope}
        \end{tikzpicture}
    \end{subfigure}\hspace{0.0005em}
    \begin{subfigure}{0.28\linewidth}
        \begin{tikzpicture}
            \node[anchor=south west,inner sep=0] (image) at (0,0) {\includegraphics[width=\linewidth, bb=0 0 1920 1080]{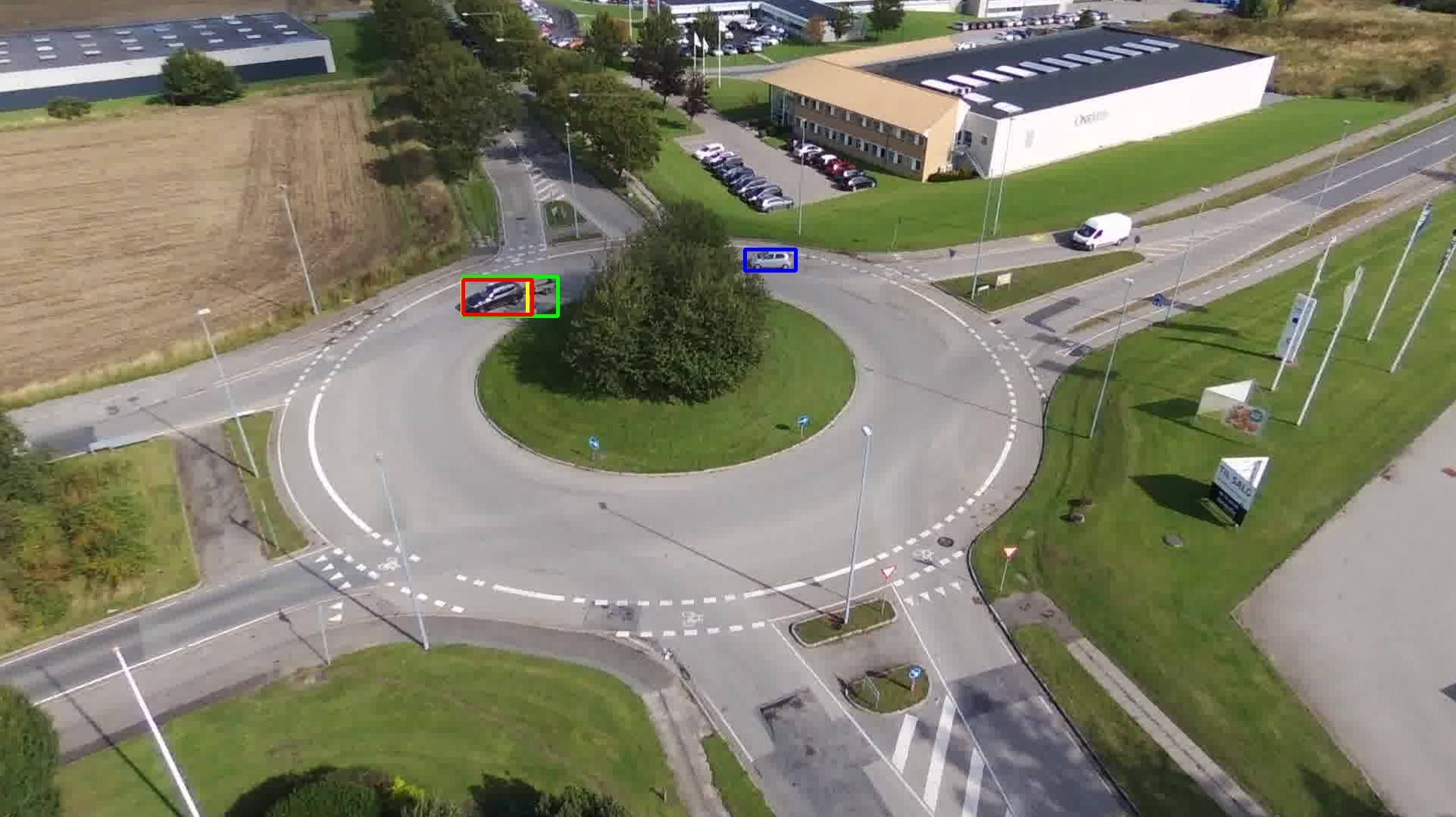}};
            \begin{scope}[x={(image.south east)},y={(image.north west)}]
                \draw[red, thick, dashdotted] (0.30,0.58) rectangle (.56,0.78);
            \end{scope}
        \end{tikzpicture}
    \end{subfigure}
    \begin{subfigure}{0.28\linewidth}
    \centering
        \includegraphics[width=\linewidth, bb=0 0 1920 1080, trim=29cm 22cm 21cm 8cm, clip=true]{Graphics/results/Seq_1_t_21_00.jpg}
    \end{subfigure}\hspace{0.0005em}
    \begin{subfigure}{0.28\linewidth}
        \includegraphics[width=\linewidth, bb=0 0 1920 1080, trim=25cm 22cm 25cm 8cm, clip=true]{Graphics/results/Seq_1_t_21_01.jpg}
    \end{subfigure}\hspace{0.0005em}
    \begin{subfigure}{0.28\linewidth}
        \includegraphics[width=\linewidth, bb=0 0 1920 1080, trim=20cm 22cm 30cm 8cm, clip=true]{Graphics/results/Seq_1_t_21_03.jpg}
    \end{subfigure}
    \begin{subfigure}{1\linewidth}
        \centering
        \includegraphics[width=.84\linewidth]{Graphics/results/DiMP_Legends.pdf}
    \end{subfigure}
    \caption{Comparison between the three DiMP variations. In~this scenario, the~2D- and 3D-DiMP methods can handle the object undergoing occlusion.}
    \label{fig:20}
\end{figure}

Figures~\ref{fig:18} and \ref{fig:20} display scenarios where the object is lost by the original DiMP variant in contrast to DiMP-2D and DiMP-3D. In~the first scenario, the~original variant, which~uses a maximum-likelihood approach, misinterprets the distractor with the object because of the low number of pixels encoding the object. In~contrast, by leveraging a particle filter, DiMP-2D and DiMP-3D handle~the presence of distractors better and are less prone to fail on small-scale objects. The~second scenario presents an occlusion situation, where~the object is hidden by the tree in the roundabout. Both~DiMP-2D and DiMP-3D recognize occlusion and can rely on their respective particle filter for predicting the position of the hidden object, until its reappearance.

\begin{figure}[H]
    \centering
    \begin{subfigure}{0.28\linewidth}
        \begin{tikzpicture}
            \node[anchor=south west,inner sep=0] (image) at (0,0) {\includegraphics[width=\linewidth, bb=0 0 1920 1080]{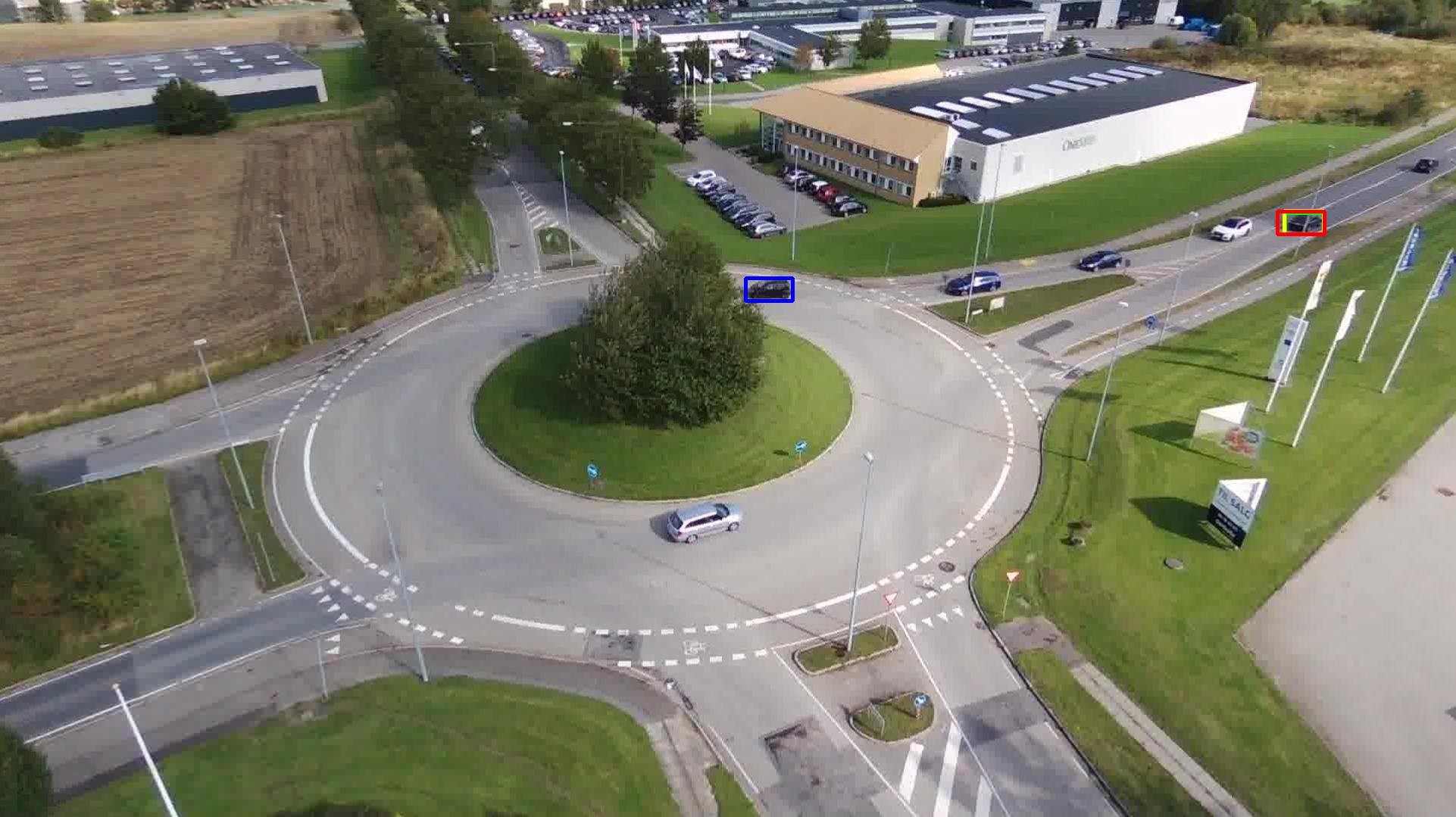}};
            \begin{scope}[x={(image.south east)},y={(image.north west)}]
                \draw[cyan!40!, ultra thin, xstep=.2, ystep=.2] (0,0) grid (1,1);
                \draw[red, thick, dashdotted] (0.74,0.60) rectangle (1.,0.82);
            \end{scope}
        \end{tikzpicture}
    \end{subfigure}\hspace{0.0005em}
    \begin{subfigure}{0.28\linewidth}
        \begin{tikzpicture}
            \node[anchor=south west,inner sep=0] (image) at (0,0) {\includegraphics[width=\linewidth, bb=0 0 1920 1080]{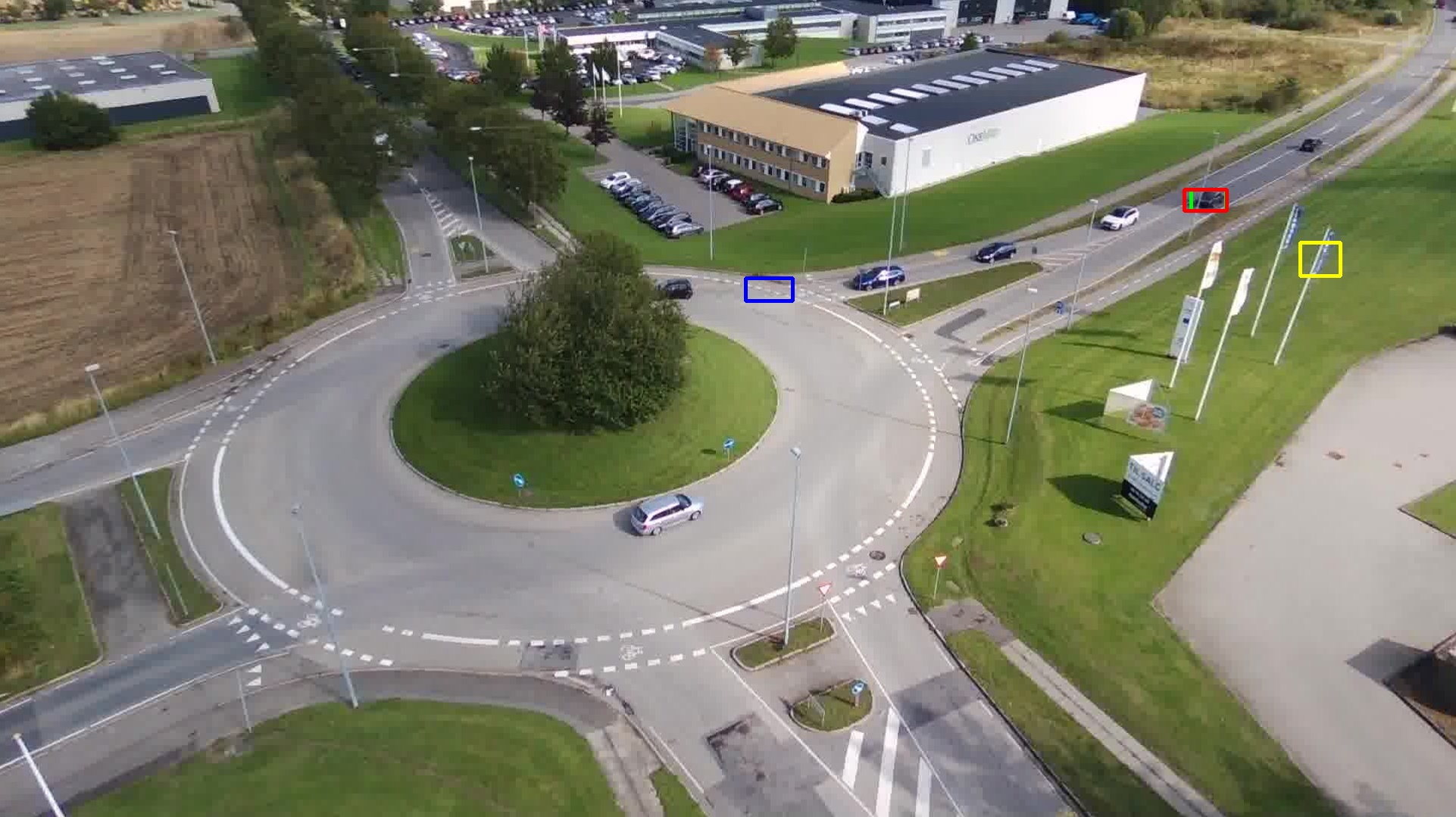}};
            \begin{scope}[x={(image.south east)},y={(image.north west)}]
                \draw[cyan!40!, ultra thin, xstep=.2, ystep=.2] (0,0) grid (1,1);
                \draw[red, thick, dashdotted] (0.74,0.62) rectangle (1.,0.84);
            \end{scope}
        \end{tikzpicture}
    \end{subfigure}\hspace{0.0005em}
    \begin{subfigure}{0.28\linewidth}
        \begin{tikzpicture}
            \node[anchor=south west,inner sep=0] (image) at (0,0) {\includegraphics[width=\linewidth, bb=0 0 1920 1080]{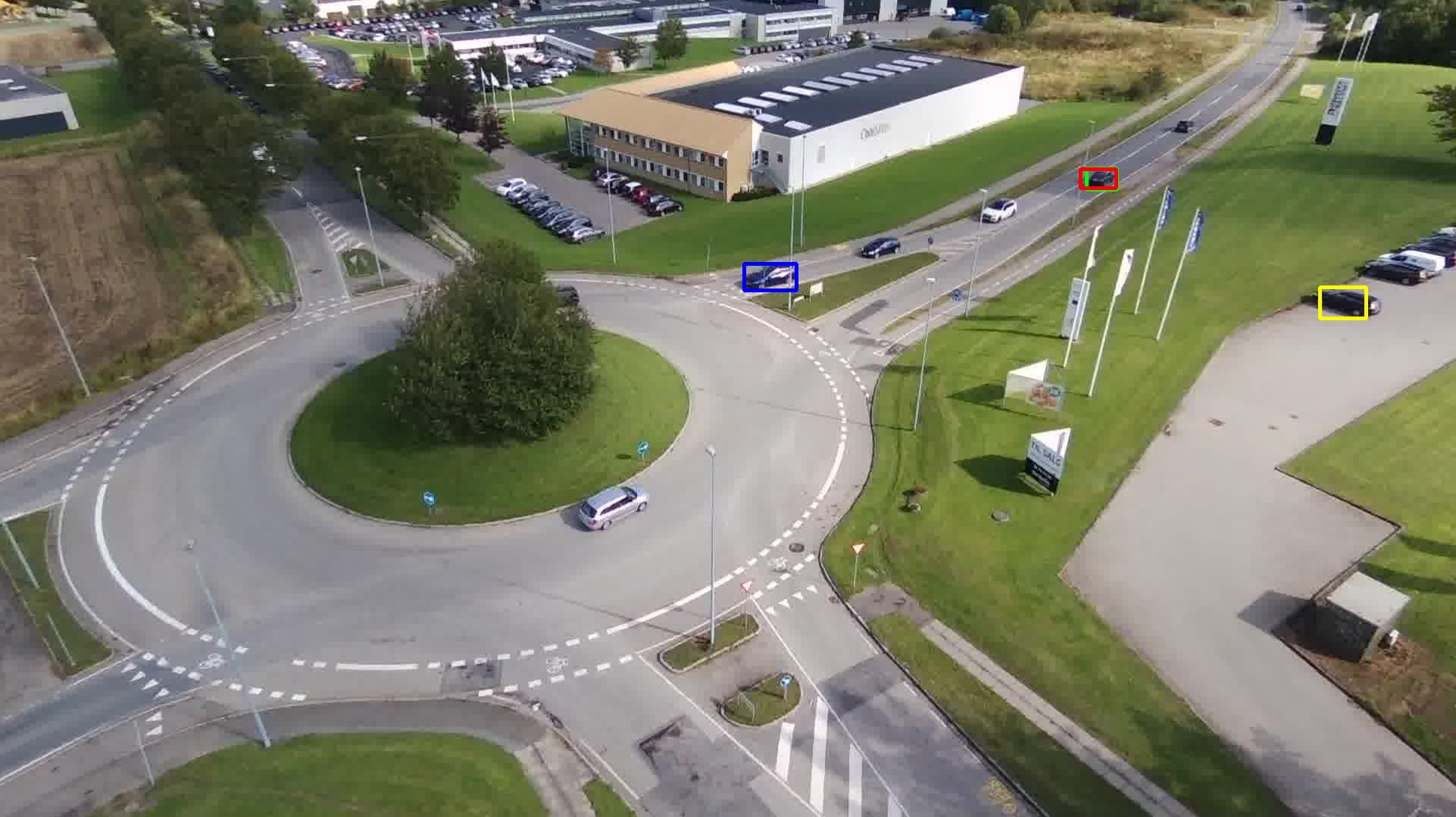}};
            \begin{scope}[x={(image.south east)},y={(image.north west)}]
                \draw[cyan!40!, ultra thin, xstep=.2, ystep=.2] (0,0) grid (1,1);
                \draw[red, thick, dashdotted] (0.69,0.62) rectangle (.96,0.84);
            \end{scope}
        \end{tikzpicture}
    \end{subfigure}
    \begin{subfigure}{0.28\linewidth}
    \centering
        \includegraphics[width=\linewidth, bb=0 0 1920 1080, trim=50cm 23cm 0cm 7cm, clip=true]{Graphics/results/Seq_1_t_8_01.jpg}
    \end{subfigure}\hspace{0.0005em}
    \begin{subfigure}{0.28\linewidth}
        \includegraphics[width=\linewidth, bb=0 0 1920 1080, trim=50cm 24cm 0cm 6cm, clip=true]{Graphics/results/Seq_1_t_8_02.jpg}
    \end{subfigure}\hspace{0.0005em}
    \begin{subfigure}{0.28\linewidth}
        \includegraphics[width=\linewidth, bb=0 0 1920 1080, trim=47cm 24cm 3cm 6cm, clip=true]{Graphics/results/Seq_1_t_8_03.jpg}
    \end{subfigure}
    \begin{subfigure}{1\linewidth}
        \centering
        \includegraphics[width=.84\linewidth]{Graphics/results/DiMP_Legends.pdf}
    \end{subfigure}
    \caption{Comparison between DiMP-2D and DiMP-3D. During~camera motion DiMP-2D is unable to track the object; whereas~DiMP-3D is able to stabilize the tracking by expressing the object position in the 3D scene space. The~light-blue grid is drawn to help visualize ego-motion.}
    \label{fig:19}
\end{figure}

\begin{figure}[H]
    \centering
    \begin{subfigure}{0.28\linewidth}
        \begin{tikzpicture}
            \node[anchor=south west,inner sep=0] (image) at (0,0) {\includegraphics[width=\linewidth, bb=0 0 1920 1080]{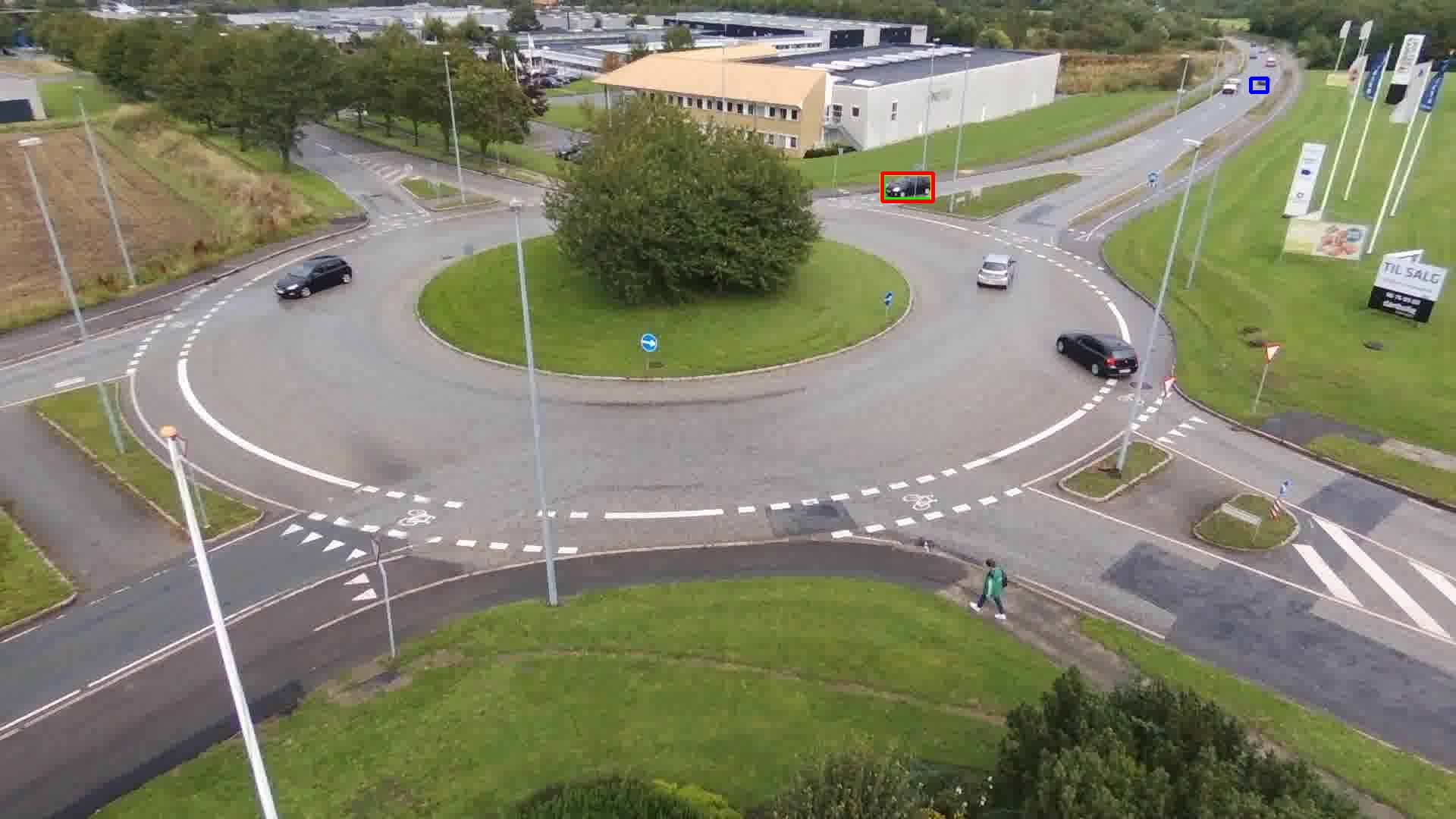}};
            \begin{scope}[x={(image.south east)},y={(image.north west)}]
                \draw[cyan!40!, ultra thin, xstep=.2, ystep=.2] (0,0) grid (1,1);
                \draw[red, thick, dashdotted] (0.47,0.63) rectangle (.73,0.85);
            \end{scope}
        \end{tikzpicture}
    \end{subfigure}\hspace{0.0005em}
    \begin{subfigure}{0.28\linewidth}
        \begin{tikzpicture}
            \node[anchor=south west,inner sep=0] (image) at (0,0) {\includegraphics[width=\linewidth, bb=0 0 1920 1080]{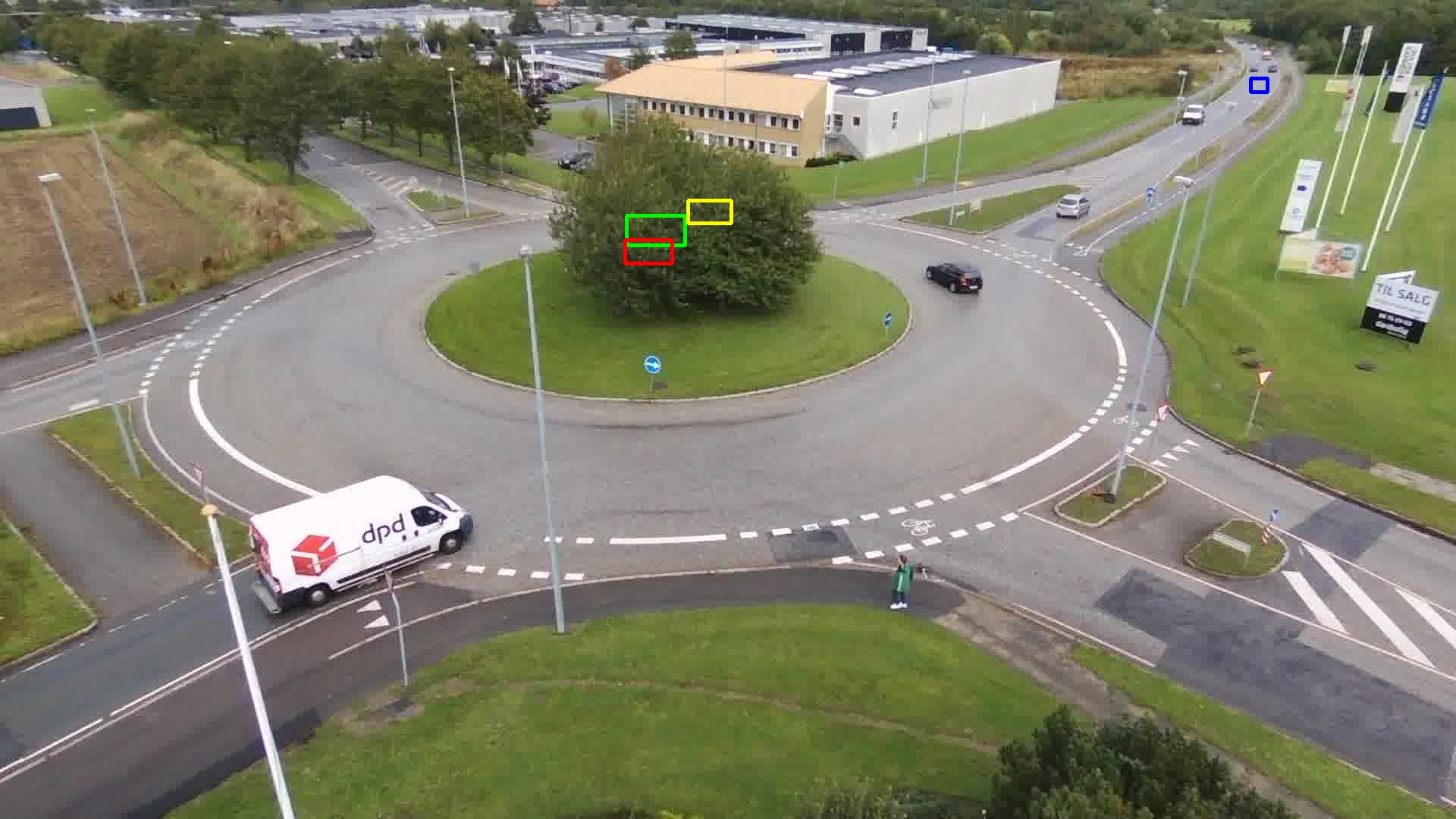}};
            \begin{scope}[x={(image.south east)},y={(image.north west)}]
                \draw[cyan!40!, ultra thin, xstep=.2, ystep=.2] (0,0) grid (1,1);
                \draw[red, thick, dashdotted] (0.35,0.55) rectangle (.62,0.77);
            \end{scope}
        \end{tikzpicture}
    \end{subfigure}\hspace{0.0005em}
    \begin{subfigure}{0.28\linewidth}
        \begin{tikzpicture}
            \node[anchor=south west,inner sep=0] (image) at (0,0) {\includegraphics[width=\linewidth, bb=0 0 1920 1080]{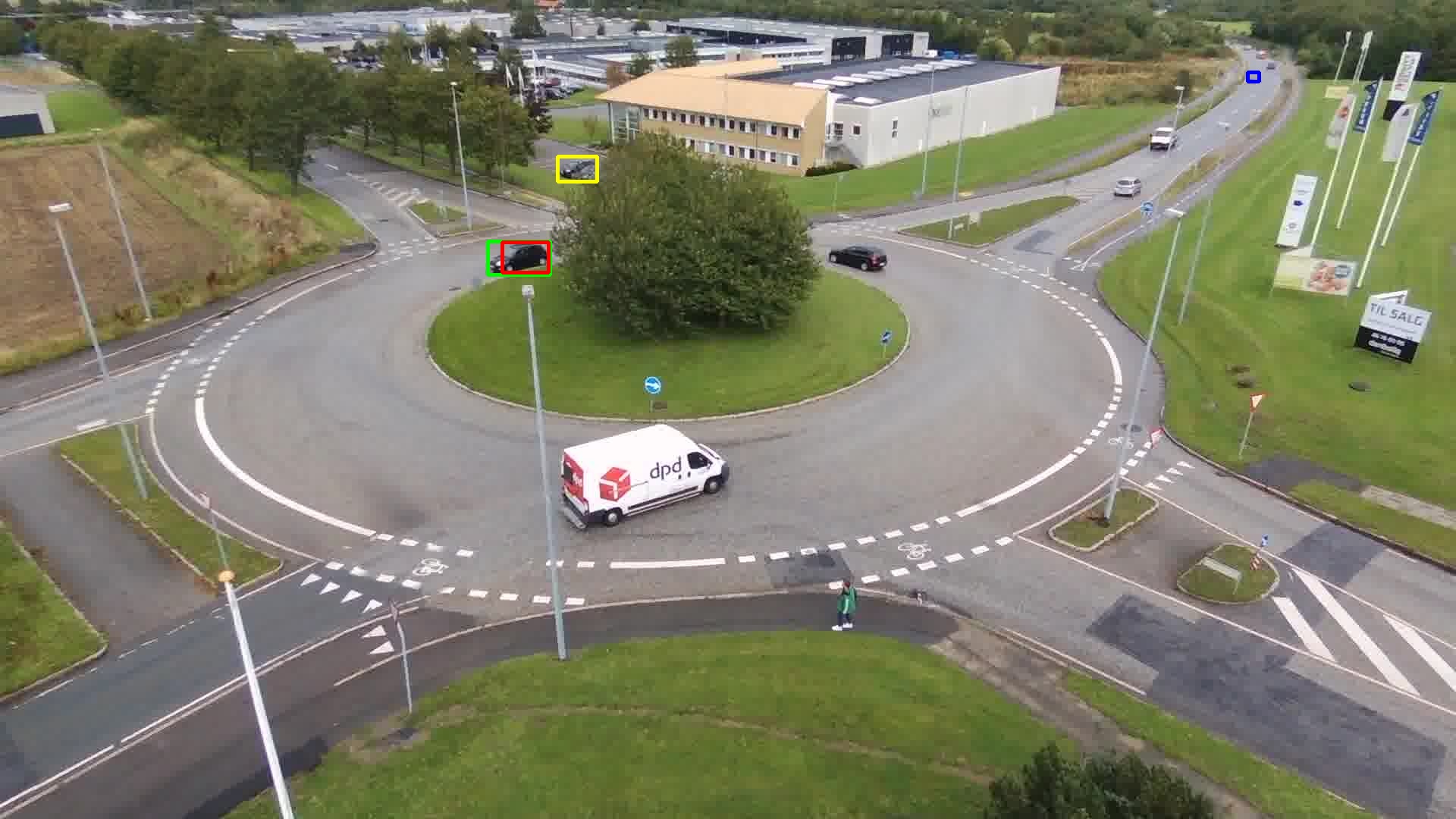}};
            \begin{scope}[x={(image.south east)},y={(image.north west)}]
                \draw[cyan!40!, ultra thin, xstep=.2, ystep=.2] (0,0) grid (1,1);
                \draw[red, thick, dashdotted] (0.26,0.6) rectangle (.51,0.82);
            \end{scope}
        \end{tikzpicture}
    \end{subfigure}
    \begin{subfigure}{0.28\linewidth}
    \centering
        \includegraphics[width=\linewidth, bb=0 0 1920 1080, trim=32cm 24cm 18cm 6cm, clip=true]{Graphics/results/Seq_0_t_6_00.jpg}
    \end{subfigure}\hspace{0.0005em}
    \begin{subfigure}{0.28\linewidth}
        \includegraphics[width=\linewidth, bb=0 0 1920 1080, trim=24cm 21cm 26cm 9cm, clip=true]{Graphics/results/Seq_0_t_6_02.jpg}
    \end{subfigure}\hspace{0.0005em}
    \begin{subfigure}{0.28\linewidth}
        \includegraphics[width=\linewidth, bb=0 0 1920 1080, trim=18cm 23cm 32cm 7cm, clip=true]{Graphics/results/Seq_0_t_6_03.jpg}
    \end{subfigure}
    \begin{subfigure}{1\linewidth}
        \centering
        \includegraphics[width=.84\linewidth]{Graphics/results/DiMP_Legends.pdf}
    \end{subfigure}
    \caption{Comparison between DiMP-2D and DiMP-3D. Only~DiMP-3D is able to robustly track the object during ego-motion while the object is occluded. The~light-blue grid is drawn to help visualize~ego-motion.}
    \label{fig:21}
\end{figure}

\begin{figure}[H]
    \centering
    \begin{subfigure}{0.28\linewidth}
        \begin{tikzpicture}
            \node[anchor=south west,inner sep=0] (image) at (0,0) {\includegraphics[width=\linewidth, bb=0 0 1920 1080]{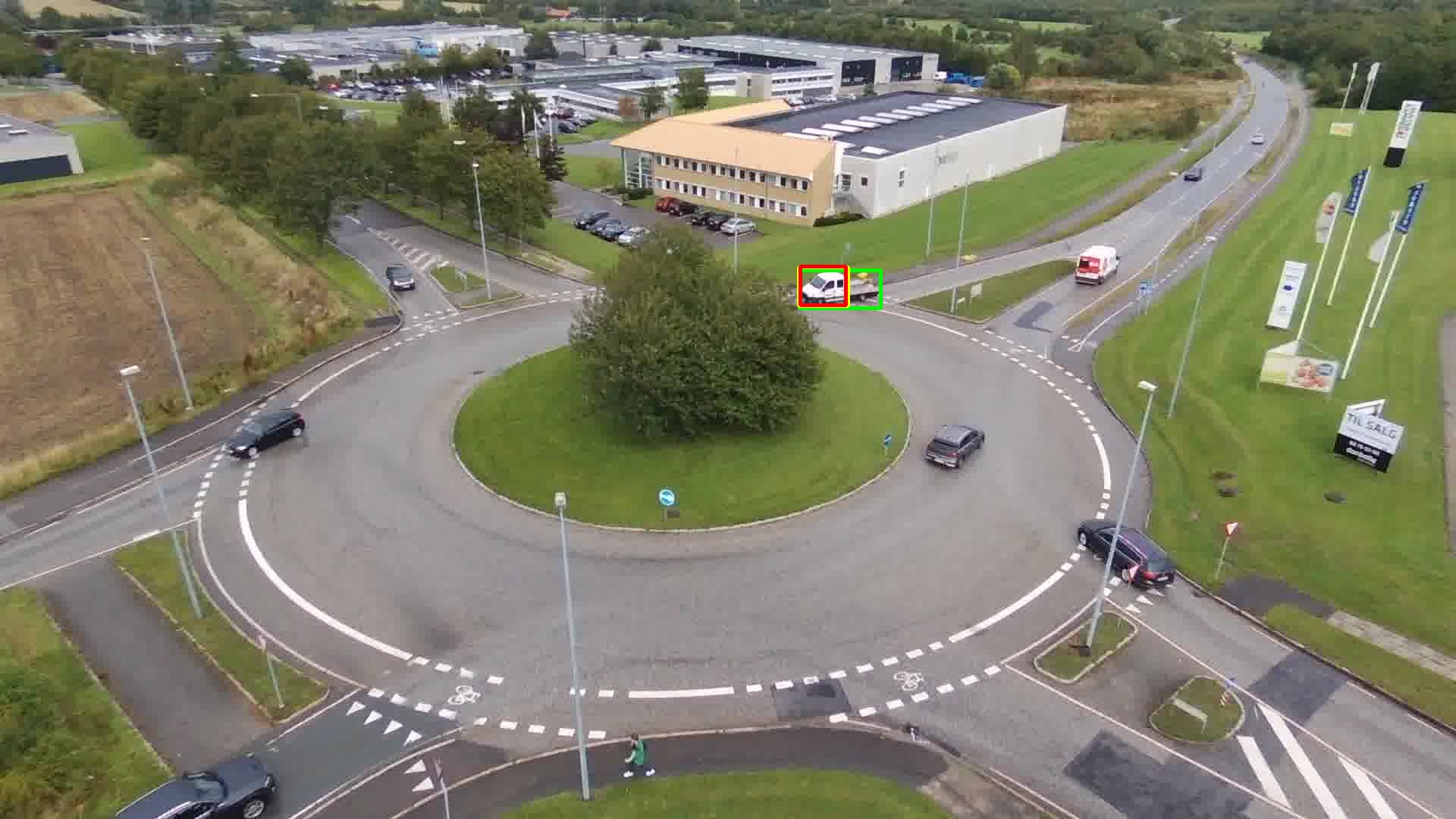}};
            \begin{scope}[x={(image.south east)},y={(image.north west)}]
                \draw[red, thick, dashdotted] (0.46,0.56) rectangle (.72,0.76);
            \end{scope}
        \end{tikzpicture}
    \end{subfigure}\hspace{0.0005em}
    \begin{subfigure}{0.28\linewidth}
        \begin{tikzpicture}
            \node[anchor=south west,inner sep=0] (image) at (0,0) {\includegraphics[width=\linewidth, bb=0 0 1920 1080]{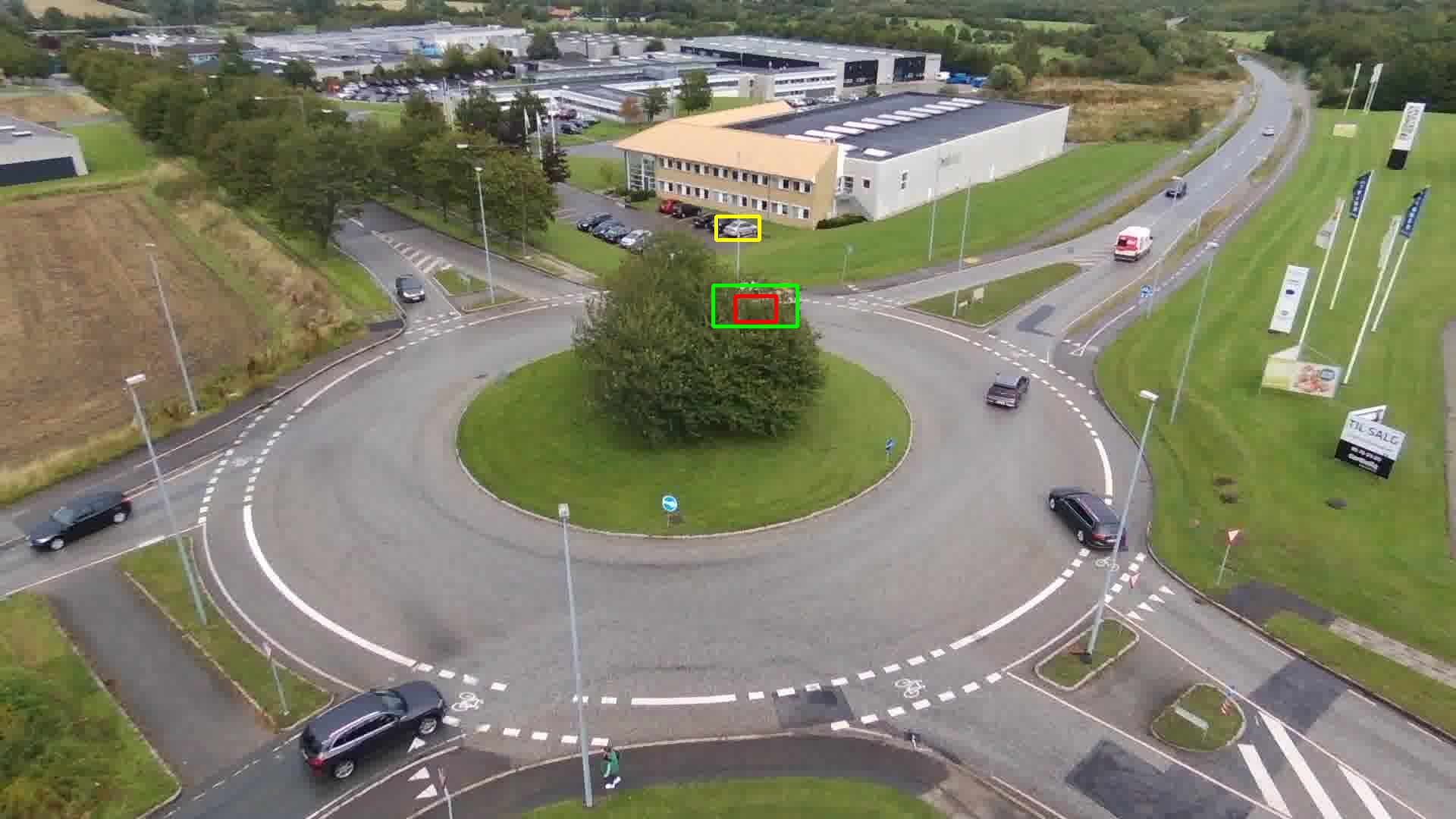}};
            \begin{scope}[x={(image.south east)},y={(image.north west)}]
                \draw[red, thick, dashdotted] (0.37,0.55) rectangle (.63,0.75);
            \end{scope}
        \end{tikzpicture}
    \end{subfigure}\hspace{0.0005em}
    \begin{subfigure}{0.28\linewidth}
        \begin{tikzpicture}
            \node[anchor=south west,inner sep=0] (image) at (0,0) {\includegraphics[width=\linewidth, bb=0 0 1920 1080]{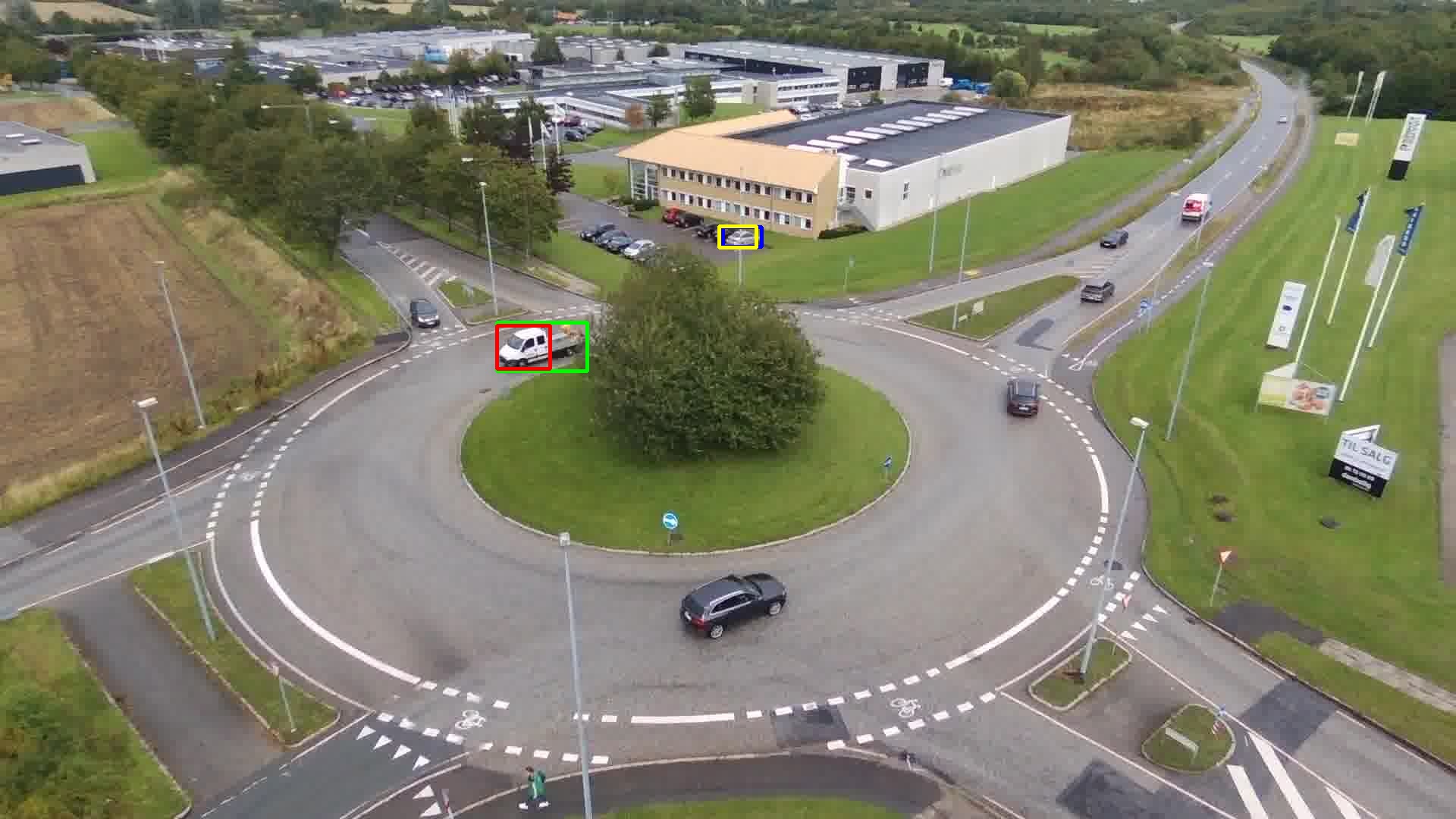}};
            \begin{scope}[x={(image.south east)},y={(image.north west)}]
                \draw[red, thick, dashdotted] (0.3,0.53) rectangle (.55,0.74);
            \end{scope}
        \end{tikzpicture}
    \end{subfigure}
    \begin{subfigure}{0.28\linewidth}
    \centering
        \includegraphics[width=\linewidth, bb=0 0 1920 1080, trim=31cm 21cm 19cm 9cm, clip=true]{Graphics/results/Seq_0_t_10_00.jpg}
    \end{subfigure}\hspace{0.0005em}
    \begin{subfigure}{0.28\linewidth}
        \includegraphics[width=\linewidth, bb=0 0 1920 1080, trim=25cm 21cm 25cm 9cm, clip=true]{Graphics/results/Seq_0_t_10_01.jpg}
    \end{subfigure}\hspace{0.0005em}
    \begin{subfigure}{0.28\linewidth}
        \includegraphics[width=\linewidth, bb=0 0 1920 1080, trim=20cm 20cm 30cm 10cm, clip=true]{Graphics/results/Seq_0_t_10_02.jpg}
    \end{subfigure}
    \begin{subfigure}{1\linewidth}
        \centering
        \includegraphics[width=.84\linewidth]{Graphics/results/DiMP_Legends.pdf}
    \end{subfigure}
    \caption{Comparison between DiMP variations. Owing~to the depth information used for occlusion identification in addition to visual cues, DiMP-3D manages to handle the object undergoing occlusion; whereas~DiMP and DiMP-2D switch to a distractor caused by relying solely on visual cues.}
    \label{fig:21c}
\end{figure}

\figref{fig:19} shows three consecutive frames where only DiMP-3D is unaffected by the ego-motion and can track the object robustly because the position of the object is expressed in the 3D scene space.
\figref{fig:21} illustrates another scenario where both DiMP-2D and DiMP-3D recognize the object as hidden, but~while predicting the position of the hidden object; only~DiMP-3D can predict robust and reasonable positions for the object even during camera motion. Relying~on an image-based state estimator is only viable when ego-motion is very minimal, as in \figref{fig:20}. In~\figref{fig:21c}, solely~DiMP-3D can identify the object undergoing occlusion, owing~to the depth information leveraged by the depth map approximations in addition to the visual cues; whereas~DiMP and DiMP-2D switch to a distractor because they solely rely on visual cues.

\begin{figure}[H]
    \centering
    \begin{subfigure}{0.28\linewidth}
        \begin{tikzpicture}
            \node[anchor=south west,inner sep=0] (image) at (0,0) {\includegraphics[width=\linewidth, bb=0 0 1920 1080]{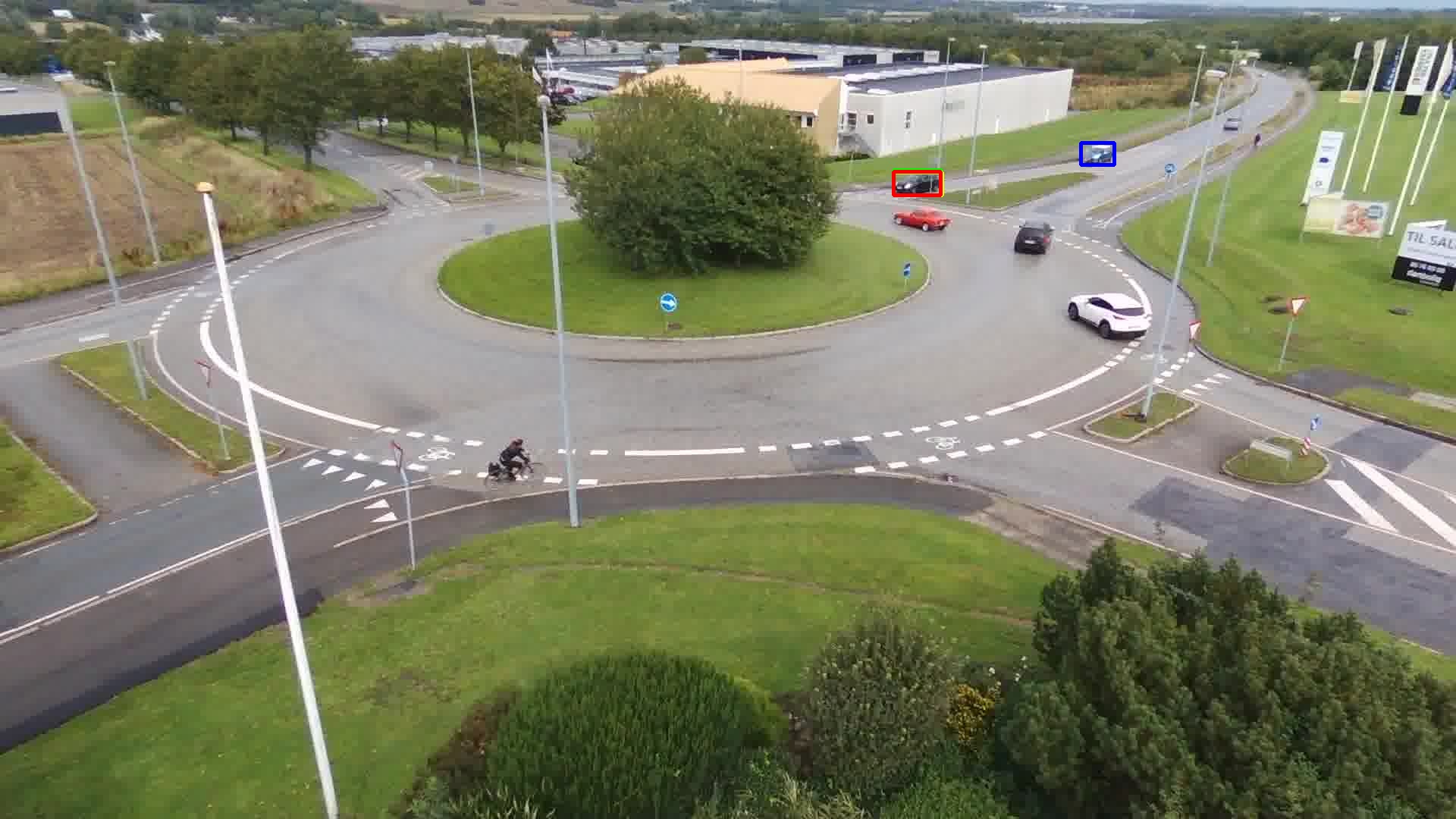}};
            \begin{scope}[x={(image.south east)},y={(image.north west)}]
                \draw[red, thick, dashdotted] (0.53,0.65) rectangle (.79,0.87);
            \end{scope}
        \end{tikzpicture}
    \end{subfigure}\hspace{0.0005em}
    \begin{subfigure}{0.28\linewidth}
        \begin{tikzpicture}
            \node[anchor=south west,inner sep=0] (image) at (0,0) {\includegraphics[width=\linewidth, bb=0 0 1920 1080]{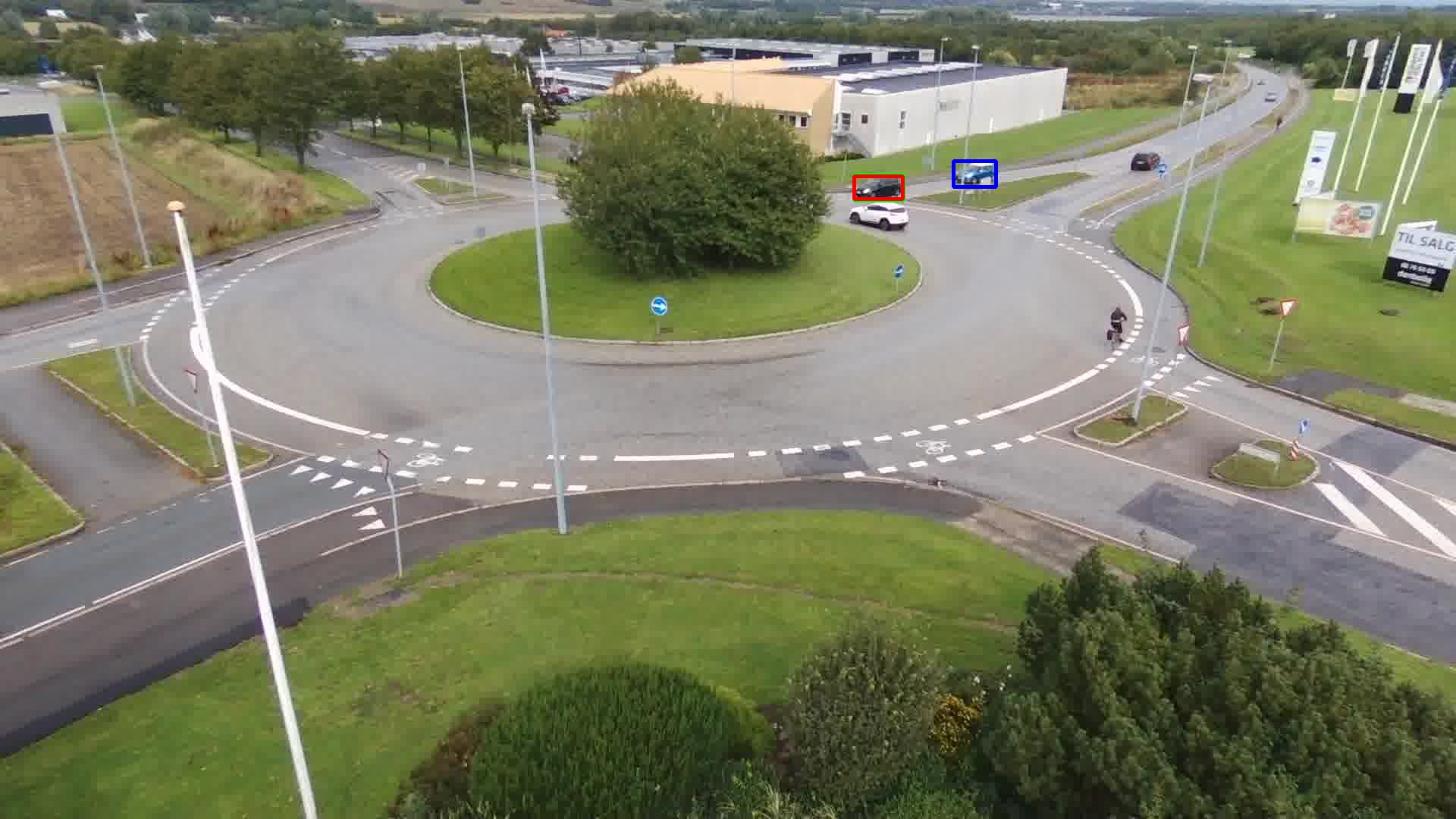}};
            \begin{scope}[x={(image.south east)},y={(image.north west)}]
                \draw[red, thick, dashdotted] (0.53,0.65) rectangle (.79,0.87);
            \end{scope}
        \end{tikzpicture}
    \end{subfigure}\hspace{0.0005em}
    \begin{subfigure}{0.28\linewidth}
        \begin{tikzpicture}
            \node[anchor=south west,inner sep=0] (image) at (0,0) {\includegraphics[width=\linewidth, bb=0 0 1920 1080]{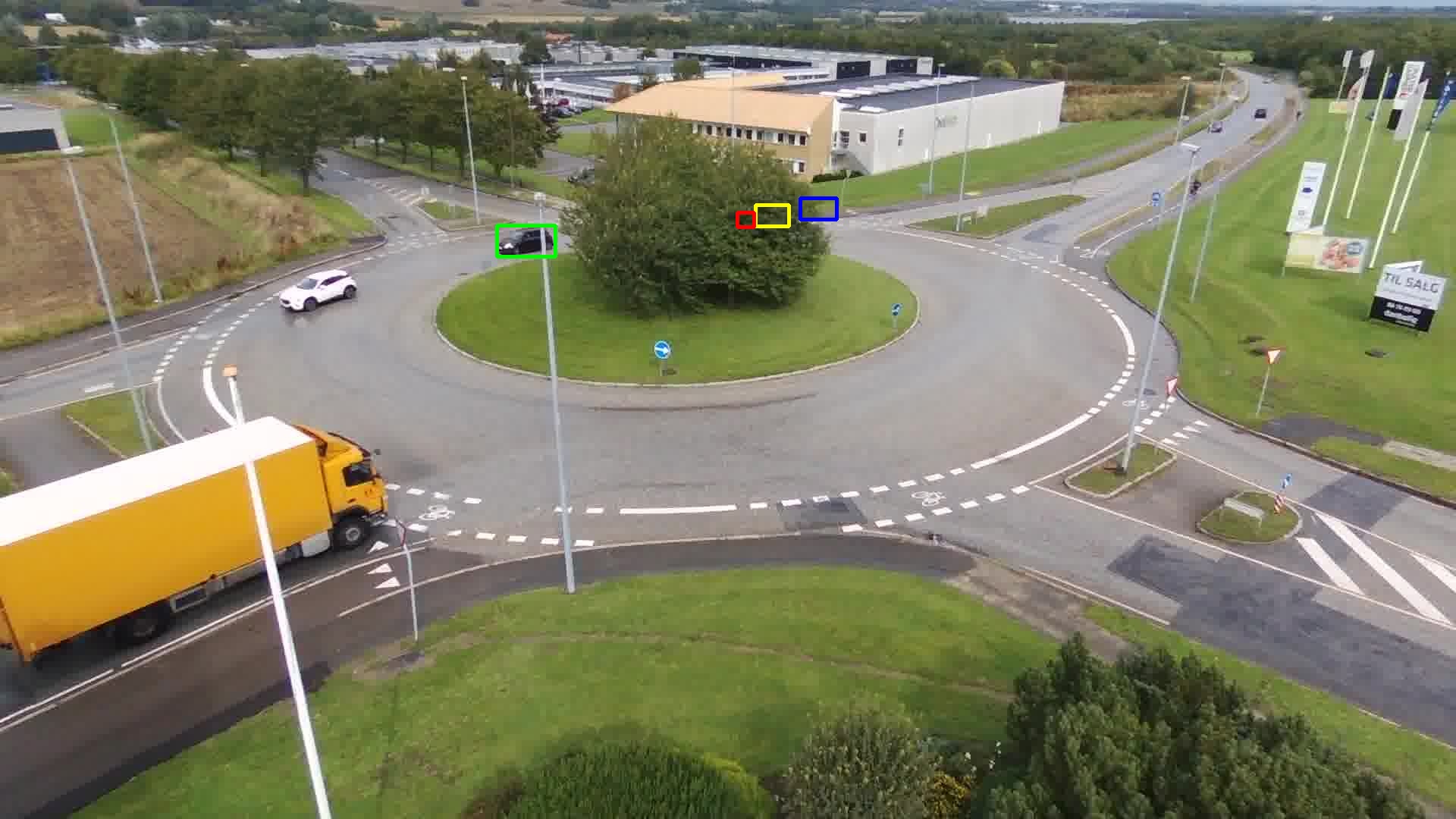}};
            \begin{scope}[x={(image.south east)},y={(image.north west)}]
                \draw[red, thick, dashdotted] (0.32,0.61) rectangle (.58,0.84);
            \end{scope}
        \end{tikzpicture}
    \end{subfigure}
    \begin{subfigure}{0.28\linewidth}
    \centering
        \includegraphics[width=\linewidth, bb=0 0 1920 1080, trim=35cm 25cm 14cm 5cm, clip=true]{Graphics/results/Seq_0_t_40_01.jpg}
    \end{subfigure}\hspace{0.0005em}
    \begin{subfigure}{0.28\linewidth}
        \includegraphics[width=\linewidth, bb=0 0 1920 1080, trim=35cm 25cm 14cm 5cm, clip=true]{Graphics/results/Seq_0_t_40_02.jpg}
    \end{subfigure}\hspace{0.0005em}
    \begin{subfigure}{0.28\linewidth}
        \includegraphics[width=\linewidth, bb=0 0 1920 1080, trim=22cm 24cm 28cm 6cm, clip=true]{Graphics/results/Seq_0_t_40_03.jpg}
    \end{subfigure}
    \begin{subfigure}{1\linewidth}
        \centering
        \includegraphics[width=.84\linewidth]{Graphics/results/DiMP_Legends.pdf}
    \end{subfigure}
    \caption{Comparison between DiMP-2D and DiMP-3D. During~the acceleration phase of the object, the~particle filter estimates the velocity of the object with a delay due to the transition model adopted.}
    \label{fig:23}
\end{figure}

Despite DiMP-3D and ATOM-3D achieving remarkable results, there~are cases where both fail. Occlusion~on rare occasions is not identified correctly because~of a strong distractor present when the object is partially hidden. This~can be prevented by elaborating a different strategy for recognizing occlusions and reappearances of the object. Another~point limiting the performance of DiMP-3D is illustrated in \figref{fig:23}, where~the object slows down at the intersection for a long period. While~the object is not moving, the~particle filter continuously updates the estimated velocity to be adequate with the observations (velocity near zero). When~the object accelerates, the~particle filter cannot match the speed instantly, due~to the transition model.

\section{Discussion}
\label{discussion}
{
Besides the challenges arising from the specific characteristics for single visual object tracking from UAVs, the~use of computer vision approaches onboard a UAV additionally faces the problem of finding an adequate compromise between computational complexity and real-time capabilities with extreme resource limitations on the platform. Although, being~not in the scope of this paper, we~explore in this section alternative design choices for integrating the current pipeline onto a UAV. Owing~to the modular design of our pipeline, we~can replace individual components with variants that are less cost-intensive.

Regarding the \emph{Visual Object Tracker} component, both~ATOM~\cite{bib:atom} and DiMP~\cite{bib:dimp} have real-time capabilities but rely on a Graphics Processing Unit (GPU) 
for inferring the position of the object. To~diminish the amount of space required and to increase the run time, the~original feature extractor---i.e., ResNet-18, ResNet-50~\cite{bib:ResNet}---could be replaced with MobileNet~\cite{bib:MobileNet}, which~is specifically designed for embedded vision applications and mobile devices. Alternatively, it~is possible to replace ATOM and DiMP with another type of tracker. Since~both trackers update the appearance model of the object on-the-fly, they~require more GPU capacities than other methods not updating the appearance model such as Siamese SOT~\cite{bib:siam_0, bib:siam_2, bib:siam_3, bib:siam_4}.

In our work, we~focused on using image-based scene reconstruction by leveraging a SfM-based method~\cite{bib:colmap0}. To~attain real-time performance, a~Visual SLAM-based method that can associate IMU formations~\cite{bib:VI-DSO,bloesch2015robust,li2013high,leutenegger2015keyframe,forster2015imu,usenko2016direct,mur2017visual} is preferred for robustly reconstructing the 3D environment on-the-fly. A~concern for the sparse reconstruction might be the storage and the processing time needed when the UAV is observing a large area. To~reduce the required storage space needed, the~point cloud can be reconstructed or partially loaded, depending~on the current UAV position in the scene~\cite{bib:subarea}.

Since the multimodal representation of the probability density function is indispensable, for~identifying distractors in the search area, a~particle filter~\cite{bib:particle_f_tuto} is utilized in our detection-by-tracking pipeline. Thus, using~a particle filter over other state estimators such as the Kalman filter~\cite{welch1995introduction} is crucial for the proposed pipeline, despite~being computationally more demanding.
Although numerous particle filter implementations do not perform well with a high number of particles, this~is not necessarily a general limitation of the approach~\cite{chao2010efficient}. In~an effort to reduce the computational time, authors~from~\cite{murray2016parallel} elaborate faster methods for the resampling step compared to common resampling~approaches.}

\section{Conclusions}
\label{conclusion}
In this paper, we~propose an approach to improve UAV onboard {single} visual object tracking. To~this end, we~combine information extracted from a visual tracker and 3D cues of the observed scene. The~3D reconstruction allows us to estimate the state in a 3D scene space rather than in a 2D image space. Therefore, we~can define a 3D transition model reflecting the dynamics of the object close to~reality.

The potential of the approach is shown on challenging real-world sequences, illustrating typical occlusion situations captured from a low-altitude UAV. The~experiments demonstrate that the presented framework has several advantages and is viable for UAV onboard visual object tracking. We~can effectively handle object occlusions, low~object sizes, the~presence of distractors, and reduced tracking errors caused by ego-motion.

\textls[-10]{A part of our future work will be to exploit a dense reconstruction rather than a sparse reconstruction; explore~different state estimators and add more context to the scene, such~as the layout of the road in the reconstruction; {and to integrate real-world coordinates through georeferencing}}.

\vspace{6pt}



\authorcontributions{Conceptualization by the first author S.V. and~the co-authors S.B. (Stefan Becker) and T.B. Data~curation, formal~analysis, investigation, methodology, software, validation, visualization, and writing---original draft, were~done by the first author S.V. The~co-authors S.B. (Stefan Becker) and~S.B. (Sebastian Bullinger) improved~the paper through their comments and corrections about the layout, content, and~the obtained results. Supervision~and project administration were done by N.S.-N. Funding~acquisition was done by M.A. All~authors have read and agreed to the published version of the manuscript.}

\funding{This research was funded by the German Ministry of Defence.}

\acknowledgments{We would like to thank Ann-Kristin Grosselfinger, who~provided the annotations for the AU-AIR-Track dataset.}

\conflictsofinterest{The authors declare no conflict of interest.}

\abbreviations{The following abbreviations are used in this manuscript:\\

\noindent
\begin{tabular}{@{}ll}
UAV & Unmanned Aerial Vehicle\\
SOT & Single Object Tracking\\
MOT & Multiple Object Tracking\\
SfM & Structure from Motion\\
SLAM & Simultaneous Localization and Mapping\\
IMU & Inertial Measurement Unit\\
RANSAC & Random Sample Consensus\\
GPS & Global Positioning System\\
IoU & Intersection-over-Union\\
GPU & Graphics Processing Unit\\
\end{tabular}}



\reftitle{References}

\bibliography{Bib.bib}

\end{document}